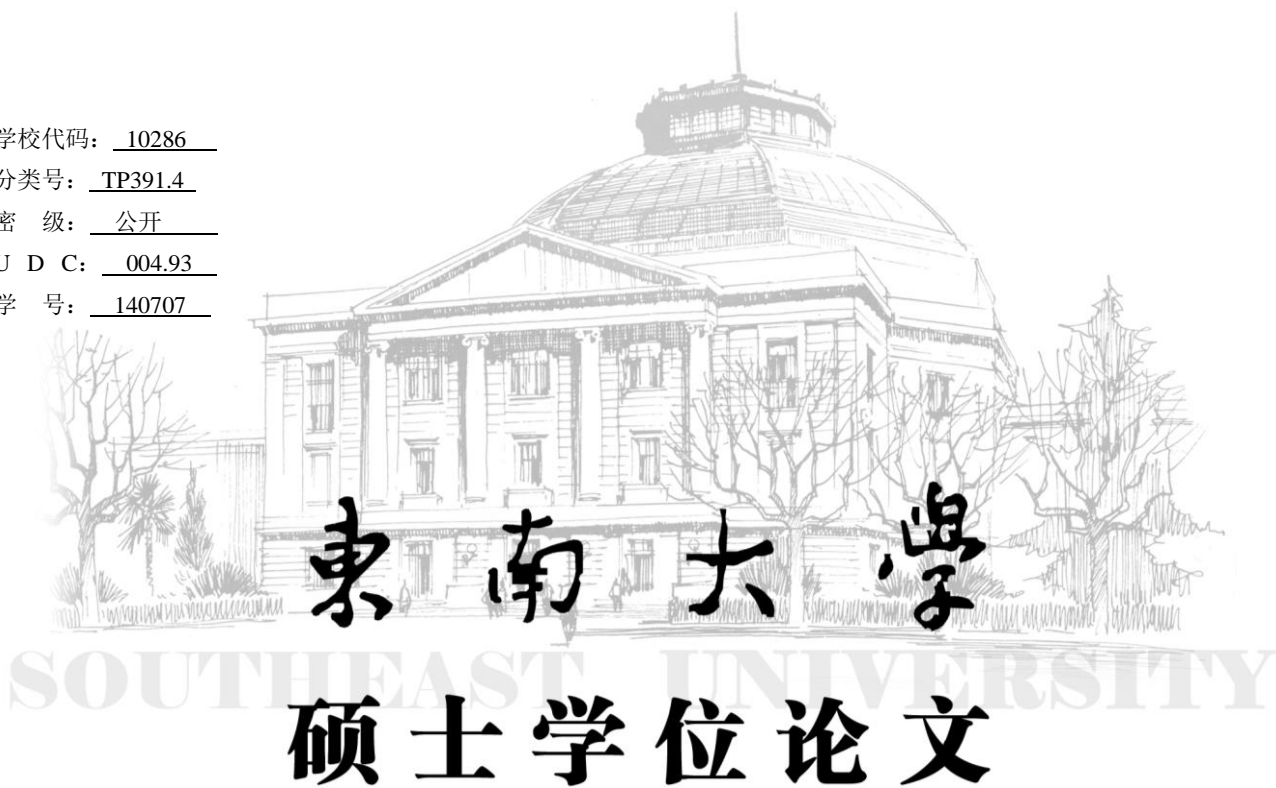

# 东 南 大 学

## 硕 士 学 位 论 文

## 基于深度学习的双模态生物特征识别研究

研究生姓名： **蒋 浩**

导 师 姓 名： **邹采荣**

申请学位类别 ___工学硕士___ 学位授予单位 **东 南 大 学**

一级学科名称 ___信息与通信工程___ 论文答辩日期 **20 年 月 日**

二级学科名称 ___信号与信息处理___ 学位授予日期 **20 年 月 日**

答辩委员会主席 ___________________ 评 阅 人 ___________________

___________________

2017 年 4 月 17 日

# 东南大学
# 硕士学位论文

## 基于深度学习的双模态生物特征识别研究

**专 业 名 称：**　　信息与通信工程

**研究生姓名：**　　　蒋　浩

**导 师 姓 名：**　　　邹采荣

# Research on Bi-mode Biometrics Based on Deep Learning

A Thesis Submitted to
Southeast University
For the Academic Degree of Master of Science

BY
Hao Jiang

Supervised by
Professor Cai Rong Zou

School of Information Science and Engineering
Southeast University
April 2017

# 东南大学学位论文独创性声明

本人声明所呈交的学位论文是我个人在导师指导下进行的研究工作及取得的研究成果。尽我所知，除了文中特别加以标注和致谢的地方外，论文中不包含其他人已经发表或撰写过的研究成果，也不包含为获得东南大学或其它教育机构的学位或证书而使用过的材料。与我一同工作的同志对本研究所做的任何贡献均已在论文中作了明确的说明并表示了谢意。

研究生签名：________________日期：________________

# 东南大学学位论文使用授权声明

东南大学、中国科学技术信息研究所、国家图书馆有权保留本人所送交学位论文的复印件和电子文档，可以采用影印、缩印或其他复制手段保存论文。本人电子文档的内容和纸质论文的内容相一致。除在保密期内的保密论文外，允许论文被查阅和借阅，可以公布（包括以电子信息形式刊登）论文的全部内容或中、英文摘要等部分内容。论文的公布（包括以电子信息形式刊登）授权东南大学研究生院办理。

研究生签名：________导师签名：________日期：__________

# 摘 要

鉴于生物特征具有优秀的独立区分特性，生物特征识别技术几乎涉及到所有区分人的相关领域。指纹、虹膜、人脸、声纹等生物特征已经广泛使用到公安部门破案侦查，移动设备解锁，目标跟踪等领域。随着电子设备使用的范围越来越广，使用的频率越来越高。具有优秀识别率的生物特征识别技术才能保证这些领域的长久发展。

由于现实生活中使用生物特征识别的环境条件的差异。单一的生物特征识别因为其自身条件的局限性，导致无论哪一种生物特征有会有其应用的局限性。例如指纹识别使用的前提是设备与目标人物必须有肢体的接触，视频监控等远距离设备无法使用指纹识别技术；在光线不足或是摄像头没有正对目标人物脸部的情况下人脸识别技术的识别率将会急剧下降；同样，虹膜识别技术也要保证目标人物眼部靠近传感器，才能实现此生物特征的后续识别过程。多生物特征融合识别技术可以很好的解决这个问题。多模态生物特征融合识别技术可以根据特征的选取和融合方法提高识别设备的准确性，普适性和鲁棒性。本文的主要研究内容和实验成果如下：

1. 研究了基于深度卷积网络（Convolutional Neural Network,CNN）和的人脸识别，创新的提出了基于 Vgg_Face 改进模型的人脸识别算法。该方法结合深度卷积网络模型的核心思想，在 Vgg_Face 模型的最后一层添加了全连接层降低人脸特征维度。在 Vgg_Face 模型的基础上微调（fine-turning）新的模型。使用 CASIA-Webface 人脸数据库训练网络模型。

2. 首先研究了说话人的感知预测系数（Perceptual Linear Predictive，PLP），然后推导了说话人的 I-Vector 特征提取方法，I-Vector 特征反映说话人的差异，并且具有优秀的跨信道性能，是目前说话人识别主流的识别特征。然后将 I-Vector 和 PLP 特征融合送入深度置信网络（Deep Belief Network，DBN）训练，生成说话人识别模型。

3. 创新的提出了人脸模型和说话人特征融合识别方法。结合 TED-LIUM 语音库和 CASIA-WebFace 人脸库，将不同的人脸和语音随机组合成新的人脸—说话人综合库。将该库提取得到的融合特征去训练 DBN。最终得到的模型识别目标人物，其识别率比单独使用人脸特征或者说话人特征的识别率有一定得提高。

**关键词：人脸识别；说话人识别；深度学习；CNN；特征融合；DBN**

# ABSTRACT

In view of the fact that biological characteristics have excellent independent distinguishing characteristics, Biometric identification technology involves almost all the relevant areas of human distinction. Fingerprints, iris, face, voiceprint and other biological features have been widely used in the public security departments to detect detection, mobile equipment unlock, target tracking and other fields. With the use of electronic devices more and more widely and the frequency is getting higher and higher. Only the Biometric identification technology with excellent recognition rate can guarantee the long-term development of these fields.

Due to differences in environmental conditions using biometrics recognition in real life. Single biometrics because of its own limitations, lead to any kind of biometrics there will be limitations of its application. For example, the premise of the use of fingerprint identification equipment is the target people must have physical contact, video surveillance and other remote devices can not use fingerprint recognition technology; The recognition rate of face recognition technology will drop sharply if there is not enough light or if the camera is not facing the target person's face; Similarly, the iris recognition technology must ensure that the target person's eye close to the sensor in order to achieve this biological characteristics of the follow-up process. Multi-biometric fusion identification technology can be a good solution to this problem. Multi-modal biometric fusion recognition technology can improve the accuracy, universality and robustness of the identification device according to the feature selection and fusion method. The main contents and experimental results of this paper are as follows:

1. Based on the Convolutional Neural Network (CNN) and the face recognition, a new face recognition algorithm based on Vgg_Face improved model is proposed. This method combines the core idea of the deep convolution network model. In the last layer of the Vgg_Face model, the full join layer is added to reduce the face feature dimension. Fine-turning a new model based on the Vgg_Face model. Use the CASIA-Webface face database to train the network model.

2. Firstly, the Perceptual Linear Predictive (PLP) of the speaker is studied, and then the speaker's I-Vector feature extraction method is deduced. The I-Vector feature reflects the speaker's differences and has excellent cross-channel performance. It's the mainstream identification characteristics of Speaker identify. And then sent the I-Vector and PLP features into the deep confidence network (Deep Belief Network, DBN) to training speaker recognition model.

3 The fusion method of face model and speaker character fusion is put forward. Combined with TED-LIUM speech database and CASIA-WebFace face database, different face and voice randomly will be combined into a new face - speaker integrated library. The dataset is extracted from the fusion feature to train the DBN. The resulting model identifies the target person whose recognition rate is higher than the recognition rate of the individual face feature or speaker feature alone.

# 目 录

# 附图目录

# 附表目录

# 第一章 绪论

## 1.1 课题的研究背景和意义

随着社会信息化技术的快速发展，人与机器之间的信息交互高度密集。生物特征识别技术是人工智能和模式识别领域研究的热门方向。生物特征作为人类重要的身份识别标志，作为交易密码或者移动设备验证密钥，具有输入便捷、容易记忆、识别迅速和不易被盗取的优点。生物特征识别已经成为目标人物查找和确认的关键手段。目前在公共安全，银行金融安全，刑侦破案等领域有广泛的应用[1]。

目标人物自身携带的独有特点都可以作为其生物特征，包括行为特征和身体特征两大类。行为特征包含走路姿态、签字、语音、面部表情等等，身体特征包含人脸、指纹、视网膜、静脉、DNA、掌纹、虹膜等等[1]。不同的生物特征都有其独有的优点和缺点，可以适用于不同的环境。同样，不同的环境会制约着某种生物特征的识别效果。为了提高生物特征识别技术对环境的鲁棒性和普适性，本文提出了人脸和说话人双模态生物特征融合识别技术。

语音和图像是人类生活中重要的两类信息交流载体，并且在人机信息传输中十分方便。机器基于数字图像和数字语音的识别技术成为目标人物身份识别的重要手段。语音和人脸两种生物特征具有远程采集，良好的隐蔽性和非侵害性等优点。可以被用在视频监控，移动设备解锁，交易保密，出入境安检等领域。

说话人语音属于行为特征，人脸属于身体特征，这两种特征之间独立，在特征空间中相互正交。人脸属于二维图像，受到光照、表情变化、姿态等条件影响，但是语音信号不会受到这些环境变化干扰。说话人语音会受到生理状态、心理状态、环境噪声等条件影响，人脸图像特征可以很好的克服此类干扰。单一特征识别在实际应用中受到环境的影响较大。这两种不同属性的生物特征相互融合具有如下优势：1.可以很好的降低生物特征密码被盗和仿制的风险；2.可以提高目标人物在恶劣环境中的识别准确率。在机场，军队，银行等公共场所复杂环境中，双模态融合能够很好的取长补短，提高整体识别系统的鲁棒性。

中国、美国等国家目前已经设立相关机构，建立全国人脸，语音、指纹等生物特征库。随着国家提出大力发展城市物联网，感知中国等一系列和人工智能相关技术。生物特征融合识别技术在未来将会有巨大的发展潜力和广阔的应用场景。本文研究的双模态生物特征融合识别技术不仅提高了现有的目标人物识别系统的识别率、鲁棒性和普适性，而且进一步的推动了国家目前要大力发展的人工智能技术。此技术在国家安防，民

众生活方式和国民生产中具有深远的影响。

## 1.2 国内外研究历史与现状

人脸和说话人特征融合识别涉及到图像、语音处理和信息融合技术等多个研究方向。过去的几十年中，人脸和说话人识别分别都取得很大的进步，作为一个交叉学科的前沿研究课题。双模态生物特征识别基于人脸和说话人识别的发展，也取得了一定得进展。下面就人脸、说话人识别的研究历史、现状和存在的问题做进一步分析。

### 1.2.1 人脸识别的研究历史与现状

在过去 40 多年，心理生理学家，神经科学家和工程师对人脸机器识别的各个方面进行了广泛的研究[2][3]。人脸识别的发展进程可以分为：1. 人脸特征提取技术发展；2. 分类器算法发展。过去几十年里已经提出几十种人脸特征提取技术，大致上可以分成三类：1. 提取面部全局特征；2. 提取面部局部特征；3. 同时提取面部局部和全局融合特征。下面将对人脸识别研究的历史分几个阶段进行介绍：

第一阶段（1950s-1990）：最早的人脸识别技术可以追溯到 20 世纪 50 年代的心理学领域（Bruner 和 Tagiuri，1954）和 60 年代的工程应用文献（Bledsoe，1964）[4]。此阶段的人脸识别通常被作为一般的模式识别来研究。主要是分析人脸的几何特征，在人脸结构、轮廓等特征提取做了较深的研究。包括眼睛、嘴巴、鼻子等明显特征间的角度，距离等参数特征。其中 Kelly（1970 年）和 Kanade（1973 年）开创性的提出机器自动识别人脸。之后一些研究室也提出将人工神经网络应用到人脸识别，但是限于算法和计算速度的限制，取得的效果不大。这一阶段的人脸识别研究限于人为主观经验对人脸提取明显轮廓特征，此类浅层特征忽略了面部的全局特征和细节纹理特征，造成了大量信息缺失[5][6][7]。此阶段人脸识别算法受环境影响很大，实际应用很少。

第二阶段（1990-2006）：Mathew 和 Alex 在 1991 年提出将主成分分析（Principal Component Analysis，PCA）应用到人脸识别，将人脸投影到一个新的空间。在新的空间坐标系中，使用人脸的最大线性无关特征组作为坐标向量。此方法对人脸进行降维，很好的解决了计算速度不够的问题。Belhumeur（1996）提出 Fisher 特征脸算法，用线性判别分析（Linear Discriminant Analysis, LDA）变换人脸特征，通过最大化类间散度矩阵与类内散度矩阵之比来构建线性子空间。此方法推进人脸识别的实际应用，在此基础上又有 Liu 和 Wechsler 在 1998 提出增强 LDA（E-LDA）算法，Lotlikar 和 Kothari（2000年）提出分步 LDA（F-LDA）算法等改进算法[8]。随着研究的深入，针对实际应用中光照、面部表情和姿态对人脸识别造成的问题，研究人员提出非线性核技术。如 Scholkopf

（1998 年）提出的内核主成分分析（KPCA）算法，Baudat 和 Anouar（2000 年）提出的广义判别分析（Generalized Discriminant Analysis，GDA）等。

Wiskott 等人在 1997 年提出弹性图匹配法（Elastic Bunch Graph Matching，EBGM），使用 Gabor 小波对人脸进行特征提取。得到若干个具有一定拓扑结构信息的特征点，构成人脸弹性图。由于 Garbor 小波变换核与人脑皮层对二维图像发射特性相同，能够对人脸空间尺度，空间位置和方向信息提取。此算法对光照和人脸表情鲁棒。Timo Ahonen 和 Abdenour Hadid 在 2006 年提出的 LBP 人脸特征，反映脸部纹理特征，对光照有一定鲁棒性。由于 LBP 特征具有旋转不变性，可以一定程度上消除脸部二维旋转造成的影响。

人脸识别技术这一阶段发展非常迅速，在一些中小型数据库上有不错的识别率。对实际应用场景中存在的问题也有较好的解决方法。基于分类器的发展，AdaBoost、SVM 等分类器和 LBP、Garbor、Sift 等人脸特征相结合取得了不错的识别效果[10]。

第三阶段（2006-至今）：随着 Hinton 在 2006 年提出 DBN，再一次把深度学习领域推向了研究热门。深度学习大大推动了人脸识别技术的发展。在 LFW 人脸识别竞赛中，经典的 Eigenface 只有 60% 的识别率，非深度学习算法最好是取得了 96.33% 的识别率，而使用深度学习方法可以取得 99.47% 的识别率，并且还在不停的上升。国外的研究团队有 MIT，剑桥大学计算机视觉组以及 Facebook，Google，Microsoft 等 IT 公司。国内的研究团队有山世光教授所在的中科院计算所与哈工大联合面像实验室，中科院自动化所的李子青教授团队，汤晓鸥老师所在的科大讯飞与香港中文大学联合实验室以及百度、商汤科技等公司。他们在近年的 LFW 人脸识别竞赛中取得了惊人的成绩。

人脸识别发展到目前为止依然有其存在的问题，自然环境中的光照、图像的抖动、拍摄角度都会很大程度影响到人脸识别效果。传统人脸识别方法借用人为经验对特征提取的不够充分，并且特征应用的场景有其局限性。深度神经网络的人脸识别方法需要大量的训练数据以及训练时间，如何使用更少的数据和时间来收敛网络同样是限制人脸识别发展的问题。

### 1.2.2 说话人识别的研究历史和现状

说话人语音作为人类行为特征，20 世纪 30 年代有研究人员开始通过说话人语音判别身份的研究。早期的科研重点主要在于人耳辨别实验和探寻听音识别的可行性方面。随着数字信号处理理论和计算机技术的发展，机器自动识别说话人技术得到了快速的发展。说话人识别技术的发展可分为语音特征提取方法和模式匹配方法两个方面。

在 1945 年，Bell 实验室的 L.G.Kesta 提出通过人眼观察语谱图来识别说话人身份，并提出类似与指纹的"声纹"概念。并于 1962 年证明了声纹可用于自动说话人识别。1963 年 Bell 实验室的 S. Pruzansky 提出通过模式匹配和概率统计方差分析来识别说话人。这种方法引起了研究人员对说话人识别的研究高潮。在 1969 年，Luck JE 开创性地提出将倒谱技术用于说话人识别。其后 B.S Atal 于 1976 年在此基础上提出线性预测倒谱特征（Linear Prediction Cepstrum Coefficient， LPCC），并将其应用于说话人识别中，大幅提高了识别率。之后研究人员又在 LPCC 的基础上提出了 LPC，LSP 等语音特征用于说话人识别。S.B.Davis 在 1980 年提出了 Mel 倒谱系数（Mel Frequency Cepstrum Coefficient， MFCC）[11]。此方法模拟人耳听觉特性，将其反映在倒谱系数上，并且对环境噪声有不错地鲁棒性。这使得说话人识别系统可用于实际应用。直到今天，Mel 倒谱系数依然广泛应用于语音识别。

说话人语音特征提取之后需要对应的建模方法生成说话人模型，因此说话人识别伴随着模式匹配方法的发展。从早期直接提取语音频谱进行匹配到动态时间规整（DTW，Itakura，1975），矢量量化（Vector Quantization，VQ，F. Soong，1987），隐马尔科夫模型（Hidden Markov Model，HMM，Jayant M Naik，1989）和人工神经网络（Artificial neural networks，ANN）的出现。在 2000 年，Reynolds 提出了最大后验概率的统一背景模型（Universal Background Model-Maximum A Posterior，UBM-MAP），使得自动说话人识别技术成功的应用于商业。其中用高斯混合模型一通用背景模型（Gaussian Mixture Model - General Background Model，GMM-UBM）提取的 I-Vector 成为如今主流的说话人识别特征[12]。从 2014 年起，NIST-ASR 比赛中，I-Vector 为统一说话人特征。

模式匹配算法的发展使说话人识别技术不断改良。近几年出现的基于深度神经网络的深度学习技术，再一次将说话人识别技术推向了新的高度。通过大量的训练数据和超大规模的神经网络来建立说话人模型，对实际应用环境更加鲁棒，识别率相较于传统的模式匹配算法有很大的提高。

## 1.3 双模态生物特征识别的研究难点

虽然双模态生物特征识别在最近几年中受到研究人员的广泛研究，但是由于实际应用环境的复杂性，双模态生物特征识别技术还没有单模态识别技术成熟。单个模态生物特征提取技术、生物特征融合技术以及识别算法都会影响到双模态生物特征识别的发展。特别是特征提取和特征融合算法上的困难，目前的研究还不够成熟和完善。特征融合方面还没有具有统一的标准化的算法框架和评判标准。概括起来又以下几个方面的困

难，如下图：

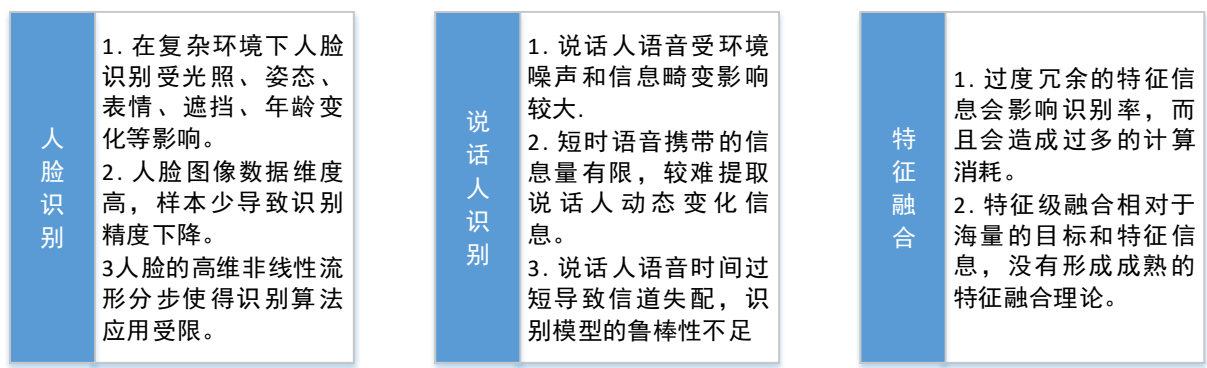

图 1.1 双模态生物特征识别的研究难点

## 1.4 课题主要内容

### 1.4.1 论文主要工作

本文主要研究基于深度学习的双模态生物特征识别，首先对现有的深度学习算法和深度神经网络结构进行了基本的研究，然后研究了目前最流行的基于深度神经网络的人脸识别算法，并且对 Vgg_Face 模型进行了改进，使得人脸特征和语音特征更好的融合识别。根据目前主流的 GMM-UBM 和全局差异性矩阵提取 I-Vector 说话人特征，并且提取说话人的感知线性预测（Perceptual Linear Prediction，PLP）系数。在数据库方面，针对大量的训练数据样本，将 TED-LIUM 语音库和 CASIA-WebFace 人脸库部分数据相结合，制作了一个语音-人脸双模态数据库。最后研究了人脸与说话人特征融合相关算法，并使用 DBN 对融合特征进行训练识别。主要进行了以下几个方面的研究：

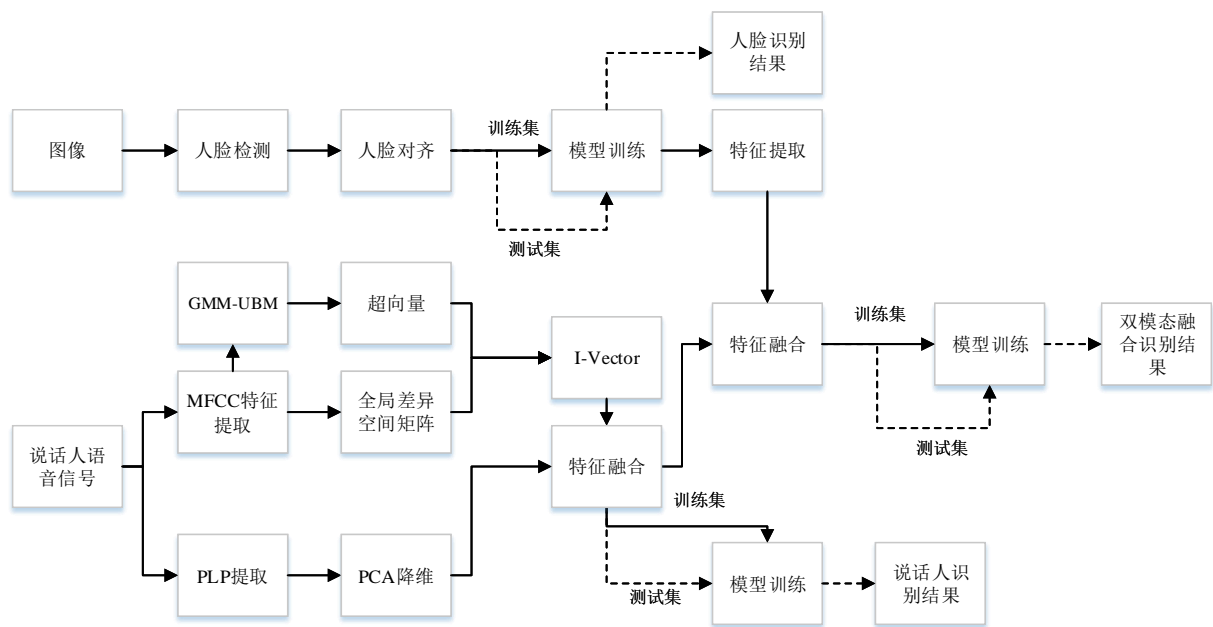

图 1.2 论文工作内容框图

### 1.4.2 论文章节安排

根据课题的研究内容，本文的章节内容安排如下：

在本章首先介绍了课题研究的背景与意义，然后详细的介绍了双模态生物特征识别（人脸识别、说话人识别）的研究历史与现状，以及课题研究的重点和难点。最后介绍了本文的主要工作内容和章节安排。

在本章中首先针对如何提高双模态生物特征识别率，介绍了深度学习的相关理论基础和其在双模态生物特征识别应用中的相关算法理论。然后介绍了经典的深度神经网络的学习算法——误差反向传播（Back Propagation ，BP）算法及其原理。详细推导了深度神经网络中的优化算法和防止过拟合算法。最后介绍了相关深度神经网络相关理论知识，包括深度置信网络（DBN）和卷积神经网络（CNN）。

本章中研究了基于 CNN 的人脸识别算法。首先针对人脸识别中的图像预处理相关算法，包括人脸检测算法和人脸对齐算法。然后研究了人脸识别的深度神经网络结构模型 Vgg_Face。针对双模态识别的需求，对 Vgg_Face 模型的网络结构进行了改进，并且取得了不错的识别效果。

本章中首先研究了说话人 I-Vector 特征的提取方法和 PLP 系数提取算法，包括语音的 MFCC 特征提取、高斯混合模型的参数估计、归一化背景模型的训练以及全局差异空间的估计。并且提出了将 DBN 与融合语音特征相结合，通过新语音特征训练网络模型。并且通过实验给出该网络模型在说话人识别中的识别效果。进一步论证本文提出的语音特征的可靠性。

本章中首先研究了信息融合的相关理论算法。针对人脸和说话人特征各自的特性，创新的提出了一种基于深度神经网络的融合特征识别方法。最后完成融合特征与 DBN 相结合的双模态特征识别实验。

# 第二章 深度学习的基本理论

机器学习是通过多个非线性信息处理层来实现无监督或者是监督的特征提取与转化、模式分类和分析的技术。深度学习是机器学习的一种，是其全新的研究子领域，由Hinton 首先提出。通过模拟人脑的学习机制，使用深度神经网络作为学习的框架，通过有监督或者无监督的学习方式，调用一系列的学习反馈算法，逐步迭代优化网络，最后生成训练模型。通过大量的数据训练出来的模型提取特征具有优秀的泛化性能。其中提取出来的高层特征信息取决于低层特征信息。鉴于目前并行计算技术的快速发展，对数据量和计算速度有很高要求的深度学习广泛应用于计算机视觉、语音等模式识别领域。相较于传统机器学习算法，深度学习大幅提高了识别效果。

## 2.1 深度神经网络理论基础

### 2.1.1 前馈深度神经网络

神经网络伴随着人类对脑部神经生物知识的认知发展形成的。神经网络的基本单元——神经元，最早是由物理学家 Pitts 和心理学家 McCulloch 根据生物神经元传播信息机理提出的[12]。其结构模型示意图如下：

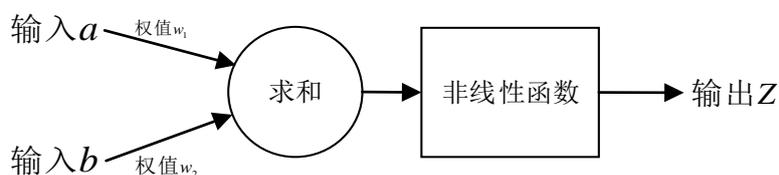

图 2.1 神经元结构模型

从图中可以得到输出 Z 是输入和权值的线性相加，然后通过一个非线性函数 $G$ 得到的。公式为：

$$Z = G(w_1 * a + w_2 * b)$$

2-1

式中 Z 为输出结果，$G$ 为非线性函数，$w_i$ 为权值系数，$a$ 和 $b$ 为输入参数。

每一个神经元包含计算与存储的功能，神经网络的计算就是修改权值的过程。多个神经元在同一个维度可以构成一层神经网络。通过已知的参数 $a$ 和 $b$ 经过神经元的计算得到未知参数 $Z$，这就是神经网络进行特征提取的原理。

最简单的神经网络为单层神经网络结构，它是由输入节点和一层计算节点组成的神经网络结构，其结构模型如下图：

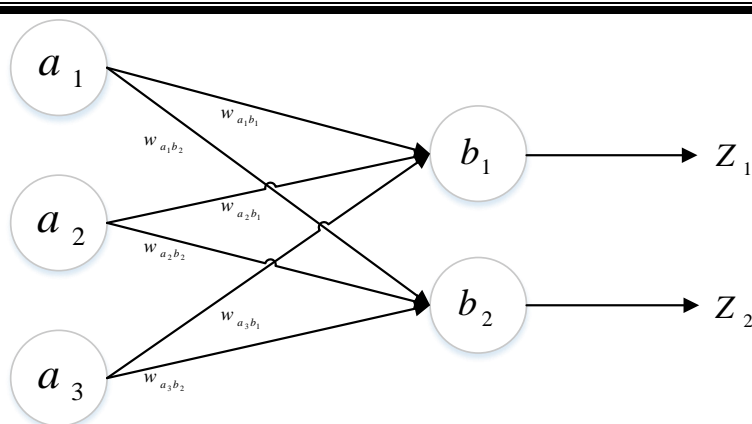

图 2.2 单层神经网络

由上图可以得到单层神经网络的计算公式为：

$$Z_1 = b_1 \left( w_{a_1b_1} * a_1 + w_{a_2b_1} * a_2 + w_{a_3b_1} * a_3 \right)$$
$$Z_2 = b_2 \left( w_{a_1b_2} * a_1 + w_{a_2b_2} * a_2 + w_{a_3b_2} * a_3 \right)$$

2-2

式中 $Z_i$ 为输出结果，$b_i$ 为非线性函数，$w_{ij}$ 为权值系数，$a_i$ 为输入参数。

单层神经网络的权值需要经过训练得到，由公式（2-2）可知其功能类似与逻辑回归模型，可以用于线性分类任务场景。单层神经网络只能完成简单的分类任务，但是可以通过两个神经层组成双层神经网络，并且通过反向传播算法或者其他微分算法来调节层与层之间的权值参数。理论证明，双层神经网络可以无限模拟任何连续函数，可以解决复杂的非线性分类任务[13][14]。双层神经网络结构如下图：

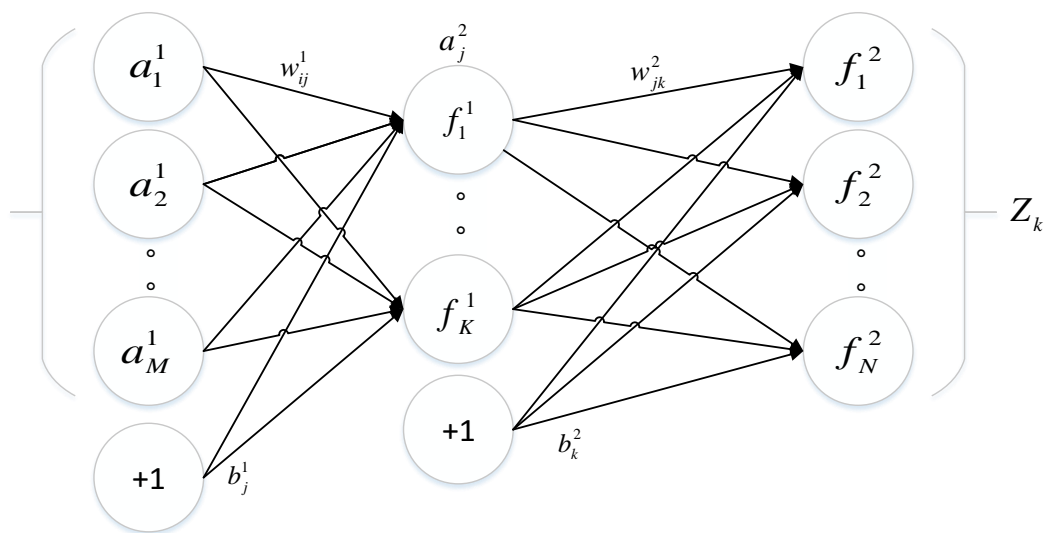

图 2.3 双层神经网络

图中 $a_i^1$ 为输入节点参数，$a_j^2$ 为隐含层节点，$b_i^1$ 和 $b_i^2$ 是偏置参数，$f_i^1$ 和 $f_i^2$ 为每个节点非线性激活函数，$Z_i$ 为输出结果。根据图中网络结构可以得到 $Z_i$ 的计算公式：

$$\begin{cases} a_j^2 = f_j^1\left(w_{ij}^1 * a_i^1 + b_i^1\right) \\ Z_k = f_k^2\left(w_{jk}^2 * a_j^2 + b_j^2\right) \end{cases} \quad 其中 \begin{cases} i = 1, 2 \cdots M \\ j = 1, 2 \cdots \mathrm{K} \\ k = 1, 2 \cdots N \end{cases} \qquad 2\text{-}3$$

在神经网络中非线性激活函数选取 sigmoid 函数或者 tanh 函数。

sigmoid 函数：

$$y = f(x) = \frac{1}{1 + e^{-x}} \qquad 2\text{-}4$$

tanh 函数：

$$y = \tanh(x) = \frac{\sinh(x)}{\cosh(x)} \qquad 2\text{-}5$$

Sigmoid 函数和 tanh 函数性质与神经学中神经元激活特性一致，并且求导容易，是神经网络中常用的激活函数。神经网络在本质上是通过激活函数与参数来拟合目标和其特征之间的函数关系[15]。

双层神经网络可以进行非线性分类其中关键的因素在于双层神经网络包含一个隐含层。根据公式 2-3 可知 $a_i^1$ 通过 $w_{ij}^1$ 进行空间变换得到 $a_j^2$，这使得本不可线性可分的 $a_i^1$ 变换到线性可分空间，然后通过第二层神经网络进行线性分类。双层神经网络的输入输出节点都是确定的，但是隐含层的节点个数可以自由变动，并且隐含层的个数对于神经网络的分类效果有影响，这就为神经网络结构构造和参数调节带来了很多变化。

由单层神经网络到双层神经网络的变化可知，增加一个隐层可以使得神经网络有更多的变化去拟合复杂的函数。函数是目标和特征之间的映射关系，因此更复杂和多样性的特征需要更加复杂的函数去提取，这就需要更多参数的神经网络去拟合[16]。深度神经网络就是基于这个原理提出的。更深的层具有更强的函数拟合能力，可以提取更复杂的特征。前馈深度神经网络结构见下图：

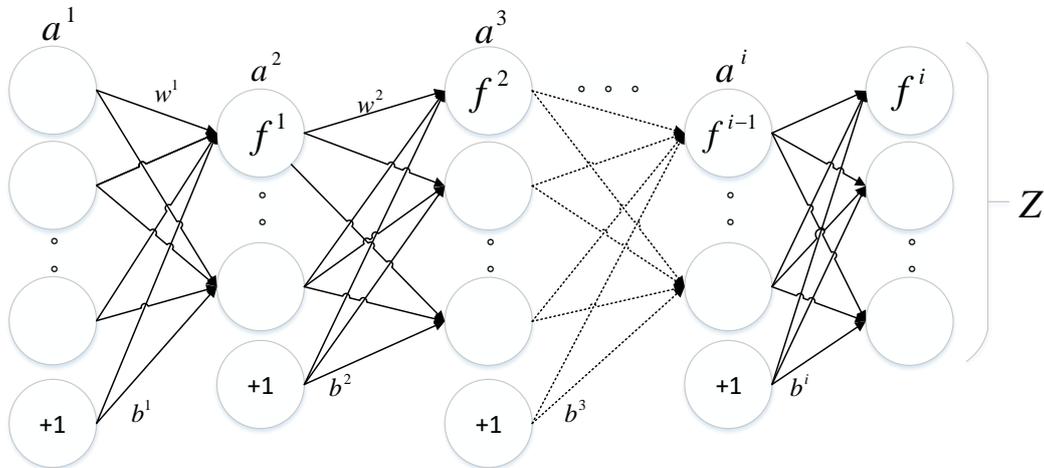

图 2.4 深度神经网络

上图即为前馈深度神经网络结构示意图，在前馈深度神经网络中，自由相邻层节点之间有信息传递，同层节点以及跨层节点之间没有信息传递。与双层神经网络类似，深层神经网络计算公式可以通过式子 2-3 推导出来：

$$\begin{cases} a^2 = f^1\left(w^1 * a^1 + b^1\right) \\ a^3 = f^2\left(w^2 * a^2 + b^2\right) \\ \vdots \\ Z = f^i\left(w^i * a^i + b^i\right) \end{cases} \qquad 2\text{-}6$$

由图 2-4 和式子 2-5 可知，前馈深度神经网络类似于一个有向无环图，并且最后的输出结果是由输入通过复合函数映射得到的。复合函数可以表示为：

$$f(x) = f^i\left(f^{i-1}\left(\cdots f^2\left(f^1(x)\right)\right)\right) \qquad 2\text{-}7$$

$f^1$ 是第一层网络函数，$f^i$ 是最后一层函数，整个网络的深度为 $i$。深度神经网络每一层都是前一层神经元更抽象的表示，最后一层输出的既是整个神经网络学习到的特征。整个网络通过一系列矩阵的线性变化和激活函数进行非线性变换，将输入数据所在的空间投影到线性可分或者稀疏空间中进行分类[17]。

### 2.1.2 反向传播（BP）算法

在实际训练一个深度神经网络模型需要选择优化模型、代价函数和输出单元的方式。其中深度神经网络中各参数需要学习算法进行迭代调节。反向传播（Back Propagation，BP）算法是通过计算结果和预期结果的误差在网络模型中反向传播，来调节参数使得整个神经网络具有一定的学习能力[18][19][20]。其中计算结果和预期结果的误差定义为代价函数。

假设网络输出预期结果为 $Y$，由式子（2-5）中 $Z = f^i\left(w^i * a^i + b^i\right)$ 可以得到关于传播系数 $w$ 的代价函数 $J(w)$：

$$J(w) = G(Z, Y) \qquad 2\text{-}8$$

其中函数 $G$ 为代价函数定义类型。那么可以得到系数 的修正式为：

$$w_j^i := w_j^i - \alpha \frac{\partial}{\partial w_j^i} J(w) \qquad 2\text{-}9$$

深层神经网络的代价函数使用的是预测值和实际值的交叉熵，使得网络收敛的更快。交叉熵代价函数如下：

$$J(w) = -\frac{1}{n}\sum_{k=1}^{n}\left(Y_k \ln Z_k + (1-Y_k)\ln(1-Z_k)\right) \qquad 2\text{-}10$$

其中 $k$ 为最后一层节点系数。浅层神经网络的代价函数常用的是最小二乘法，即：

$$J(w) = \frac{1}{2}\sum_{k=1}^{n}(Z_k - Y_k) = \frac{1}{2}\sum_{k=1}^{n}\left(f_k^i\left(w_k^i * a_k^i + b_k^i\right) - Y_k\right)^2 \qquad 2\text{-}11$$

对 $J(w)$ 对 $w$ 求偏倒可以得到：

$$
\begin{aligned}
\frac{\partial}{\partial w_j} J(w) &= \frac{\partial}{\partial w_j} \frac{1}{2}\sum_{k=1}^{n}\left(f_k^i\left(w_k^i * a_k^i + b_k^i\right) - Y_k\right)^2 \\
&= 2 \cdot \frac{1}{2}\left(f_j^i\left(w_j^i * a_j^i + b_j^i\right) - Y_j\right) \cdot \frac{\partial}{\partial w_j}\left(f_j^i\left(w_j^i * a_j^i + b_j^i\right) - Y_k\right) \\
&= \left(f_j^i\left(w_j^i * a_j^i + b_j^i\right) - Y_j\right) \cdot a_j^i \cdot f_j^{i'}
\end{aligned}
\qquad 2\text{-}12
$$

其中 $\alpha$ 为学习率。由式 2-9 和 2-12 可知 $w$ 的修正和实际值与预测值的差还有激活函数 $f^i$ 的梯度有关。

在整个网络参数调节的过程中，希望是实际值和预测值的误差越大，在反向传播的过程中参数调节的幅度越大，整个网络收敛的越快。但是激活函数 $f^i$ 常使用的形式类似于 sigmoid 函数，如下图：

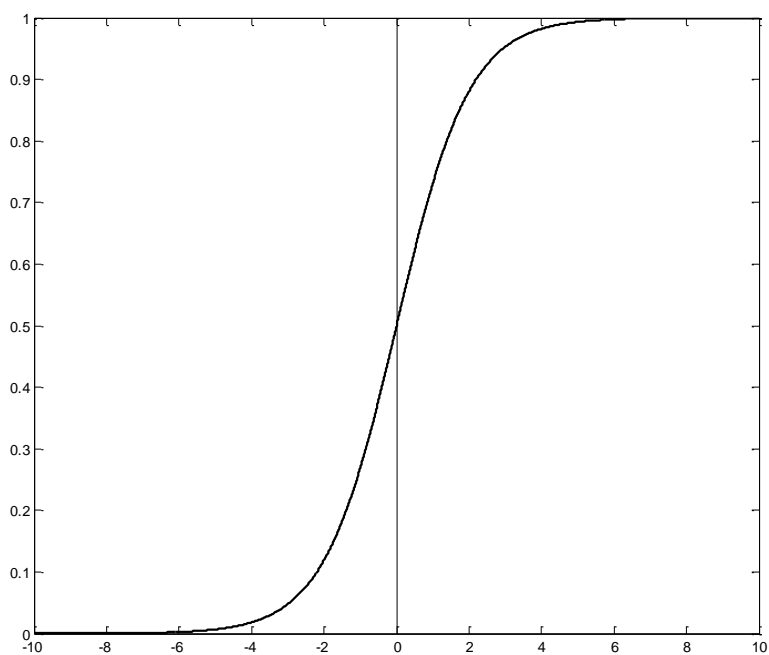

图 2.5 Sigmoid 函数示意图

可以从图中得到，在误差较大的地方，$f^i$ 的梯度下降非常慢，这使得代价函数随着误差增大收敛越慢[21]。特别是在深度神经网络的收敛运算中，这样的收敛特性很难运用到实际当中。接下推倒交叉熵代价函数关于 $w$ 的修正公式：

$$\frac{\partial}{\partial w_j} J(w) = \frac{\partial}{\partial w_j} \left( -\frac{1}{n} \sum_{i=1}^{n} \left( Y_i \ln Z_i + (1 - Y_i) \ln (1 - Z_i) \right) \right)$$

$$= -\frac{1}{n} \left( \frac{Y_j}{f_j^i \left( w_j^i * a_j^i + b_j^i \right)} - \frac{1 - Y_j}{1 - f_j^i \left( w_j^i * a_j^i + b_j^i \right)} \right) \cdot f_j^{i'} \left( w_j^i * a_j^i + b_j^i \right) \cdot a_j^i$$

$$= -\frac{1}{n} \frac{f_j^{i'} \left( w_j^i * a_j^i + b_j^i \right) \cdot a_j^i}{f_j^i \left( w_j^i * a_j^i + b_j^i \right) \cdot \left( 1 - f_j^i \left( w_j^i * a_j^i + b_j^i \right) \right)} \left( f_j^i \left( w_j^i * a_j^i + b_j^i \right) - Y_j \right)$$

2-13

其中 $f^i$ 激活函数为 sigmoid 函数，对其求导可得：

$$f_j^{i'}(x) = \left( \frac{1}{1 + e^{-x}} \right)^{'}$$

$$= -\left( 1 + e^{-x} \right) \cdot \left( e^{-x} \right)^{'}$$

$$= \frac{1}{\left( 1 + e^{-x} \right)^2} \cdot \left( e^{-x} \right)$$

$$= \frac{1}{1 + e^{-x}} \cdot \left( 1 - \frac{1}{1 + e^{-x}} \right)$$

$$= f_j^i(x) \left( 1 - f_j^i(x) \right)$$

2-14

将式子（2-14）带入（2-13）可得：

$$\frac{\partial}{\partial w_j} J(w) = \frac{1}{n} a_j^i \cdot \left( f_j^i \left( w_j^i * a_j^i + b_j^i \right) - Y_j \right)$$

2-15

从上式可得激活函数的梯度可被消除，参数修正只与预测值和实际值的误差有关，这使得误差越大，参数修正的越快。

从上面的推倒可见神经网络每层参数的收敛与代价函数的梯度有关，假设要调节第 $m$ 层参数，那么根据式子（2-9）可知：

$$w_j^m = w_j^{m+1} - \alpha \frac{\partial}{\partial w_j^{m+1}} J_j(w)$$

2-16

根据微积分中的求偏倒链式法则，可以求得任何层参数调节公式为：

$$w_j^m = w_j^{m+1} - \alpha \frac{\partial}{\partial w_j^{m+1}} J_j(w)$$

$$= w_j^{m+1} - \alpha \frac{\partial}{\partial w_j^{m-1}} J_j(w) \cdot \frac{\partial}{\partial w_j^{m-2}} w_j^{m-1} \cdots \frac{\partial}{\partial w_j^{m+1}} w_j^{m+2}$$

2-17

通过式子（2-17）即可通过预测值和实际值的代价函数调节任意层参数，从最后一层调节到第一层，然后再进行一次前馈计算得到新的预测值。反复迭代直到预测值和实际值的误差小于设定的阈值[22]。

反向传播算法不是计算神经网络梯度的唯一算法或是最优算法，但是它是一个非常

实用的算法，在目前的很多实际应用的深度神经网络模型中都使用它作为网络学习算法。

## 2.2 深度学习优化算法

在深度神经网络的算法中都涉及到优化问题。神经网络的参数优化过程也是一个学习的过程。在实际问题建模中，直接符合凸优化条件的函数非常少。特别是在深度神经网络的多维特征输入和多隐含层的情况下，针对非凸优化条件的函数的优化算法成为研究深度神经网络的关键部分[23][24]。优化指的是改变自变量 $x$ 来最大化或者最小化函数 $f(x)$。通常以最小化 $f(x)$ 来表示最优化问题，最大化的情况可以用最小化 $-f(x)$ 来实现。

在传统的机器学习中，常用的优化算法从大方向上可以归纳为一阶优化算法和二阶优化算法。其中一阶优化算法包含随机梯度下降法（Stochastic gradient descent,SGD），自适应梯度下降法（Adaptive gradient, AdaGrad），自适应学习率法（Adaptive Delta, AdaDelta）等。二阶优化算法包含牛顿法（Newton's method），共轭梯度法（Conjugate gradient），L-BFGS 等[25][26]。

一阶优化算法需要设定学习率，通过大量的迭代训练才能找到满足损失函数的最优解。二阶优化算法可以通过少量的迭代训练可完成寻优。但是对于高维特征而言，二阶优化算法每次的迭代代价要比一阶优化算法计算代价大得多[27]。对于人脸和语音等高维特征数据来说，实际中使用一阶优化算法要比二阶优化算法更具有优势。

### 2.2.1 随机梯度下降法

对于一个简单的一维函数 $y = f(x)$，如图 2-6。其导数为 $f'(x)$，表示函数在点 $x$ 的斜率。由此可以的到在函数上输入较小的自变量变化 $\alpha$ 可以在输出上得到相应的变化：$f(x+\alpha) \approx f(x) + \alpha f'(x)$。因此导数对于优化函数有很大的作用，其可以给出如何改变自变量可以最小化函数值。假如已知自变量微小变化 $\varepsilon$，得到 $f(x - \varepsilon|f'(x)|)$ 比 $f(x)$ 小，则可以反向移动自变量来减小输出，这种方法称为梯度下降（gradient descent）[28]。

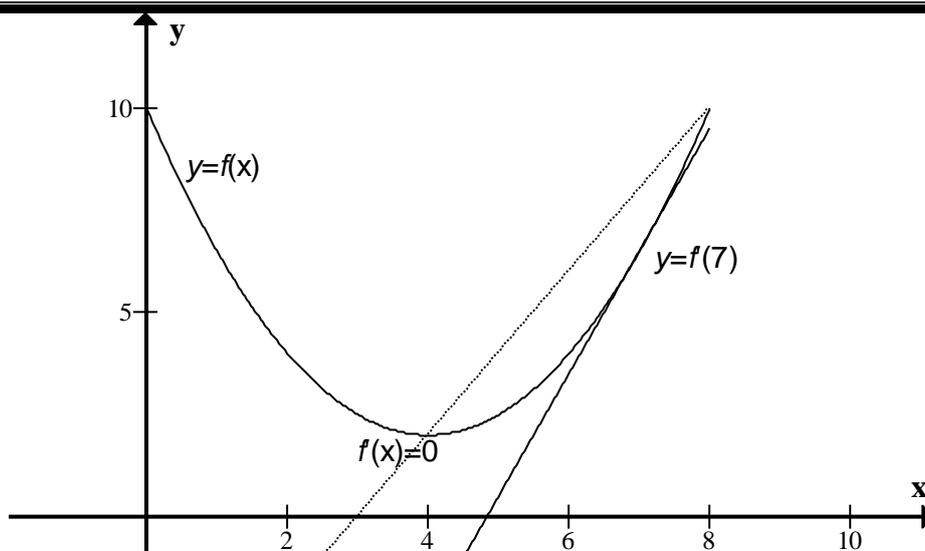

图2.6 一维凸函数梯度下降示意图

根据上图可知，当函数导数为正的时候，函数自变量往负向移动；当导数为负时，自变量向正向移动，即可得到函数的最小值。当导数 $f'(x)=0$，则无法提供自变量变化的方向，这时称这个点为临界点。对于凸优化条件函数的最小值在其局部最小值得地方。当待优化函数是一个复杂的非凸条件函数时，见下图。局部最小值不一定是全局最小值。

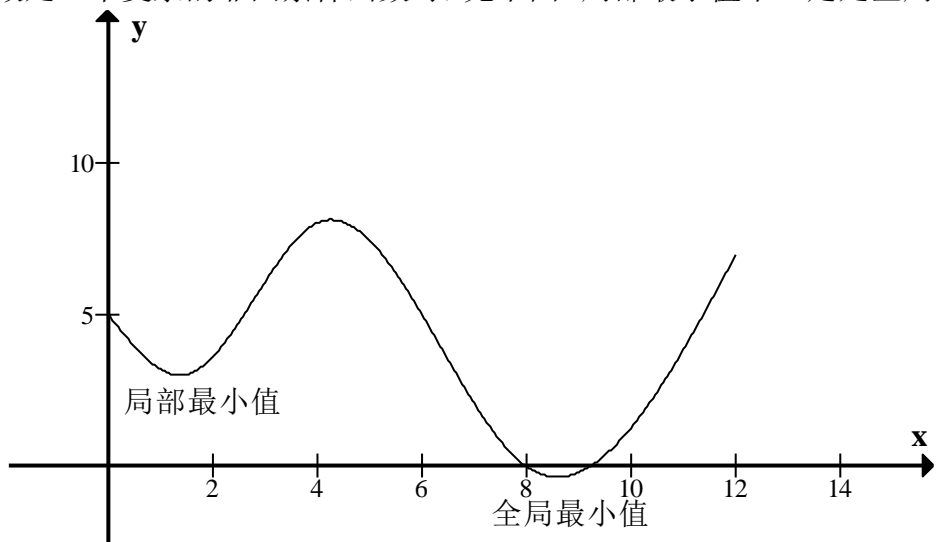

图2.7 复杂非凸优化函数示意图

在深度学习中，优化函数得到的不一定是最优的局部最小值。在寻找函数 $f(x)$ 最小值的过程中，不一定需要得到全局最优值。优化函数得到的最优解一定要满足代价函数显著低的值才可以接受。由于输入多维特征的时候，这些使得优化问题变得非常难[29]。针对多维特征输入函数的最优化，提出了偏导数 $\frac{\partial}{\partial x_j}f(x_j)$ 的概念。对于一个梯度而言是求一个向量的导数，而函数 $f(x)$ 的导数是包含所有偏导的向量，梯度记为 $\nabla_x f(x)$。梯度第 $j$ 个元素是函数 $f(x)$ 关于 $x_j$ 的偏导数。根据上面的一维情况可知，多维条件下，临界点即为梯度所有元素为0的点。

定义一个单位向量 $v$，那么函数 $f(x)$ 在其方向的方向导数即为此函数在 $v$ 方向上的斜率。假设存在一个变量 $\beta$，函数 $f(x+\beta v)$ 关于 $\beta$ 的导数（$\beta=0$ 时）即为在单位向量 $v$ 的方向导数：

$$\frac{\partial}{\partial \beta} f(x+\beta v) = v^T \nabla_x f(x) \qquad 2\text{-}18$$

在最小化函数 $f(x)$ 的过程中，希望得到 $f(x)$ 下降的最快以减少训练时间。在函数下降最快的方向计算其方向导数：

$$\min_{v,v^T v=1} v^T \nabla f(x)$$
$$= \min_{v,v^T v=1} \|v\|_2 \|\nabla f(x)\|_2 \cos\theta \qquad 2\text{-}19$$

其中梯度和向量 $v$ 的夹角为 $\theta$。忽略和单位向量无关联项，并且 $\|v\|_2 = 1$。可以简化 2-19 为 $\min_v \cos\theta$。在单位向量 $v$ 和梯度方向相反的时候可以得到最小值。也就是负梯度向量上移动自变量可以减小函数 $f(x)$。这种方法是多维输入函数的梯度下降法。其中自变量更新函数为：

$$x^{'} = x - \varepsilon \nabla_x f(x) \qquad 2\text{-}20$$

其中 $\varepsilon$ 是学习率（learning rate），学习率是自变量迭代更新的权值。学习率根据训练数据集和深度神经网络的结构所设定[30]。若设定的学习率为一个固定的值，在函数最小化的前期收敛相对较快，但是由于数值固定，很容易造成函数收敛后期自变量变化过大使得函数梯度发散，从而无法达到最优值。因此目前使用梯度下降法的学习率随着训练循环次数的增加而逐渐减小。在训练的后期依然能够很好地逐步地收敛到最优解。

在深度神经网路中，梯度下降法面对大数据集时通常很慢而且十分不可靠。由于训练函数的非凸性质，梯度下降在这种情况下可能不能保证在接受的时间内达到代价函数的最小值。

随机梯度下降法（stochastic gradient descent,SGD）是梯度下降法针对大数据集的一个优化算法[31]。在训练过程中，整体的损失函数可以分解到每个样本的损失函数。例如损失函数可以表示成：

$$J(\theta) = E_{x,y \sim \hat{p}_{data}} L(x,y,\theta) = \frac{1}{m} \sum_{i=1}^{m} L(x^i, y^i, \theta) \qquad 2\text{-}21$$

上式对训练样本数量的计算代价为 $O(m)$，当训练集中样本有 10 万个，这时计算一次梯度就需要消耗很长的时间。这里提出 Minibatch 的概念，一个 Minibatch 是在样本集合里面随机抽取一定数值的样本[32]。这个数值相对于远样本集合来说非常小。假定在 10 万个样本里面随机抽取 200 个样本为一个 Minibatch，一共有 500 个 Minibatch。每次

一个训练迭代循环在一个 Minibatch 上完成。并计算其梯度：

$$g = \frac{1}{m'} \nabla_\theta \sum_{i=1}^{m'} L\left(x^i, y^i, \theta\right) \qquad 2\text{-}22$$

其中 $m'$ 为 Minibatch 的样本数量。训练中用该梯度来完成自变量的更新 $\theta - \varepsilon g \to \theta$。直到整个集合样本都完成对参数的迭代优化或者函数的最优解达到损失函数能够接受的阈值。因此在训练样本无限大时，随机梯度下降法不需要对所有样本进行训练。只要模型收敛到设定的最优测试误差范围内就可以停止训练。由此可知 SGD 法关于训练样本数量级的计算代价为 $O(1)$，可以较快的找到代价函数能够接受的一个非常接近最小值的解。

### 2.2.2 常用自适应学习率优化算法

优化算法的学习率设定对模型训练的性能有很大的影响。如果系统能够自适应调节学习率将会更加快速和精确的调节参数，并且可以防止函数优化到鞍部。上小节介绍的随机梯度下降法的学习率是人为设置的，若设置较大会造成迭代到一定程度后梯度的发散；若设置较小，模型的训练又会变得异常缓慢。研究学者提出了基于梯度下降，适应于模型参数的学习率的优化算法：

自适应梯度下降算法：

AdaGrad 算法能够独立地自适应于所有模型[33]。其将迭代速率反比于所有历史梯度总和的平方根。这种方法产生的更新步长使得具有较大偏导的参数下降较为缓慢，而对于具有较小偏导的参数有相对快速下降。总体效果会使得参数的优化在平缓梯度处获得较大步进，在梯度较陡处获得较小的步进长度。

假设此次迭代之前的所有更新消息为：

$$\left(\nabla L(W)\right)_{t'} \quad t' \in \{1, 2, \cdots, t\} \qquad 2\text{-}23$$

那么每一个自变量的更新公式为：

$$\left(W_{t+1}\right)_i = \left(W_t\right)_i - \varepsilon \frac{\left(\nabla L(W_t)\right)_i}{\sqrt{\sum_{t'=1}^{t} \left(\nabla L(W_{t'})\right)_i^2 + \delta}} \qquad 2\text{-}24$$

其中 $\delta$ 为 $10^{-7}$ 级的常量防止分母为零，$\varepsilon$ 为全局学习率，$W$ 为参数权重。

这种算法需要划分一部分内存出来存放过往的梯度信息。在凸条件函数优化问题上能够很快地收敛到最优值。但是随着海量的数据输入和内存的限制，载入所有的历史梯度信息将变得越来越困难。到训练的中后期更新公式的分母趋近于 0，将会导致训练提前结束。

RMSProp 算法：

为了克服 AdaGrad 算法在海量数据输入和非凸函数优化问题上存在缺陷，Hinton 在 2012 年提出了 RMSProp 算法[35]。将 AdaGrad 算法中的梯度累加改为指数加权。因为 AdaGrad 算法在非凸函数优化过程中会通过很多局部最小值，最终达到一个最后的局部最小值点。但是其累积了之前的所有梯度信息，导致到达准确的最小值的凸结构时，学习速率变得非常慢。所以 RMSProp 算法的基于指数衰减的方法可以丢弃时间比较久远的历史梯度信息。这样在找到准确的凸结构时还能快速收敛到最优解。

设定一个衰减系数 $\rho$ 和全局学习率 $\varepsilon$。$t$ 时刻之前的梯度总和为：

$$\sum_{t'=1}^{t}\left(\nabla L\left(W_{t'}\right)\right)_{i}^{2} \quad t' \in \left\{1, 2, \cdots, t\text{-}1\right\} \qquad 2\text{-}25$$

那么 $t$ 时刻的梯度的指数加权均方和为：

$$\gamma = \rho \sum_{t'=1}^{t}\left(\nabla L\left(W_{t'}\right)\right)_{i}^{2} + (1-\rho)\left(\nabla L\left(W_{t}\right)\right)_{i}^{2} \quad t' \in \left\{1, 2, \cdots, t\text{-}1\right\} \qquad 2\text{-}26$$

根据上式可以得到参数更新公式为：

$$\left(W_{t+1}\right)_{i} = \left(W_{t}\right)_{i} - \varepsilon \frac{\left(\nabla L\left(W_{t}\right)\right)_{i}}{\sqrt{\gamma + \delta}} \qquad 2\text{-}27$$

根据式 2-26 可知每循环一次，历史梯度和需要乘上衰减系数 $\rho$。当训练到一定时间，间隔久远的梯度历史信息将会被丢弃。RMSProp 算法依然需要提供全局学习率，若设置的较高，会使得收敛过于敏感，调节变化太大。

## 2.3 深度神经网络的正则化

深度学习的重点不是无限求于训练数据的识别率高，而是识别模型能够在训练集合外的数据上具有优秀的泛化能力。因此在深度学习中有必要增加训练误差为代价防止过拟合来减少测试误差，这种方式称为正则化[36]。正则化不是为了对训练误差的修改而是为了减少优化算法的泛化误差。大多数正则化方法都是对参数估计量进行正则化，以增加参数估计量的偏差来减少方差的减少。在实际应用中最好的拟合模型（具有最小的泛化误差）具有适当的正则化。下面介绍深度学习中常用的几个正则化方法。

### 2.3.1 参数范数惩罚项

假设目标函数为 $J$，对其正则化的方式添加一个参数范数惩罚项 $\Omega(\beta)$，来限制模型对训练集合的学习。那么添加正则项后的目标函数为：

$$J\left(x, y; \beta\right) = J\left(x, y; \beta\right) + \varphi \Omega(\beta) \quad \varphi \in \left[0, +\infty\right) \qquad 2\text{-}28$$

若系数 $\varphi = 0$ 表示未添加正则项，系数越大表示惩罚程度越高。

在深度神经网络收敛的时候，由于非线性函数的非凸性，会有局部最优的情况。深度神经网络模型在过拟合的情况下，为了使每个估计点都在预测结果中，会导致最终的

拟合函数波动较大。导致在某些小区间内，函数值变化剧烈。为了防止训练模型过拟合，通过增加参数范数惩罚项[37]，来减少修正步长[37]。选取不同的参数惩罚项也会对目标函数正则化有不同的影响。每种正则项包含的固有数学特性，都有其应用的场景。

### 2.3.1.1 $L^2$ 范数正则化

使用 $L^2$ 参数范数惩罚的正则化方法是通过给目标函数添加正则项 $\Omega(\beta)=0.5\cdot\|w\|_2^2$ 使得参数权重更加靠近原点。

假定没有设定偏置项，那么权重 $\beta=w$，可以将模型的目标函数表示成：

$$\tilde{J}(x,y;\beta)=\tilde{J}(x,y;w)=J(x,y;w)+\frac{\varphi}{2}w^Tw \qquad 2\text{-}29$$

那么目标函数的梯度为：

$$\nabla_w\tilde{J}(x,y;w)=\nabla_wJ(x,y;w)+\varphi w \qquad 2\text{-}30$$

将上式带入参数更新公式中：

$$\begin{aligned} &w-\varepsilon\left(\nabla_w\tilde{J}(x,y;w)\right)\\ &=w-\varepsilon\left(\nabla_wJ(x,y;w)+\varphi w\right)\to w \end{aligned} \qquad 2\text{-}31$$

换种表达方式可以得到更新公式为：

$$w\leftarrow(1-\varepsilon\varphi)w-\varepsilon\nabla_wJ(x,y;w) \qquad 2\text{-}32$$

可以从上式得到，参数权重每一步更新都会带有权重衰减。结合式 2-11 可以推导出权重衰减会对目标函数的训练带来好处。代价函数设为均方误差和：

$$(xw-y)^T(xw-y) \qquad 2\text{-}33$$

添加正则项之后代价函数为：

$$(xw-y)^T(xw-y)+\frac{\varphi}{2}w^Tw \qquad 2\text{-}34$$

参数权重 $w$ 计算公式将会更新为：

$$w=(x^Tx+\varphi I)^{-1}x^Ty \qquad 2\text{-}35$$

从上式可知 $L^2$ 范数让模型的训练能够对高方差的输入 $x$ 输出协方差较小的特征权重。虽然会牺牲训练数据的准确率，但是提高了测试数据集的泛化能力[38]。

### 2.3.1.2 $L^1$ 范数正则化

类似于 $L^2$ 范数，$L^1$ 正则化表示形式为：

$$\Omega(\beta)=\|w\|_1=\sum_i|w_i| \qquad 2\text{-}36$$

惩罚项为参数权重的绝对值总和。那么可以表示出正则化后目标函数：

$$\tilde{J}(x,y;w) = J(x,y;w) + \frac{\varphi}{2}\|w\|_1 \qquad\qquad 2\text{-}37$$

相应的梯度为：

$$\nabla_w \tilde{J}(x,y;w) = \nabla_w J(x,y;w) + \varphi\,\mathrm{sign}(w) \qquad\qquad 2\text{-}38$$

因此正则化后的目标函数受到参数 $w$ 的影响。根据 $L^1$ 范数的特性可知当系数 $w$ 为负值的时候，更新之后，系数会变大；当系数 $w$ 为正值的时候，更新之后，系数会变小。这样可以使得整个网络的权重系数接近于 0，减小网络复杂度，防止过拟合。相比于 $L^2$ 范数，$L^1$ 正则化会产生更加稀疏解[39][40]。因此 $L^1$ 正则化的稀疏特性较为广泛的应用于特征选择场景中。从特征子集中选取更优的子集，从而简化训练问题。

## 2.3.2 Dropout 正则化

在介绍 Dropout 正则化方法前先介绍 Bagging（Bootstrap Aggregating）正则化方法[41]。在机器学习模型训练中，Bremen 在 1994 年提出 Bagging 算法，通过结合数个模型来降低泛化误差。最后通过所有模型来表决最后输出值，此方法又被称为模型平均法。在深度神经网络中，由于深度神经网络的训练一个模型就需要消耗大量的内存和时间，通过结合数个模型来降低泛化误差的方法不现实。但是可以将 Bagging 正则化方法的思想应用到 Dropout 正则化中[42][43]。

Dropout 算法在深度神经网络训练过程中近似 Bagging，但是不是训练多个完整的深度神经网络模型。Dropout 算法是将深度神经网络模型的训练拆分成多时段，多片段的模式。具体来说就是 Dropout 是训练基本网络减去非输出单元形成的子网络[44]。如图 2-8。在模型训练的时候随机选取某一些隐含层的参数节点不工作。剩下的网络作为完整的神经网络进行更新。被暂停的节点的权值保留下来，等下次没有被选取成禁用的时候继续参与迭代更新。这种技术减少了神经元之间复杂的共适性。因为一个神经元不能依赖其他特定的神经元。因此，不得不去学习随机子集神经元间的鲁棒性的有用连接。神经元作为要给预测的模型，Dropout 可以确保模型在丢失某些体线索来减少权重连接，然后增加模型在缺失个体连接信息情况下的鲁棒性。

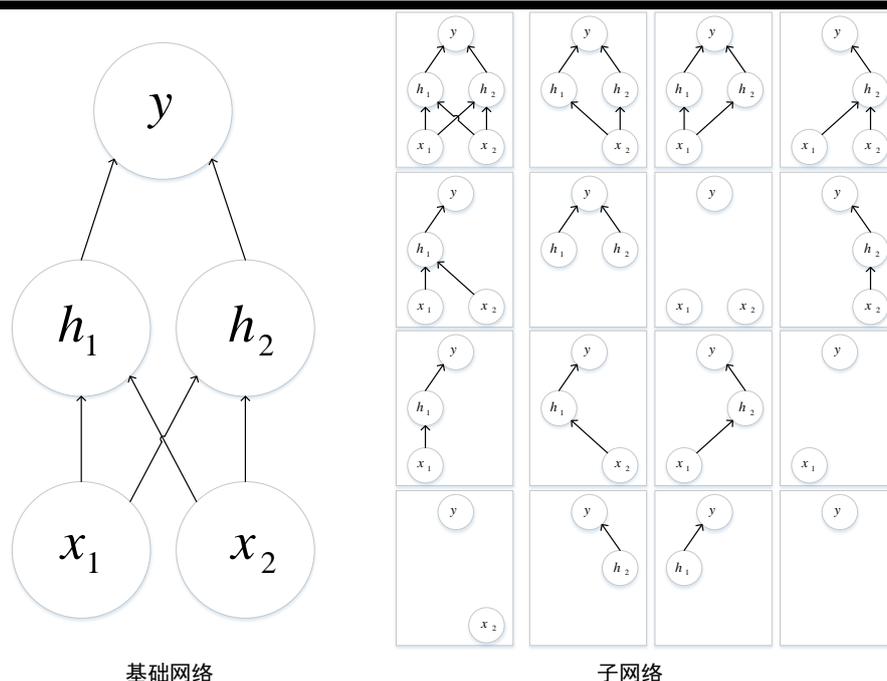

基础网络　　　　　　　子网络

图 2.8 Dropout 正则化原理示意图

在使用 Dropout 正则化时，考虑到之前的 Minibatch 样本集。每次加载 Minibatch 中的一个样本神经网络就会随机选取网络中输入和隐藏单元的掩码。这些独立采样的掩码可以开闭相应的响应单元。这样抽样后的神经网络在继续之前的训练模式进行向前传播。

假定相对于每个掩码 $\eta$，其子模型的概率分布函数为 $p(y|x,\mu)$。关于所有的掩码的算术平均值可以表示为：

$$\sum_u p(\eta)p(y|x,\eta) \qquad\qquad 2\text{-}39$$

其中 $p(\eta)$ 是训练模型采用掩码 $\eta$ 的概率。由于深度神经网络包含的网络模型太复杂。如果遍历每种可能的子网络，会造成巨大的计算量，不够现实。实际应用中一般取 10~15 个掩码足够有很好的表现。Dropout 相比于其它的计算开销较小的正则化项（权重衰减，稀疏激活）等正则化更有效。Dropout 正则化方法还可以和其他形式的正则化方式（如 $L^1$、$L^2$ 范数等）合并，进一步提升正则化性能。

Dropout 正则化不限制于模型或者训练过程，能够广泛使用在 SGD 或其他优化算法模型上。其在模型的某一个迭代过程中效果不是很明显，但在完整的训练过程中有很好地泛化能力[45]。Dropout 的缺点是在少数量样本集中不能得到好的效果。

## 2.4 深度神经网络介绍

### 2.4.1 深度置信网络

深度置信网络（Deep belief network, DBN）是由 Hinton 在 2006 年提出来的，这是

第一个应用于深度框架的非卷积神经网络模型。在 MNIST 手写数字数据集上超过了凸目标函数内核 SVM。在当时具有相当重要的意义，并由此掀起了新一轮的深度学习的研究热潮。

DBN 是有多个受限玻尔兹曼机和一个分类器层所构成。模型的训练分为无监督预训练与有监督微调两个步骤。下面对 DBN 的网络重要组成部分——受限玻尔兹曼机（Restricted Boltzmann Machine, RBM）和其训练方式进行研究。

### 2.4.1.1 受限玻尔兹曼机

受限玻尔兹曼机是特殊的马尔科夫随机场。RBM 是由一个隐含层和一个可见单元层构成。这两个层之间的单元全部两两连接，但是层内单元之间互不连接。也就是说 RBM 是一个层内无连接，层间全连接的双向图结构。如下图。

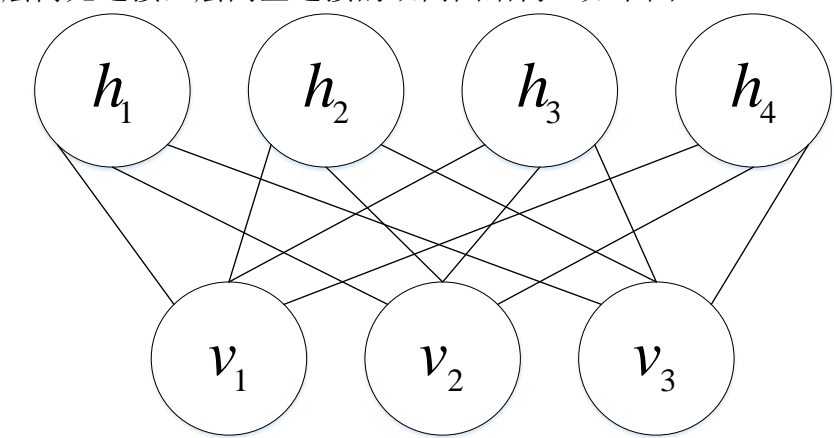

图 2.9 RBM 结构示意图

上图中 $v$ 是可视层单元，$h$ 是隐含层单元。假定 RBM 的单元之间的相互联系可以用能量的方式来表示。设定参数 $b$，$c$ 和 $W$ 是可以学习到的。为了方便公式的简化定义 $\beta = \{b, c, W\}$。那么能量函数可以表示成：

$$E(v, h, \beta) = -b^T v - c^T h - v^T W h \qquad 2\text{-}410$$

那么隐含层和可视层单元之间的条件概率公式为：

$$p(h|v) = \prod_j p(h_j|v)$$
$$p(v|h) = \prod_i p(v_i|h) \qquad 2\text{-}41$$

根据独立条件分布对上式进行的因式分解可以得到：

$$\begin{cases} p\left(h_j = 1 \mid v\right) = \sigma\left(v^T W_{:,j} + c_j\right) \\ p\left(h_j = 0 \mid v\right) = 1 - \sigma\left(v^T W_{:,j} + c_j\right) \\ p\left(v_i = 1 \mid h\right) = \sigma\left(h^T W_{:,i} + b_i\right) \\ p\left(v_i = 0 \mid h\right) = 1 - \sigma\left(h^T W_{:,i} + b_i\right) \end{cases} \qquad 2\text{-}42$$

其中 $\sigma(x) = 1/(1 + \exp(-x))$。对于可视层节点 $v_i$ 来说，可以用最大似然估计来计算参数 $\beta$：

$$\hat{\beta} = \arg\max_\beta \sum_{i=1}^N \log p(v_i \mid \beta) \qquad 2\text{-}43$$

将 $\hat{\beta}$ 展开：

$$\begin{aligned} L(\beta) &= \sum_{i=1}^N \log p(v_i \mid \beta) \\ &= \sum_{i=1}^N \log \sum_h p(v_i, h \mid \beta) \\ &= \sum_{i=1}^N (\log \sum_h \exp[-E(v_i, h \mid \beta)] - \log \sum_v \sum_h \exp[-E(v, h \mid \beta)]) \end{aligned} \qquad 2\text{-}44$$

然后对进行求梯度：

$$\frac{\partial L(\beta)}{\partial \beta} = \sum_{i=1}^N \left( \left\langle \frac{\partial(-E(v_i, h \mid \beta))}{\partial \beta} \right\rangle_{p(h \mid v_i, \beta)} - \left\langle \frac{\partial(-E(v, h \mid \beta))}{\partial \beta} \right\rangle_{p(v, h \mid \beta)} \right) \qquad 2\text{-}45$$

根据上式提取出 $\beta$ 中的 $W$，可以得到 $W$ 的导数为：

$$\frac{\partial L(\beta)}{\partial W_{ij}} = \sum_h p(h \mid v) v_i h_j - \sum_{v,h} p(v, h) v_i h_j \qquad 2\text{-}46$$

可以得到：

$$\Delta W_{ij} = E_1(v_i h_j) - E_2(v_i h_j) \qquad 2\text{-}47$$

式中

同理可得：

$$\begin{aligned} \Delta a_i &= E_1(v_i) - E_2(v_i) \\ \Delta b_j &= E_1(h_j) - E_2(h_j) \end{aligned} \qquad 2\text{-}48$$

其中 $E_1(x)$ 是训练数据集中样本数据的观测值期望。$E_2(x)$ 是模型计算所得分布上的期望。根据式 2-42 可以很容易求得 $E_1(x)$。但是 $E_2(x)$ 需要知道全部样本的分布才能得到。对比散度近似计算梯度可以近似得到期望值，由此可以通过训练样本初始化之后的吉布斯采样来近似代替 $E_2(x)$。

### 2.4.1.2 DBN 的训练方式

将多个 RBM 叠加在一起，然后增加一个分类器层即可构成 DBN-DNN，如图 2-10。

无监督预训练的过程是，从网络的底部由下而上进行预训练。首先使用随机值预训练最底层的高斯-伯努利或者伯努利-伯努利 RBM，将其隐含层的节点激活概率作为下一个 RBM 的输入。然后计算第二个 RBM 的隐含层节点激活概率输入到第三个 RBM 中，依次类推，直到分类器层得到最后结果。根据提出的逐层贪婪学习理论可以达到整个网络结构的最大似然学习。此过程是无监督的学习，不用提供任何标签。

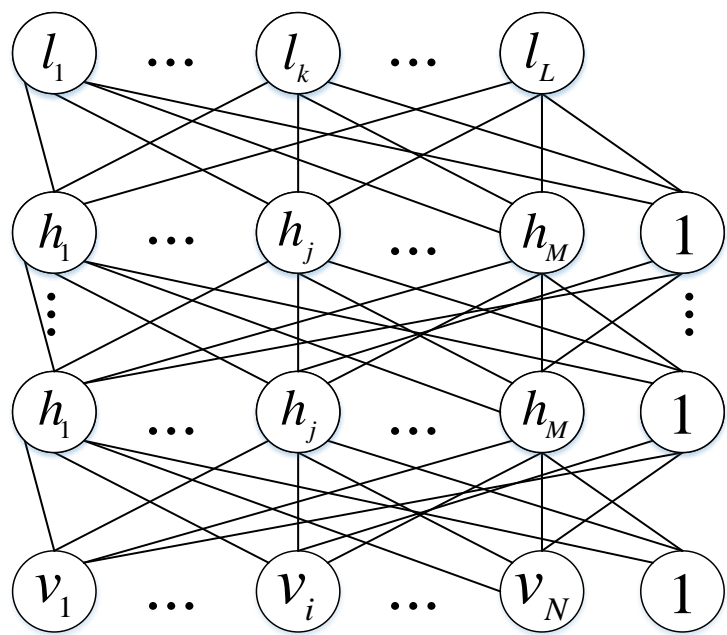

图 2.10 DBN 结构示意图

在提供训练样本数据后，通过样本数据信息和标签信息，使用反向传播算法来精调整个网络的参数权值。DBN 可以学习到样本之间深度的复杂特性，可以被用于人脸识别，语音识别等识别领域。

### 2.4.2 卷积神经网络

卷积神经网络（Convolutional Neural Network，CNN）通常用来处理具有网络结构特性的数据。例如语音（在时间序列上具有关联特性）和图像（在二维空间类具有关联特性）数据，可以通过神经网络进行处理。卷积神经网络的关键点在于网络中至少有一个层使用了卷积运算，其中卷积是特殊的线性运算。数据经过卷积运算之后，会生成一组线性激活函数，之后要通过非线性激活函数对前一层的进行调整。然后还要使用池化函数对卷积层的输出进行调整，如下图：

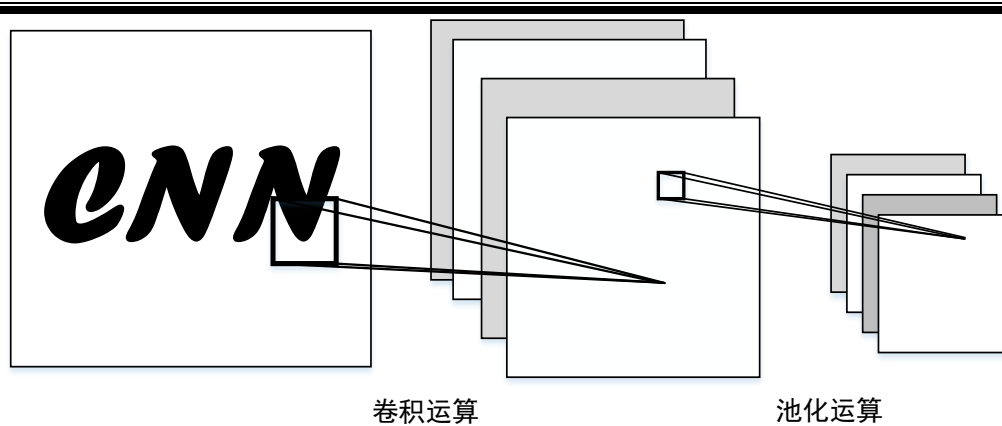

卷积运算                        池化运算

图 2.11 卷积神经网络单元结构示意图

### 2.4.2.1 卷积层

本文使用卷积神经网络处理人脸数据，在此使用图像作为介绍卷积神经网络的数据例子。传统的深度神经网络用矩阵乘法来表示输入数据和输出数据之间的联系。矩阵中的每一个参数都表述了输入单元和输出单元之间的某种交互信息。假设输入图像的大小为 500*500 像素，并且隐含层的节点数和输入层节点数相同。那么这一对神经元之间需要参数为 500^4，针对于深度神经网络，参数量还会急速增长。这样大规模的参数基本不可能进行训练。

通过研究人类视觉对外部图像的感知，可以知道人首先是从图像局部开始观测然后到全局观测。视网膜观测到的图像空间中的像素点的局部范围是有联系的。得到图像局部像素信息之后再扩展到较大范围的像素范围，进而得到整幅图片的信息。卷积神经网络具有类似的稀疏交互（Sparse interactions）的特征。通过卷积核函数对图像局部像素提取信息来减少特征量，如图：.

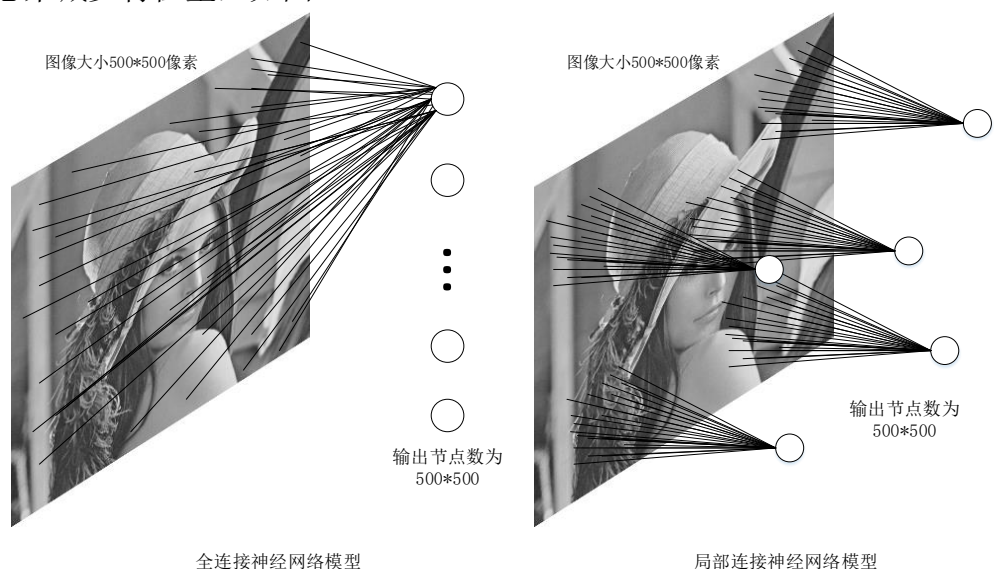

图 2.12 卷积神经网络中卷积核减少参数量原理图

从上图中可以看到全连接神经网络的隐含层每一个节点要和图片的每一个像素点都要有参数链接，一共有500^2个节点，总共需要500$^4$个参数。根据人类生物视觉原理的到的局部连接神经网络模型，假设每一个神经元对应于20*20像素大小的像素范围。那么此时参数个数为500$^2$*20$^2$。对于此前500$^4$个参数减少了非常多。

即便如此，这样的参数数目对于深度神经网络来说还是太大了。参数共享（parameter sharing）被用在卷积神经网络中。其定义在一个模型的多个函数中使用相同的权值参数。输出节点与图像数据之间的联系通过卷积核进行计算，如下图：

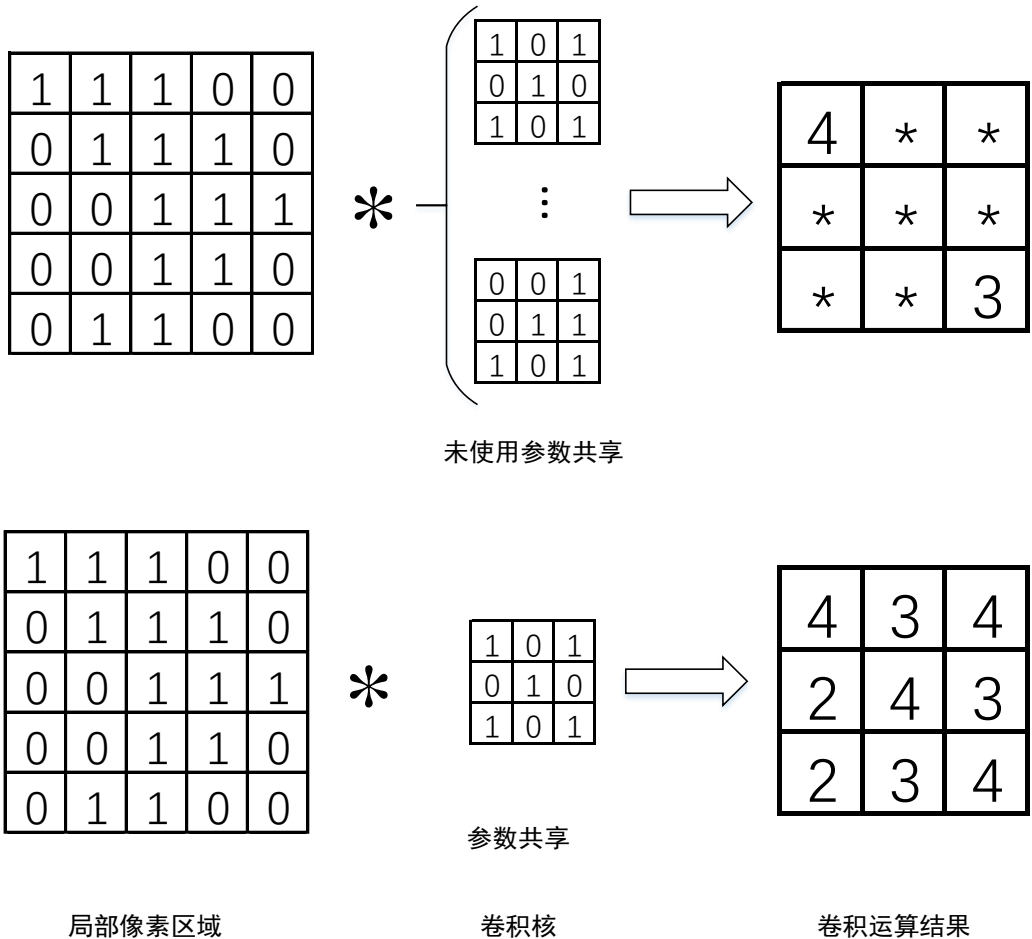

图 2.13 参数共享原理示意图

上图表示的是3*3卷积核在5*5大小图片上进行卷积的过程。未使用参数共享的时候，每一个神经网络节点都需要一个特定的卷积核得到。若定义图像的某部分的统计特性和其他部分是一样的，也就是说该部分的特性学习方式在其他部分也同样使用。那么神经节点可以共用一个卷积核。每一种卷积核代表着一种图像特征提取的方式。每一个神经元节点都使用了参数共享之后的卷积核，训练参数数量为20$^2$=400。这比500$^2$*20$^2$要少得多。

使用了参数共享之后的参数数目急剧下降。但是目前只使用了一种卷积核，这在图

像特征提取的应用中，特征提取显然是不够充分的。这可以通过添加卷积核种类，来使特征更加充分。假如有 100 种卷积核，那么这一百种卷积核会生成 100 张图片出来。每张图都可以看成是原图像经过不同通道之后产生的数据。

### 2.4.2.2 池化层

假设此卷积神经网络只包含一个卷积层，图像数据通过卷积层之后，需要对提取到的特征通过分类器（如 Softmax）进行分类。根据上面的假设，500*500 像素的图片，使用20*20的卷积核进行卷积，得到的特征数目为$(500-20+1)*(500-20+1)=231361$。假如有 100 种卷积核，那么得到的特征数目为 2313610。这么庞大的特征维度对于分类器来说太大了。

因此，池化层提出来可以很好地解决这个问题。池化函数使用局部输出的总体统计值来代替整个局部数据的输出。比如，最大池化（max pooling）函数使用相邻矩阵内的最大值替输出；均值池化函数使用相邻矩形内的均值代替输出等等。输入数据具有一定小的平移时，池化函数具有对数据的近似不变性。假如池化函数作用的矩形范围为5*5，那么特征维度可以降成原来的1/25。

针对500*500大小的图片，一个卷积层和一个池化层对其进行特征提取肯定是不够的。目前针对图像识别较好的卷积神经网络模型，有 VggNet、GoogleNet、ImageNet、ResNet 等。每种 CNN 模型都可以根据训练数据进行层参数调节，模型越复杂，越需要大量的数据对其进行训练，保证模型的收敛性和泛化性能。

在设计深度神经网络结构的时候，会根据实际情况除了卷积部分和池化部分还会添加其他的处理模块。例如在卷积和池化之间加入激活函数（ReLU、Sigmoid，TanH 等），或者在池化部分之后加入局部响应归一化进一步提高模型的泛化能力。

## 2.5 本章小结

本章主要介绍了深度神经网络的相关理论知识，首先介绍了最基本的前馈神经网络以及网络参数和神经元之间的关系。然后介绍了双层神经网络，并由此延伸到深度神经网络的相关概念，之后介绍了计算参数的复合函数和几种激活函数的形式。对于深层神经网络参数学习优化的问题，介绍了神经网络的反向传播算法。此算法在后续的深度神经网络模型的训练中都有使用，是一个相对重要的概念。

在介绍了基本的神经网络模型和学习机制之后，提出了具体的深度神经网络的优化算法。其中随机梯度下降法，是一阶优化算法中较为重要的方法。自适应梯度下降法、自适应学习率法、加速梯度下降法等一阶优化算法，都是基于随机梯度下降法变化得到

的。然后介绍了优化算法本身，学习率设置和参数优化的技巧。

　　深度神经网络不光需要优秀的优化算法，深度网络模型在优化的过程中很容易陷入局部优化和过拟合的情况，因此提出了深度神经网络的正则化。正则化保证了网络模型在训练集合外的数据上具有优秀的泛化能力。这是一个优秀的神经网络模型所具备的特性。最后根据本文要使用的深度神经网络，介绍了深度置信网络和卷积神经网络两种深度神经网络，并进行了相关函数和方法得推导。

# 第三章 基于 CNN 的人脸特征提取方法

人脸识别属于机器视觉模式识别，鉴于人脸识别在复杂的空间内具有三维形态可变性，人脸的采集会受到光照、表情、遮挡、姿态等影响。这些影响因素会限制识别算法的性能。人脸图像送入识别算法前，要经过一些预处理。比如降噪、模糊处理、人脸对齐等等。本文基于 CNN 的人脸识别过程如图 3-1。

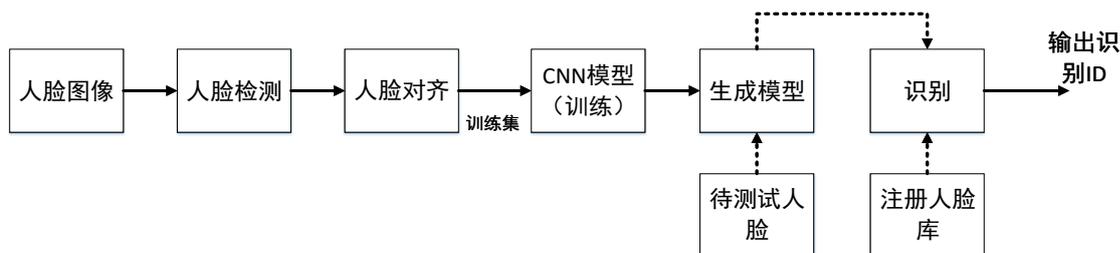

图 3.1 基于 CNN 人脸识别系统框图

## 3.1 人脸预处理

人脸图像的预处理是人脸识别系统中不可或缺的部分。本小结从人脸检测和人脸对齐两个方面来研究人脸图像预处理问题。

### 3.1.1 人脸检测

目前人脸检测的算法模型有很多，例如弹性模型匹配、肤色模型、SVM 法、ANN 模型、AdaBoost 法等。目前 AdaBoost 学习算法是相对简单高效的。对于大数据量的人脸图像检测来说，具有识别速度上很大的优势。AdaBoost 算法首先对训练样本进行学习得到弱分类器。然后根据弱分类器的识别结果来降低正确样本的权值，提高错误样本的权值。根据重新设定权值的样本再次学习，重复上面的步骤。经过多次循环之后得到一系列弱分类器。最后把这些弱分类器乘上相应的权值求和，就可得到具有很强分类能力的强分类器[46]。

人脸检测问题中，AdaBoost 使用样本数据的 Haar-like 特征，如图 3-2。计算方法是将图中白色区域的像素减去黑色区域的像素，人脸区域和非人脸区域的 Haar-like 特征相减的值肯定是不一样的。因此可以通过使用人脸的 Haar-like 特征，训练分类器。

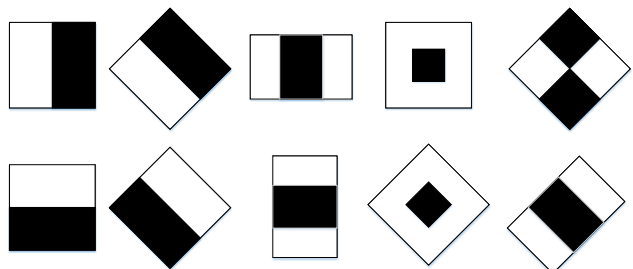

图 3.2 部分 Haar-like 特征示意图

根据上图可以得到 Haar-like 特征值为 V = S(白)-S(黑)，改变模块中分割线的位置，就可以得到不同的特征。假设使用图 3-2 中左上角第一个和左下角第一个两个类型模块共同构造，并假设其大小为12*12，可以得到 2808 个特征。因此使用 Haar-like 特征可以构造出大量的信息。根据 Haar-like 的构造可以检测出包括边缘特征、中心特征、线性特征、和对角线特征。由这些种类的特征构造特征模板。但是由于特征值数量很大，并且要在大量的人脸和背景图片上进行学习，因此计算非常大。为了提高算法的速度，引入积分图来表示图片数据。

积分图表示的是把图像的起点到各个点之间所形成的矩形区域像素的和作为元素保存在数组内存中[47][48]。每次要计算某个区域的时候，可以直接索引数组元素，从而不需要计算这个区域的像素和。这个算法通过牺牲内存，提高计算速度。

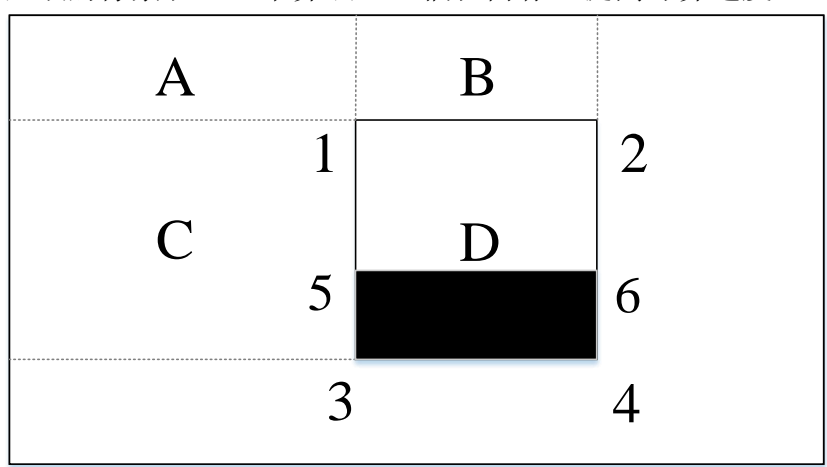

图 3.3 积分图示意图

定义像素和计算公式为：

$$F(x, y) = \sum_{x' \leq x} \sum_{y' \leq y} I(x', y')$$

3-1

其中 $F(x, y)$ 为点图像中 $(x, y)$ 左上角矩形范围的像素和，$I(x', y')$ 表示图中点所在的像素值。因此在图 3-3 中1, 2, 3, 4四个点在数组中保存的值为：

$$F(1) = S(A)$$
$$F(2) = S(A) + S(B)$$
$$F(3) = S(A) + S(C)$$
$$F(4) = S(A) + S(B) + S(C) + S(D)$$

3-2

如果要求 D 区域的像素和，可以通过查找1, 2, 3, 4在数组中的值直接求得：

$$S(D) = F(1) + F(4) - F(2) - F(3)$$

3-3

若要求 D 区域的 Haar-like 特征，通过数组中存储的值和索引信息直接得到：

3-4

每次求 Haar-like 特征值时的时间消耗在于索引数组中坐标所在的值，可以大幅提高计算速度。

特征提取出来之后送入 AdaBoost 算法模块进行训练，得到多个弱分类器，然后将这些弱分类器乘上对应的权重合成强分类器。假设输入的一组特征样本为：$(x_1, y_1),(x_2, y_2)\ldots(x_n, y_n)$，$x_i$ 表示特征值，$y_i$ 表示特征对应的标签，0（非脸）或者 1（脸）。然后初始化训练样本的权值：$W_i = \dfrac{1}{2N}$ 和 $W_i = \dfrac{1}{2M}$ 分别表示人脸特征和非人脸特征的初始化权值。其中 $N$ 和 $M$ 分别为人脸样本和非人脸样本的总数。

假设训练过程循环 $K$ 次，那么 AdaBoost 学习过程为：

i.  权值归一化：

$$l_{k,i} = \frac{W_{k,i}}{\sum_{j=1}^{M} W_{k,j}}$$

3-5

ii. 然后用特征 $z$ 训练弱分类器 $h(x, z, p, \beta)$，其中 $p$ 和 $\beta$ 是弱分类器的参数，可以求得该分类器对特征的加权错误率为：

$$\varepsilon_z = \sum_i l_i \left| h(x_i, z, p, \beta) - y_i \right|$$

3-6

iii. 根据最小错误原则来选择最佳弱分类器 $h_k(x)$，其中最小错误率为：

$$\varepsilon_k = \min_{z, p, \beta} \sum_i l_i \left| h(x_i, z, p, \beta) - y_i \right|$$

3-7

iv. 根据求得的弱分类器来调整样本的权值：

$$W_{i+1,i} = W_{k,i} \gamma_k^{1-e_i} \quad \text{其中} \gamma_k = \frac{\varepsilon_k}{1 - \varepsilon_k}$$

3-8

v.  结合弱分类器和权值集成强分类器为：

$$C(x) = \begin{cases} 1 & \sum_{k=1}^{K} \lambda_k h_k(x) \geq \frac{1}{2} \sum_{k=1}^{K} \lambda_k \\ 0 & \text{其他} \end{cases}$$

3-9

其中 $\lambda_k = \log \dfrac{1}{\gamma_k}$ 。

前面提到的假如只有 2 个 12*12 的 Haar-like 可以生产 2808 个特征。若加入更多的 Haar-like 模板，特征总数会急剧上升。这样庞大的数据量即使用了积分图来提升，后面的 AdaBoost 的训练过程也会消耗大量的时间。因此用 AdaBoost 训练 Haar-like 特征的人脸检测算法训练过程消耗较大，但是识别过程较为迅速[49]。图 3-4 为本文实现 AdaBoost 法的识别结果。在人脸变形 3-D 畸变不大的时候，检测效果能得到保障。

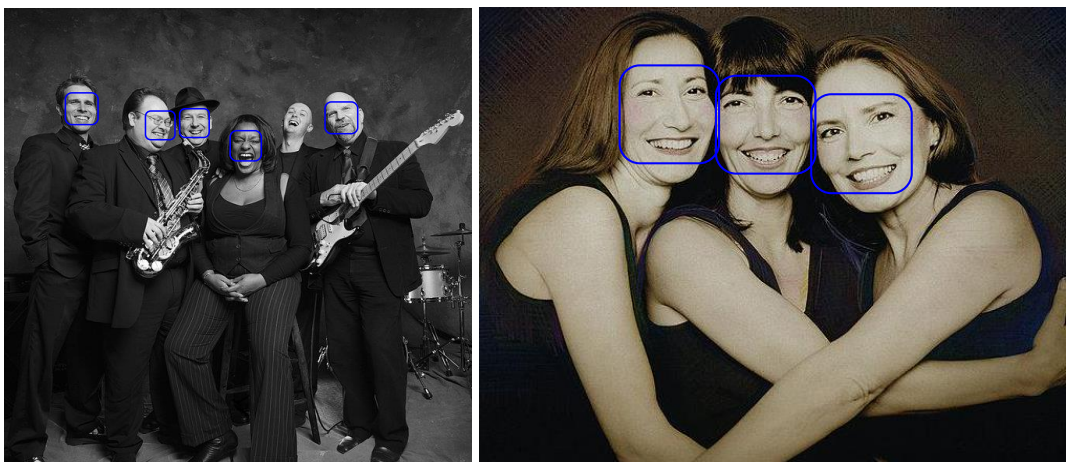

图 3.4 AdaBoost 人脸检测结果

### 3.1.2 人脸对齐

上小节算法的主要功能是将人脸从图像中找出来。但是实际环境中，摄像头拍摄到的人脸往往不是中规中矩的状态，会出现倾斜、偏侧、上下倾角等情况。这样的人脸数据送入训练或者识别都是会影响到系统的性能[50]。因此人脸识别的矫正预处理对提高人脸识别性能十分重要。

人脸对齐的主要思路是在已检测到的人脸图像上对人脸进行特征点检测。然后通过找到的特征点进行仿射矩阵计算。通过仿射变换将姿态不正的人脸调节成端正的中性脸。

如图 3-5，其中图（A）是人脸关键点准确位置，图（B）中的关键点是初始化位置。现在人脸关键点检测可以建模成将图（B）中的点变化到图（A）的位置。这个模型可以看成机器学习中的非线性优化问题。

假设图（A）中的关键点坐标为 $X$，图（B）中点关键点坐标为 $x$。则可以假定 $l(x) \in R^{n*1}$ 为图中 $n$ 个特征点的坐标。然后对每一个特征点求其 sift 特征，这样每个特征点变化成 128 维特征。假设特征提取函数为 $h$，那么 $h(l(x)) \in R^{128n*1}$。因此上述的建模问题可以由最小化函数：

$$f(x + \Delta x) = \left\| h(l(x + \Delta x)) - h(l(X)) \right\|^2 \qquad \text{3-10}$$

求解变量 $\Delta x$ 使得函数 $f(x + \Delta x)$ 最小。对非线性目标函数求解，通常使用二阶导数进行优化，比如牛顿法[51]。但是牛顿法优化非线性函数有几个限制条件：1. 牛顿法要求目标函数是二次可微的；2. Hessian 矩阵在局部最优时才是正定的，在其他地方有可能不是正定的，这会导致解出来的梯度方向不一定是下降的方向。3. 针对求人脸特征点

这种情况，每个特征的的 sift 特征为 128 维。60 个特征点生成的向量会有 7680。对 7680*7680 的逆矩阵求解，计算量太大。

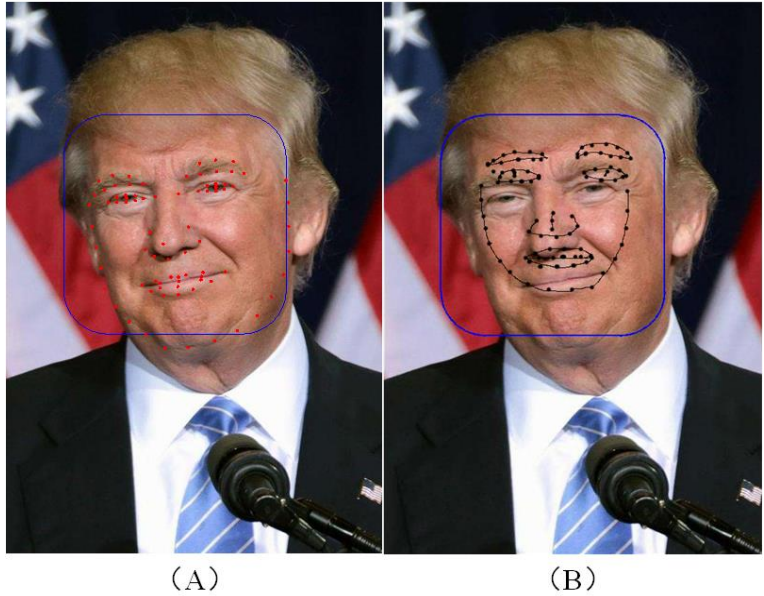

（A） （B）

图 3.5 人脸特征点检测优化示意图

用牛顿法对式 3-10 进行求解可以得到：

$$x_k = x_{k-1} - 2H^{-1}J_l^T\left(\phi_{k-1} - \phi_x\right)$$ 3-11

其中 $\phi_x = h\left(l\left(X\right)\right)$、$\phi_{k-1} = h\left(l\left(x_{k-1}\right)\right)$、$H$ 为 Hessian 矩阵、$J$ 为 Jacobian 矩阵。将式 3-11 进一步分解可以得到：

$$x_k = x_{k-1} - 2H^{-1}J_l^T\phi_{k-1} + 2H^{-1}J_l^T\phi_X$$ 3-12

其中由于矩阵 $H$ 和 $J$ 的维度太高，计算量太大。在此求 $H$ 和 $J$ 得乘积。将上式转化成：

$$x_k = x_{k-1} + C_{k-1}\phi_{k-1} + b_{k-1}$$ 3-13

其中 $C_{k-1} = 2H^{-1}J_l^T$、$b_{k-1} = 2H^{-1}J_l^T\phi_X$。因此优化问题变成了求解 $C_{k-1}$ 和 $b_{k-1}$ 的问题。到此处开始使用监督下降法（Supervised Descent Method,SDM）来对其求解[52]。对于训练人脸图像集 $d$，进行迭代训练。假设第一次训练，其目标函数为：

$$\arg\min_{C_0,b_0}\sum_{d_i}\left\|\Delta X_i - C_0\phi_0^i - b_0\right\|^2$$ 3-14

通过训练集得到 $C_0$ 和 $b_0$，对于第 $k$ 次迭代可以得到参数 $C_{k-1}$ 和 $b_{k-1}$。通过人脸训练集逐步迭代找到最优的参数对人脸的特征点进行收敛。人脸的特征点标定效果如下图。

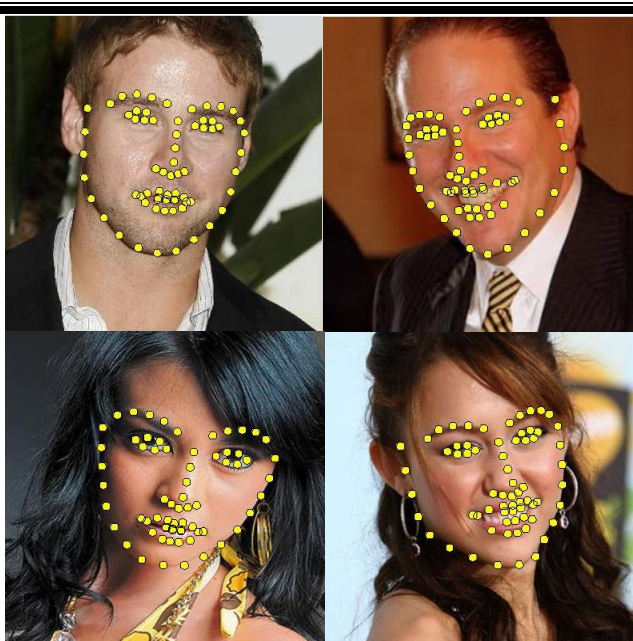

图 3.6 基于 SDM 的人脸特征点标定

在得到人脸的特征点的坐标信息之后，给定一个目标模板的特征点坐标信息。使用仿射变换，可以对人脸进行对齐。给出标准平均脸的特征点坐标位置为：$(x_i, y_i)$，假设通过 SDM 定位得到的特征点坐标为：$(x_i^{'}, y_i^{'})$。对于第 $i$ 个点，对应的变换方程为：

$$\begin{bmatrix} x_i \\ y_i \\ 1 \end{bmatrix} = \begin{bmatrix} a & b & c \\ d & e & f \\ 0 & 0 & 1 \end{bmatrix} \begin{bmatrix} x_i^{'} \\ y_i^{'} \\ 1 \end{bmatrix} = \mathbf{M} \begin{bmatrix} x_i^{'} \\ y_i^{'} \\ 1 \end{bmatrix} \qquad 3\text{-}15$$

假设 SDM 定位出来的特征点个数为 n，可以写出变换方程为：

$$\begin{bmatrix} x_1 \\ y_1 \\ x_2 \\ y_2 \\ \vdots \\ x_n \\ y_n \end{bmatrix} = \begin{bmatrix} x_1^{'} & y_1^{'} & 0 & 0 & 1 & 0 \\ 0 & 0 & x_1^{'} & y_1^{'} & 0 & 1 \\ x_2^{'} & y_2^{'} & 0 & 0 & 1 & 0 \\ 0 & 0 & x_2^{'} & y_2^{'} & 0 & 1 \\ \vdots & \vdots & \vdots & \vdots & \vdots & \vdots \\ x_n^{'} & y_n^{'} & 0 & 0 & 1 & 0 \\ 0 & 0 & x_n^{'} & y_n^{'} & 0 & 1 \end{bmatrix} \begin{bmatrix} a \\ b \\ c \\ d \\ e \\ f \end{bmatrix} \qquad 3\text{-}16$$

其中未知变量为 $a,b,c,d,e,f$，设为向量 $h$。可将上式变换成：$Z = Kh$。根据式 3-15 可知三个特征点可以求得向量 $h$。为了减小变换矩阵的误差，取用特征点往往大于 3 个。使方程的个数大于未知数个数形成超定方程。然后通过最小二乘法求方程的解：

$$h = \left( K^T K \right)^{-1} K^T Z \qquad 3\text{-}17$$

通过上式求得位置向量 $h$，然后对原图的像素进行仿射变换[53]，得到对齐后的结果见下图。

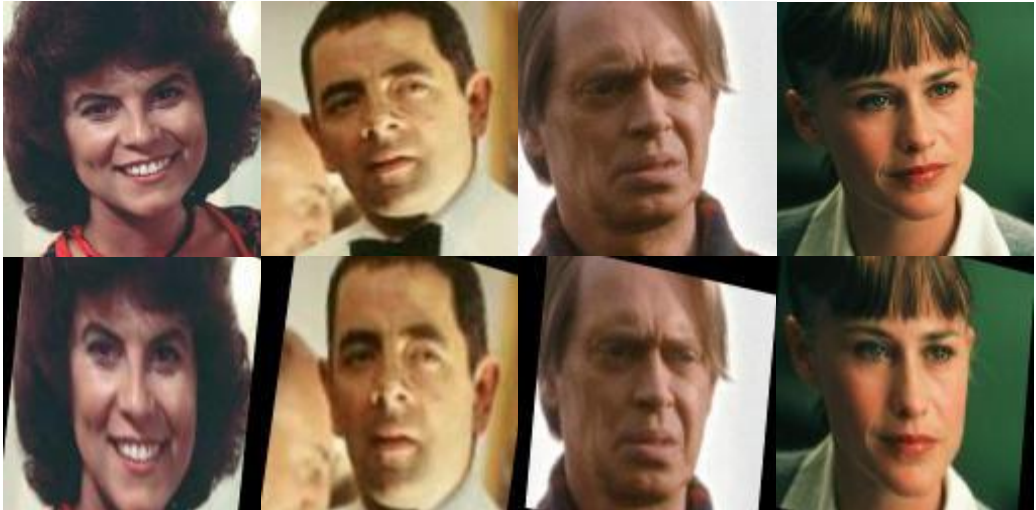

图 3.7 基于特征点仿射变换的人脸对齐

## 3.2 人脸识别模型及其改进

目前基于深度学习的图片分类的网络模型有很多，例如多伦多大学的 AlexNet、牛津大学的 VggNet、Google 公司的 GoogleNet 等。将这些图片分类模型用人脸数据库进行训练，并且对网络结构的调节和参数的微调，可以训练出具有优秀泛化能力的人脸特征提取模型[54]。

本文中提取人脸特征的模型是基于 Vgg_Face 模型的改进型。在介绍 Vgg_Face 模型前，首先来看多伦多大学的 Alex Krizhevsky 提出来的深度卷积神经网络 AlexNet。AlexNet 结构很大程度上推动了深度神经网络的发展，在人脸识别问题上 Vgg_Face 模型也是在 AlexNet 结构的基础上做了更深层的推广。在 ImageNet LSVRC-2012 比赛中，120 万张图片分成 1000 类。AlexNet 对其测试集的 Top-1 误差率为 27.5%，Top-5 的误差率为 15.3%。其中包含了五个卷积层和三个全连接层，如图 3-8。

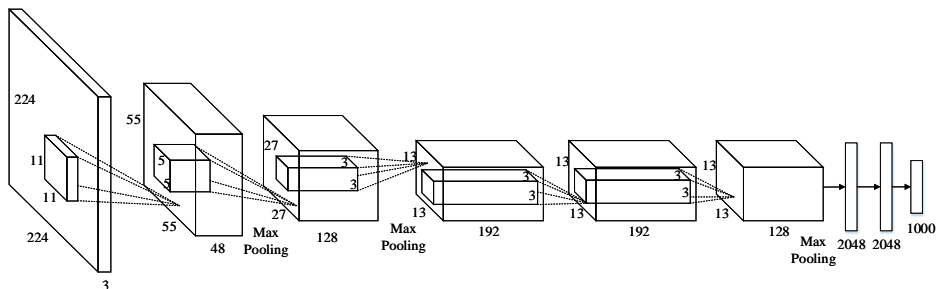

图 3.8 AlexNet 网络结构示意图

上图中，图像数据送入网络中，首先通过前五层卷积层进行特征的抽象提取，然后进入后三层的全连接层进一步处理，最后得到的全连接层数据载送入 Softmax 等分类器中进行分类。最终得到 1000 个类别的分布概率。在 AlexNet 网络结构中，卷积层包含了卷积，激活，池化、局部响应归一化（Local Response Normalization，LRN）四个部

分。

Omkar M. Parkhi 提出的 VGGNet 模型在总体结构上和 AlexNet 一致，但是提出了去掉 LRN 层的结构。其认为 LRN 层在大数据量的情况下对识别率的影响不大，但是增加了模型训练的计算量。VGG_Face 模型见表 1。

表 1 VggNet 网络结构模型

| 卷积层结构类别 | | | | | |
|---|---|---|---|---|---|
| A | A-LRN | B | C | D | E |
| 11 个参数层 | 11 个参数层 | 13 个参数层 | 16 个参数层 | 16 个参数层 | 19 个参数层 |
| 输入图片（224×224 RGB） | | | | | |
| Conv3-64 | Conv3-64 +LRN | Conv3-64 Conv3-64 | Conv3-64 Conv3-64 | Conv3-64 Conv3-64 | Conv3-64 Conv3-64 |
| MaxPooling | | | | | |
| Conv3-128 | Conv3-128 | Conv3-128 Conv3-128 | Conv3-128 Conv3-128 | Conv3-128 Conv3-128 | Conv3-128 Conv3-128 |
| MaxPooling | | | | | |
| Conv3-256 | Conv3-256 Conv3-256 | Conv3-256 Conv3-256 | Conv3-256 Conv3-256 Conv1-256 | Conv3-256 Conv3-256 Conv3-256 | Conv3-256 Conv3-256 Conv3-256 Conv3-256 |
| MaxPooling | | | | | |
| Conv3-512 Conv3-512 | Conv3-512 Conv3-512 | Conv3-512 Conv3-512 | Conv3-512 Conv3-512 Conv1-512 | Conv3-512 Conv3-512 Conv3-512 | Conv3-512 Conv3-512 Conv3-512 Conv3-512 |
| MaxPooling | | | | | |
| Conv3-512 Conv3-512 | Conv3-512 Conv3-512 | Conv3-512 Conv3-512 | Conv3-512 Conv3-512 Conv1-512 | Conv3-512 Conv3-512 Conv3-512 | Conv3-512 Conv3-512 Conv3-512 Conv3-512 |
| MaxPooling | | | | | |
| FullContact-4096 | | | | | |
| FullContact-4096 | | | | | |
| FullContact-1000 | | | | | |
| Softmax | | | | | |

其中 Conv3-64 表示此卷积层使用的是 3×3 的卷积核，卷积核的种类为 64。VggNet 和 AlexNet 不同。在池化层中间，AlexNet 使用一个卷积层，而 VggNet 使用多个卷积层。其通过减小卷积核的尺寸，然后增加卷积层的数目来达到抽象数据特征的作用。

Omkar M. Parkhi 在其 2015 年的论文中，使用的是 VggNet 的 E 网络结构，使用 2.6 兆的照片对模型进行训练，并且提出了将 Embedding learning 和 Vgg_Face 相结合的方法。在此之前模型使用 Softmax 分类器，在 LFW 库上可达到 97.27%，在 YouTobe Faces Datasets 上达到 92.8%的识别率。本论文中使用的人脸部分的特征提取模型正是基于此模型上的改进模型。

本课题将语音 I-Vector 特征和图像卷积层后的特征相融合，然后将融合特征送入

DBN 进一步训练提取抽象特征。为了提高双模态识别的性能，需要对人脸识别模型进行一定的改动。基于本课题的试验结果，在 Vgg_Face 网络结构的倒数第二个全连接层添加一个节点为 2048 的全连接层，以提高双模态特征融合训练的识别率。改进后的模型结构如下表：

表 2 基于 Vgg_Face 的改进模型

| 输入图片（224×224 RGB） |
| :---: |
| Conv3-64 |
| Conv3-64 |
| MaxPooling |
| Conv3-128 |
| Conv3-128 |
| MaxPooling |
| Conv3-256 |
| Conv3-256 |
| Conv3-256 |
| Conv3-256 |
| MaxPooling |
| Conv3-512 |
| Conv3-512 |
| Conv3-512 |
| Conv3-512 |
| MaxPooling |
| Conv3-512 |
| Conv3-512 |
| Conv3-512 |
| Conv3-512 |
| MaxPooling |
| FullContact-4096 |
| FullContact-4096 |
| FullContact-2048 |
| FullContact-** |
| SoftMax |

## 3.3 人脸识别改进网络模型训练

本文使用的训练方法是在 Vgg_Face 模型上使用自建的数据库对其进行调整（Fine Tune）。由于实验环境的限制，在本文周期内通过重新训练 Vgg_Face 模型各个参数是不现实也是没有效率的事情。并且在本文中，对 Vgg_Face 改进的地方在于模型的后几层全连接层，因此前面的卷积层部分的网络参数结构可以继续延续。而 Fine Turn 的优势在于不需要从头训练整个网络的参数，只需要通过自己的训练样本和重新设定学习参数就可以在较少的学习迭代次数后得到预期的结果。因为 Vgg_Face 模型通过 2.6M 张人脸图像进行训练，整个模型的参数对人脸具有很好地类内区分性能。通过自己的数据对模型参数进行训练微调，不失为一种高效的办法。

训练模型使用的人脸数据库为 CASIA-WebFace 库，这个库包含了 10575 个人，一

共 494414 张人脸照片，如图 3-9。照片是网上半自动获取的名人脸部照片，每个人的姿态表情都不一样，并且光照条件也很丰富，这对训练鲁棒性强的模型非常有帮助。但是这个库的样本不平衡，一个类别中样本包含最多的有 400 多张照片，最少的只有 20 多张照片。并且对姿态较大的样本进行挑选之后，类别间的样本数目相差较大。这对模型的训练十分不利。因此，在对样本进行预处理之后，挑选样本数目为 50 以上的类别作为训练集，并且每个样本集中只选取 50 个样本。并命名为 CASIA-WebFace-A。剩余样本数目小于 50 的类别划分到测试集，命名为 CASIA-WebFace-B。

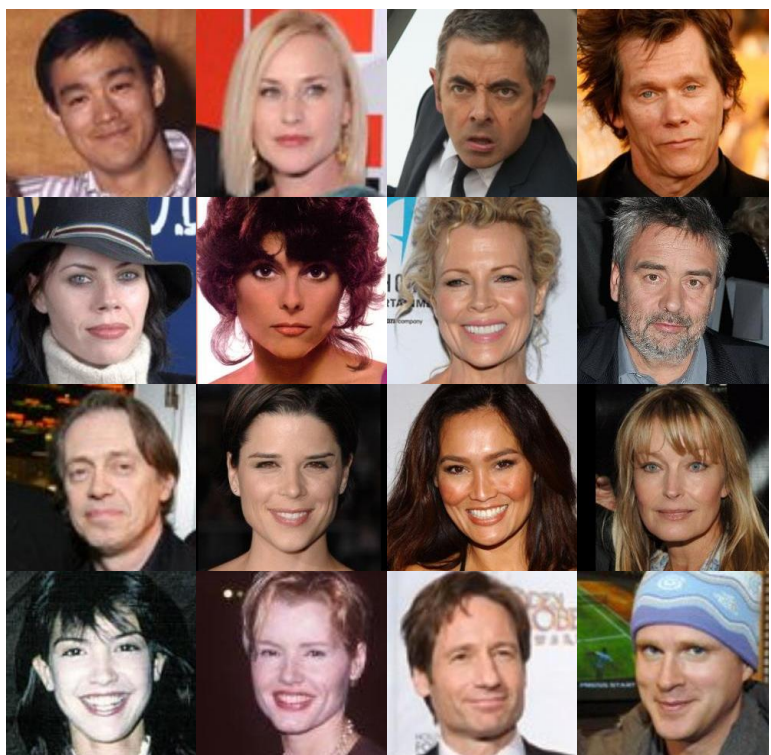

图 3.9 CASIA-WebFace 库缩略图

使用 3.1 节中的人脸预处理方法，对 CASIA-WebFace 进行处理。

1. 使用基于 Haar-Like 的 AdaBoost 方法进行人脸检测。对检测不到的人脸图片进行剔除；
2. 使用 SDM 算法对人脸关键点进行标定；
3. 通过标定的关键点计算仿射变换矩阵，然后做人脸对齐处理。最后将人脸部分抠出；

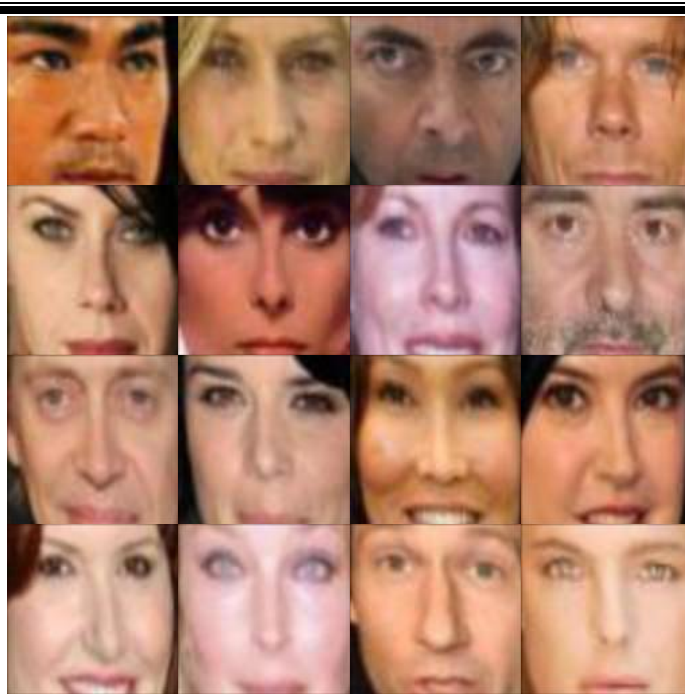

图 3.10 人脸图像预处理后结果

　　最终处理完的数据如上图。经过对数据库中某些类别进行区分，CASIA-WebFace-A 中的类别数总共为 8216 个，CASIA-WebFace-B 中的类别数为 2359。在 CASIA-WebFace-A 中随机选取 45 个样本作为模型训练集，剩下的 5 个样本作为训练测试集（在模型训练过程中，每经过固定迭代次数，对网络模型识别率测试的数据集）。CASIA-WebFace-B 为开集测试库，在第 5 章中会和语音库一起组成融合特征库。为保证和其它模型对比的公平性，在本章中选取了 LFW 为模型开集测试库，LFW 为目前国际上人脸识别公开测试库。LFW 库包含 5749 类共 13233 张图片，其中 1680 人包含两张以上照片。为保证训练模型的性能，本文使用相同的人脸检测算法和人脸对齐算法对 LFW 库进行预处理，在没有检测出人脸的数据，进行人为的框定。使用 LFW 官方提供的测试集，包含正（同一个人）测试对共 3000 对，负（不同的人）测试对 3000 对。通过这 6000 对人脸对训练好的模型进行验证。详见表 3。

表 3　人脸数据样本分布

| 样本集名称 | 训练集 | 训练测试集 | 正测试集 | 负测试集 |
|---|---|---|---|---|
| 数量（类别*每类样本数） | 8216*45 | 8216*5 | 3000 对 | 3000 对 |

　　本文人脸部分的模型训练使用的深度学习训练平台为卷积神经网络框架（Convolution Architecture For Feature Extraction，Caffe）。Caffe 使用文本模式来定义模型和优化，并非代码模式。对于新的训练任务配置起来十分灵活，可以很方便的供使用者快速编辑。并且 Caffe 的计算速度非常快可以搭配 GPU 并行计算，适合目前前沿的模

型和海量数据库的训练。

训练人脸识别模型步骤如下：

1. 首先将设计好的人脸数据（训练集和训练测试集）原始图片（128*128pi）转化成 Caffe 中设定的 Lightning Memory-Mapped Database（LMDB）数据格式（训练集.lmdb 和训练测试集.lmdb）。由于训练过程中需要反复读取大量的训练数据。若训练数据用原始格式存放在硬盘中，每次访问训练文件的 I/O 开销将会非常大。LMDB 使用内存映射的方法保存文件，文件内部的寻址开销很小，通过简单的指针寻址完成文件的赋值和传输。几万个文件的数据集通过一个文件库管理，在进行训练的时候，可以大幅减小文件 I/O 开销影响训练的效率；

2. 对训练集中的数据求均值。神经网络训练中的数据预处理可以加快模型的收敛速度。由于图像数据之间具有较强的相关性，在对特征进行分类的时候，由于相关特性的存在会导致特征之间的区分度会存在于空间的狭小部分。对数据进行预处理即在送入训练之前对数据进行一定的人为去相关操作，增加特征之间的区分度。例如对模型的白化处理，PCA 或者减均值都可以一定程度上的降低数据间的相关程度。由于求协方差矩阵会消耗大量时间和空间，通过减去均值来扩大数据的分布范围是权衡计算量和收敛时间之间性价比较高的方法；

3. 在完成训练数据的预处理之后，通过 Prototxt 文本编辑的方式设定可视化的训练网络的结构。根据表 2 中对训练网络的设计，编辑 Prototxt 文件；

4. 完成网络结构的编辑之后需要设定模型训练的相关参数，包括学习率，优化函数、学习率收敛方式、激活函数等；

5. 调用 Vgg_Face 模型，将最后两层的全连接层的网络结构名称设置为自定义名称，保证在 Fine Tune 的时候能够重新调节整个网络结构的参数；

6. 通过训练结果的损失曲线，识别率曲线来重新调参，直到训练模型达到预定效果；

本文根据硬件条件以及选用的数据库，设定参数如下：

1. 原始图像数据像素大小为128*128；

2. 初始学习率为 0.0005；

3. 学习率下降策略为"Step"；

4. 学习率下降步长为 20000 迭代次数；

5. 学习率下降 Gamma 参数为 0.96；

6. 总迭代次数为 120000 次；

7. 训练数据的 Batchsize 为 64，训练测试数据的 Batchsize 为 32；

本次模型训练的硬件参数为：

表 4 人脸识别模型训练硬件参数

| 名称 | 参数 |
|------|------|
| CPU | Intel i5 3450 主频 3.1GHz 6M 三级缓存 |
| 内存 | Kinston DDR3 16G |
| 显卡 | NVIDIA GeForce GTX Titan X 显存 12G |

## 3.4 人脸特征提取及识别

识别模型训练过程及结果如下图：

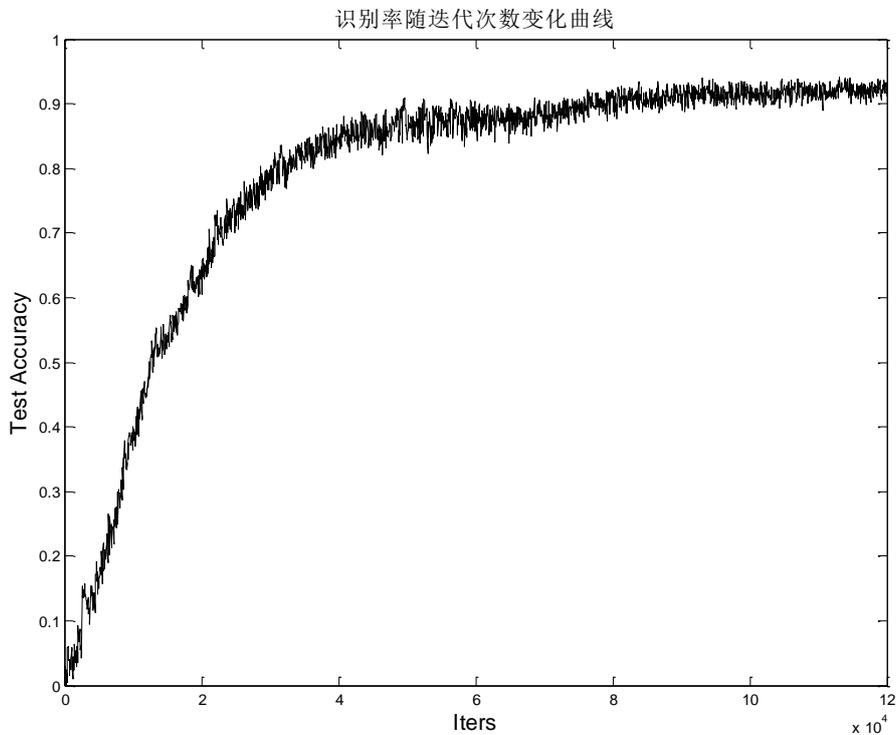

图 3.11 识别率曲线

上图为使用 CASIA-WebFace-A 库对网络模型训练过程，模型对训练测试集中样本的识别率变化曲线。从上图可以看到，在模型最开始训练的时候识别率不为零，这是由于 Vgg_Face 在其原来的训练库上训练的参数具有一定的泛化能力，由于增加了一个全连接层导致改进网络对 CASIA-WebFace 库最初始的识别率没有直接用 Vgg_Face 进行识别的识别率高。经过大概 800 次的训练之后，改进网络的识别率开始快速上升。由测试集损失函数的变化曲线（图 3-12）也可以看出在经过 800 次的迭代之后损失函数值开始迅速下降。如果不调用 Vgg_Face 训练好的参数，用改进网络训练这么大的数据，实际

的训练速度远没有这么快。大概经过 6 万次的迭代之后识别率的值开始趋于稳定，最终经过 12 万次的训练得到了 94.13% 的识别率。

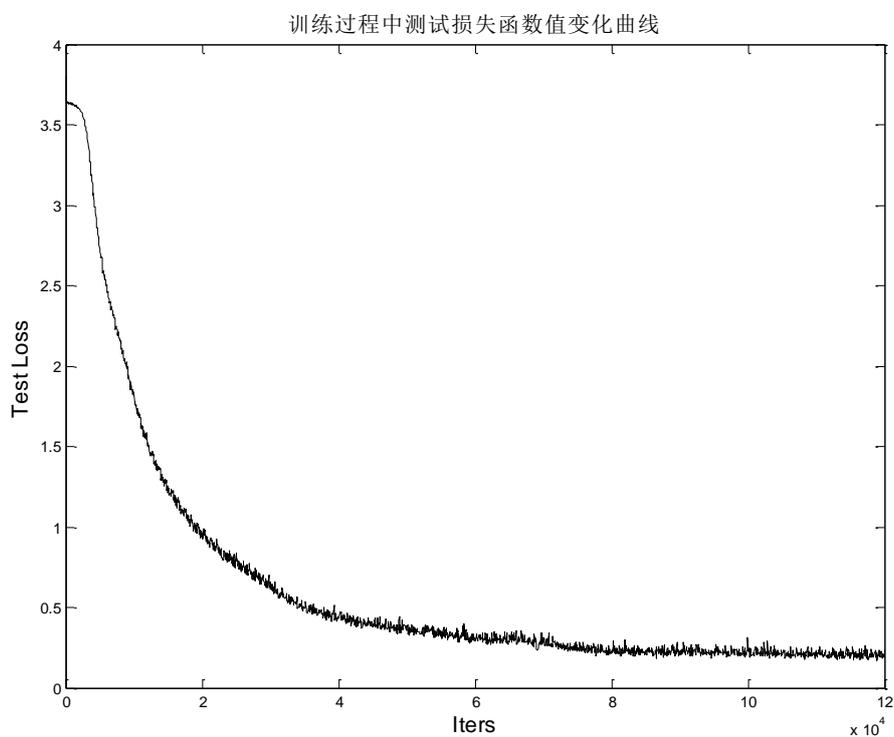

图 3.12 测试集损失函数变化曲线

模型训练好之后，就可以将人脸图像输入到模型中进行特征提取。在表 2 中的每一层都可以提取出特征。其中在卷积层部分的特征可以通过 Caffe 进行可视化。

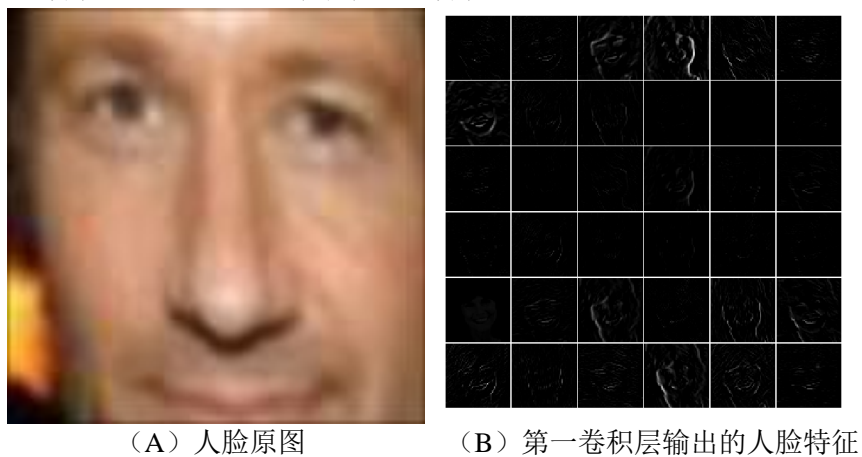

（A）人脸原图　　　　　　（B）第一卷积层输出的人脸特征

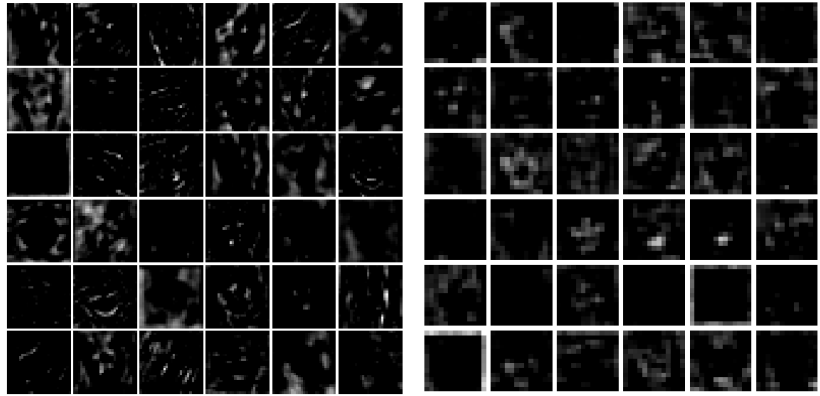

（C）第二卷积层输出的人脸特征　　（D）第三卷积层输出的人脸特征
图 3.13 图像在深度神经网络每层特征示意图

可以从上图中看到，横向对比：在卷积层处理之后，即使是同一层输出的特征，由于使用的卷积核不同，生成的特征形式也不一样。纵向对比：同一幅图片经过不同的卷积层之后得到的特征形式也发生了变化。从低层到高层，可以看到图像的特征逐步抽象。然后经过 4 个全连接层进一步进行抽象学习，最终的特征经过 Softmax 进行分类。

对训练好的识别模型，本文在 LFW 库上做开集测试。这个库中的人脸的姿态，光照影响较为剧烈，使用本文提出的人脸检测和矫正算法进行处理后，有一部分人脸未检出。未检测出的照片通过手动标定裁剪。将两张照片分别通过训练好的模型，经过逐层计算并提取出倒数第二个全连接层输出的数据，将这层输出的数据作为图像特征，对这两个数据做余弦距离计算。由于不同人脸在复杂的环境下进行特征提取，得到的特征向量可能存在度量标准不同意，余弦距离由于其对向量的绝对值不敏感，而是对特征的方向较为敏感，余弦距离可以很好地反映人脸特征向量的方向相似度。

LFW 人脸库的测试集提供了 10 组照片集，每组照片集里面有 600 对照片，同属一个人和不同人的分别有 300 对。通过模型计算出 6000 对照片的特征向量，并且计算器余弦距离数值。找出 6000 个余弦距离值中的最大值和最小值，用最大值减去最小值，以此差值做为归一化分母，将 6000 对余弦距离值减去最小值除以该分母。从而将所有的特征向量的余弦距离归一化到 $[0,1]$。通过调整阈值来统计这 6000 对数据中的正确接受（Ture Positve）数目、漏报（False Negative）数目、误报（False Positive）数目和正确拒绝（Ture Negative）数目。部分阈值的识别具体结果见表 5。

表 5 改进网络模型在 LFW 库识别结果（部分）

| 样本类别<br>归一化阈值 | 正样本（正确 Ture Positive/错误<br>False Negative） | 负样本（正确 False Positive/错误<br>Ture Negative） |
|---|---|---|
| 0.3800 | 2724/276 | 51/2949 |

| 0.4165 | 2849/151 | 151/2849 |
| 0.5000 | 2901/99 | 247/2753 |
| 0.6500 | 2992/8 | 1268/1732 |

统计所有选取得阈值的 Ture Positive 率和 False Positive 率画出 ROC 曲线（图 3-14），并且给出了本文的 ROC 曲线和 Baidu、DeepID3 和 Vgg_Face 的 ROC 曲线比较（图 3-15）。

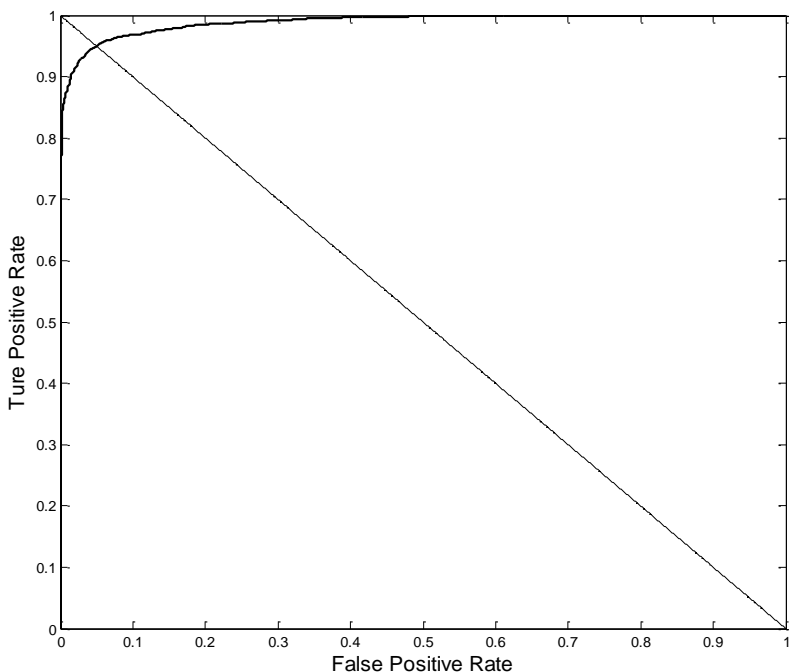

图 3.14 识别模型在 LFW 数据库上 ROC 曲线

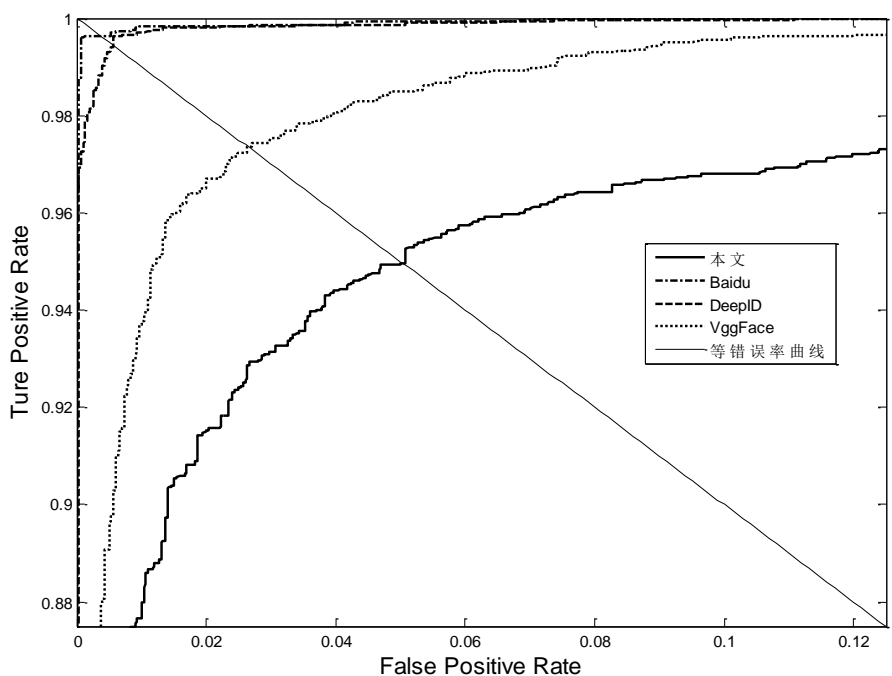

图 3.15 ROC 曲线比较

可以从图中看到，在正确识别率和误识别率相等的时候，此时的系统识别率为 94.97%。此时归一化阈值取值为 0.4165。本文的识别模型是基于 Vgg_Face 模型更改的，为了适用于双模态特征融合识别的需求，在 Vgg_Face 的网络层上增加了一个全连接层。并且从新通过 CASIA-WebFace-A 库对模型参数进行训练。可以从图 3-15 中看出本文的模型和 Vgg_Face 模型的性能上还存在一定的差距。分析具体的原因有以下几点：

1. 训练的数据集没有 Vgg_Face 原始训练数据集好，数据集中包含较多的噪声数据。

2. 网络结构进行了一定的改变，导致最后的性能没有 Vgg_Face 的优秀

3. 在网络模型的训练过程中，参数调节的还不够理想。导致最终的结果和 Vgg_Face 存在差距。

虽然目前改进型 Vgg_Face 模型没有 Vgg_Face 原模型的识别效果好，但是双模态生物特征识别不止需要人脸部分，还需要融合语音特征。本文人脸特征提取模型得到的特征要小于 Vgg_Face 的 4096 维特征，在后续的融合识别模型中训练速度会更快。并且本文的改进型模型提取出来的特征在第五章中，可以很好地和语音特征相融合。在开集测试中，识别效果要优于 Vgg_Face 提取出来的特征。

## 3.5 本章小结

在本章中，首先介绍了基于 CNN 的人脸特征提取所需要的数据预处理方法。包括基于 Haar-Like 的 AdaBoost 人脸检测算法，此算法因其实现简单，识别速度快而广泛应用于工业和学术研究中。然后介绍了 SDM 人脸特征点标定算法，人脸特征点标定是人脸对齐的关键步骤。此算法通过每次对训练样本得到参数的收敛步长，很好地解决了优化函数计算量巨大的问题。在本次试验中取得了很优异的实际效果。

在提出本文使用的人脸图像预处理技术之后，介绍了目前人脸识别效果较为优秀的网络模型，并且根据本文的需求提出了相关创新点。对原有的网络模型进行修改，在全连接层部分做出了基于双模态生物特征识别需求的改进。其中改进后的识别结果见第 5 章。在不结合语音特征的情况下，改进模型在人脸识别部分相较于 Vgg_Face 网络模型在 LFW 人脸库上降低了 2.3 个百分点。这个识别率的降低有三个部分的原因：1.增加了一个全连接层导致网络复杂度增加，使得最终网络模型收敛效果没有 Vgg_Face 好；2. 由于使用的人脸数据库不同，本文使用了人脸预处理方法包括人脸检测和人脸对齐，Vgg_Face 对其使用的人脸数据库数据并未进行人脸图像对齐等预处理；3. 在网络模型的训练过程中，调参技术没有 Vgg_Face 的优秀，调参对于深度神经网络的训练至关重

要。

　　最后使用训练好的模型对数据样本进行特征提取，对整个网络结构中的一部分层的输出特征进行可视化，并对其出现的原因进行了解释。帮助读者更加形象的理解基于卷积神经网络的人脸识别特征提取的过程。

# 第四章 说话人特征提取

本章主要提出本文的说话人特征提取系统。说话人识别系统研究对象为语音信息，在系统结构中首先提出了语音信息的预处理方法，然后对处理过后的语音进行特征提取，其中包括 1.GMM-UBM 提取说话人 I-Vector 特征；2. 提取说话人语音预测线性系数（Perceptual Linear Predictive，PLP）。最后将 I-Vector 和 PLP 特征融合送入深度神经网络中训练模型，并用此模型验证该特征对于说话人识别的效果。特征提取系统框图如下：

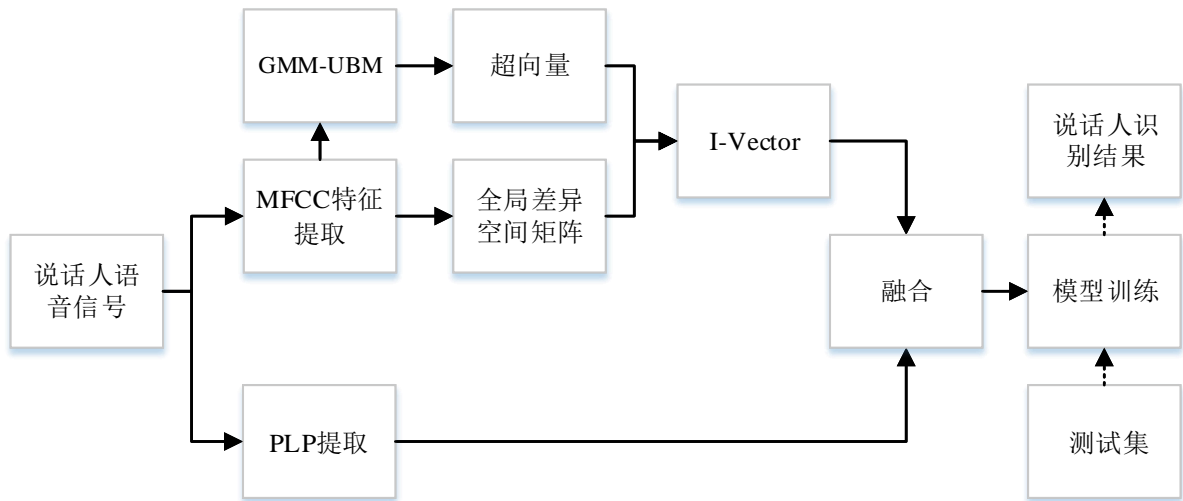

图 4.1 说话人识别系统框图

说话人识别的原理可以从一段语音产生的机理来分析。首先人耳来判断一句话是不是某个人说出来的，需要听到含有此人特性的语音信号。由此可以得出某段语音信号中必定包含此人的说话特性。一段语音产生需要经过声带的振动，然后穿过喉咙、口腔和空气、耳道、鼓膜、耳蜗等部分，最后被听觉神经接受送入大脑分析。对于机器来说，传感器接收到语音信号相当于听觉神经接受到语音信号。当然，由于主观对声音的感知不同，机器接收到的语音信号和人耳听到的信号是不一样的。那么通过机器做说话人识别，需要对声带振动、喉咙、口腔和空气、耳道、鼓膜、耳蜗这部分进行建模，然后去掉空气、耳道、鼓膜、耳蜗对信号产生的影响。模型剩下的参数即可以用来区分每个说话人[54][55]。

## 4.1 说话人语音信号 PLP 特征提取

感知线性预测算法中，使用了很多已知的人耳听觉特性。例如人耳主观声响感知曲线和掩蔽效应。由于人耳的特殊构造，人耳主观听觉感受到的响度和语音信号的频率有关，并不是所有频率上的响度感知都是一样的，通过实验可以统计出人耳对语音响度感

知随着频率的变化特性如图 4-2。可以看到在声强较低的时候，人耳对语音信号的响度感知受频率影响较大，而高声强的时候，声响主观感受随频率变化较小[56]。

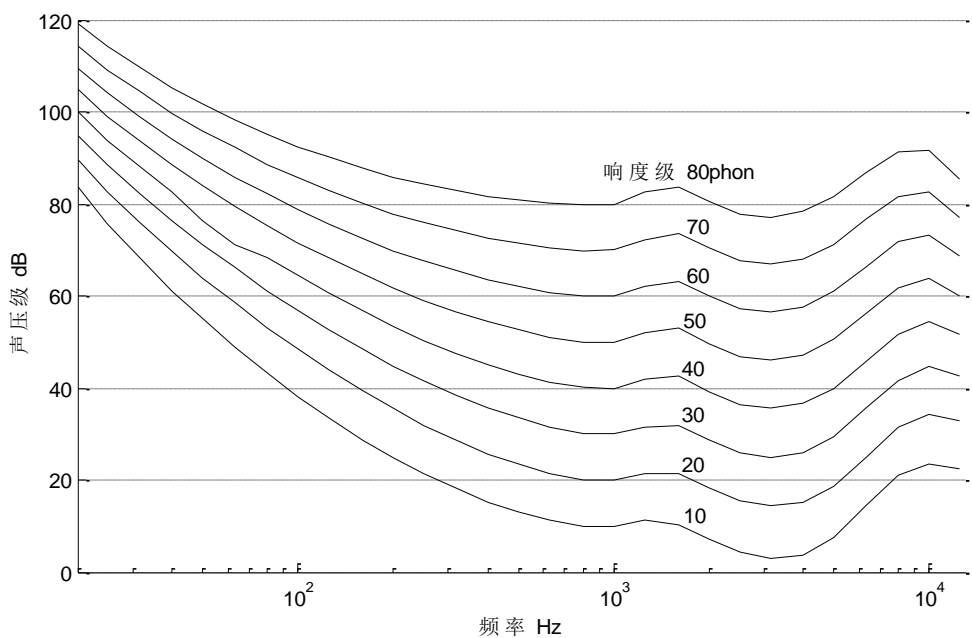

图 4.2 人耳听觉等响度图

人耳除了有等响度效应外，还具有声音掩蔽效应。掩蔽效应分为频域掩蔽和时域掩蔽，本文讨论的是频域掩蔽。即在同一时刻，当人耳接收到不同频率并且在相应频率上声音响度不同时，声强较高的频率信号会掩蔽掉其掩蔽阈值内其他频率信号的信息。其原理图如下图。

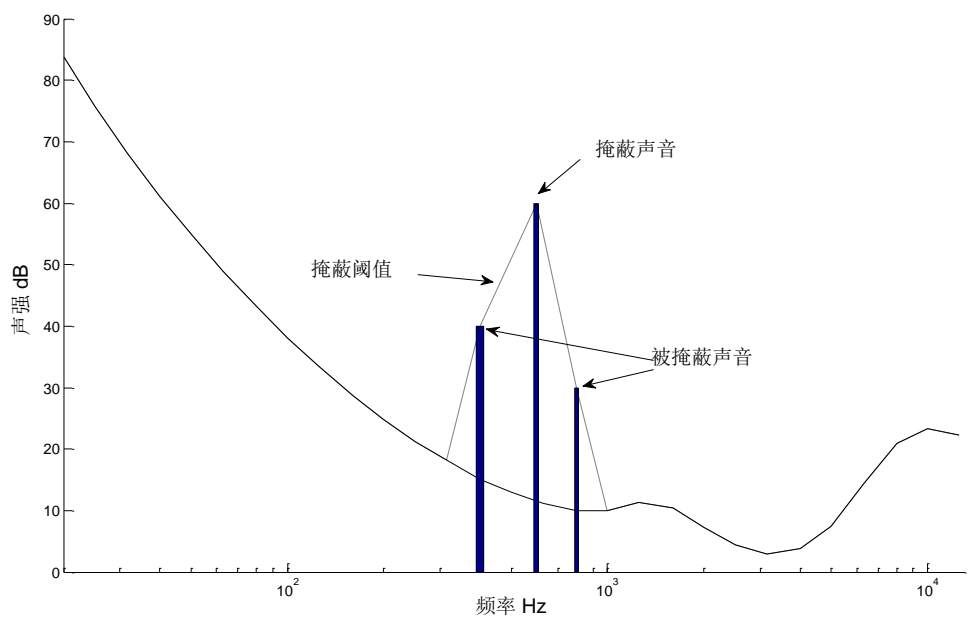

图 4.3 人耳频域掩蔽效应示意图

假设将从声带发生到听觉神经接受语音信号之间的一系列过程设置成一个黑箱，此黑箱定义为 $v$，声音激励为 $e$，接收到的语音信号为 $s$。那么可以模型化此过程为：

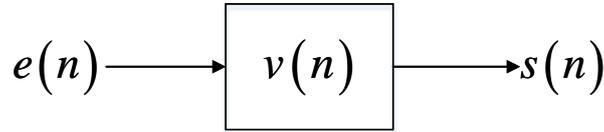

图 4.4 语音信号传输示意图

对于语音信号 $s(n)$，由声门脉冲激励 $e(n)$ 和声道响应 $v(n)$ 滤波得到。即：

$$s(n) = e(n) * v(n)$$ 4-1

对于信号的倒谱变换，上式可以变成：

$$\hat{s}(n) = \hat{e}(n) + \hat{v}(n)$$ 4-2

在倒谱域里，声道响应 $\hat{v}(n)$ 随着 增大会急速衰减。例如在采样率为10kHz 的时候，在间隔[-25 25]外的幅值衰减的很小，在此范围外的值可以忽略不计。但是对于发出浊音的时候，人类的基音周期在 10kHz 采样率下 $n$ 的变化范围在[25 200]之间。可以看出在倒谱域内基音信息和声道信息可以区分开来。采取相应的倒谱算法就可以将 $e(n)$ 和 $v(n)$ 分离并且分别恢复出来。

根据图 4-4，假设已知信号 $s(n)$ 和激励 $e(n)$，那么对于未知系统来说可以设定一个全极点模型来表示 $v(n)$：

$$v(z) = \frac{1}{1 + \sum_{i=1}^{p} a_i z^{-i}}$$ 4-3

PLP 使用的是基于参数的线性预测算法求得全极点系数。感知线性预测结合了人耳的听觉模型，通过临界频带分析、等响度预加重等过程，最终预测出语音信号的全极点模型参数[57]。PLP 系统框图如下：

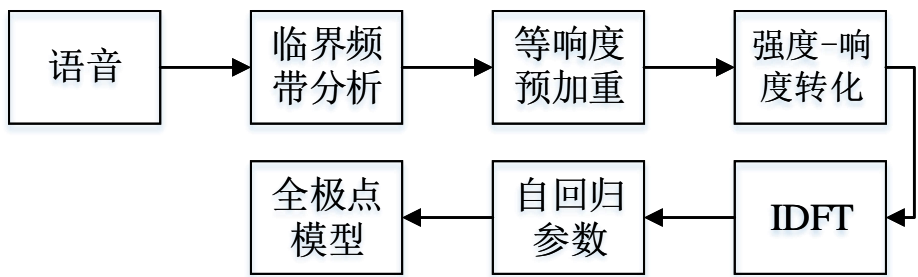

图 4.5 PLP 求解流程图

求解 PLP 的具体过程如下：

（一）临界频带分析

首先给语音信号加汉明窗：$w(n) = 0.54 + 0.46\cos\left[2\pi n / (N-1)\right]$，其中 $N$ 为窗长。窗长设置为 20ms。然后对加窗信号使用 FFT，变换到频域。则信号的短时功率谱可以

通过其频域的实部和虚部构成：

$$p(w) = \text{Re}\big[S(w)\big]^2 + \text{Im}\big[S(w)\big]^2 \qquad 4\text{-}4$$

然后将功率谱 $p(w)$ 从线性频率轴映射到人耳感知 Bark 域 $\Omega$：

$$\Omega(w) = 6\ln\left\{ w/1200\pi + \big[(w/1200\pi)^2 + 1\big]^{0.5} \right\} \qquad 4\text{-}5$$

其中 $w$ 是角频率（rad/s），在 Bark 域中，每一 Bark-Hz 为 1rad/s。功率谱映射到角频率之后，其将与临界带掩蔽曲线 $\mu(\Omega)$ 卷积。临界带曲线为：

$$\mu(\Omega) = \begin{cases} 0 & \Omega < -1.3 \\ 10^{2.5(\Omega + 0.5)} & -1.3 \le \Omega \le -0.5 \\ 1 & -0.5 < \Omega < 0.5 \\ 10^{0.5 - \Omega} & 0.5 \le \Omega \le 2.5 \\ 0 & \Omega > 2.5 \end{cases} \qquad 4\text{-}6$$

上式表示的模拟临界带掩蔽曲线和功率谱离散卷积之后产生临界带功率谱：

$$\beta(\Omega) = \sum_{\Omega = -1.3}^{2.5} p(\Omega - \Omega_i)\mu(\Omega) \qquad 4\text{-}7$$

由于临界带掩蔽曲线的截断效应可以有效地降低功率谱的分辨率。然后对其进行 1rad/s 间隔采样，得到 $\nu(\Omega)$。

(二) 等响度预加重

可以从图 4-3 看出来，人耳对于中高频相对敏感，根据研究表明，中耳道可以对 1-5KHz 的语音信号响度提示 10~20dB。为体现人耳对声音信号的这一特性，对功率谱进行等响预加重。对采样之后的功率谱进行等响度曲线预加重：

$$\Phi(\Omega(w)) = \text{E}(w)\big[\Omega(w)\big] \qquad 4\text{-}8$$

其中函数 $\text{E}(w)$ 为近似的人耳对不同频率的灵敏度，其灵敏度控制在 40dB。可以具体化为：

$$\text{E}(w) = \frac{(w^2 + 56.8 \times 10^6)w^4}{(w^2 + 6.3 \times 10^6)^2 \times (w^2 + 0.38 \times 10^9)(w^6 + 9.58 \times 10^{26})} \qquad 4\text{-}9$$

(三) 强度-响度转化

对功率谱进行采样和预加重之后需要经过强度-响度转化过程，这一步骤是为了模拟人耳听觉曲线也是为了降低临界带功率谱幅值的变化。其变化公式为：

$$\Theta(\Omega) = \Phi(\Omega)^{0.33} \qquad 4\text{-}10$$

接收到的语音信号强度和人耳感知响度符合上式的非线性关系。

(四) 全极点模型逼近

对 $\Theta(\Omega)$ 进行 IDFT 之后可以得到语音信号的短时自相关函数 $\gamma(\tau)$，然后使用全极

点模型来逼近 $\gamma(\tau)$，即做线性预测分析。使用 Levinson-Durbin 递归法求得 阶线性预测系数。通过 12 阶的全极点模型逼近人耳听觉谱，对声音信号提取出 24 维 PLP 特征向量。

高阶的感知线性预测系数中包含了语音内容信息和说话人信息，对于不基于文本的说话人时别来说可以抑制语音内容信息的特征，此抑制过程可以交由后续的深度神经网络完成。

## 4.2 说话人语音 I-Vector 特征提取

最近几年文本无关说话人识别中，广泛使用的 I-Vector 参数特征由 Dehak N 在 2010 年提出。此前广泛使用的联合因子分析（JFA）算法通过构建说话人信息子空间和语音信道子空间进行失配信道补偿来提取说话人识别特征，并且取得了比较好的效果。但是研究表明，JFA 算法中的语音信道子空间同样可以提取出说话人特有的特征信息。但是 JFA 算法求说话人均值矢量的时候会损失一些说话人信息[58]。I-Vector 正是基于此理论提出的，使用全变量信息子空间进行估计。其中全变量信息子空间是一个包括说话人信息和语音信道信息的矩阵。使用它估计的 I-Vector 特征包含了说话人信息和语音信道信息。

如图 4-1 所示，提取语音信号 I-Vector 特征需要经过 MFCC 参数提取、GMM-UBM 训练、超向量提取和求全局差异空间矩阵几个步骤。下面将对这几个部分作进一步分析。

### 4.2.1 语音信号 MFCC 提取

根据式子 4-2 可以知道，在倒谱域上可以将激励脉冲信号和声道信息很好地区分出来。通过对人耳耳蜗的频率响应特性进行分析，人耳在 Mel 频率尺度上对语音频率呈线性感知特性。不同于 PLP 对全极点模型参数的估计，MFCC 的提取过程需要将功率谱变化到 Mel 频率[59]。具体过程如下：

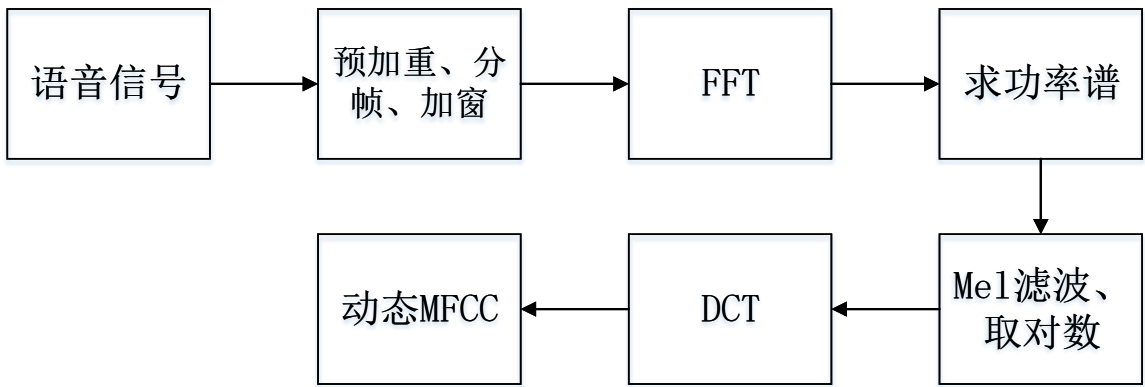

图 4.6 MFCC 求解框图

首先对语音信号预加重，处理公式为：$s'(n) = s(n) - a \cdot s(n-1)$　其中 $a \in [0.9,1]$。然

后对其分帧处理，分帧间隔大概为 25ms 左右，帧叠为 5ms。分帧之后给语音信号加汉明窗：$w(n)=0.54+0.46\cos[2\pi n/(N-1)]$。为了推导方便，现假设分帧加窗之后的语音信号为 $s_i(n)$。对其求 FFT 后得到：

$$s_i(z) = e_i(z)v_i(z) \tag{4-11}$$

根据式子 4-4 对每一帧信号求其功率谱得到 $X_n(k)$。

Mel 谱和实际频率的关系为：

$$\text{Mel}(f) = 2595 \lg\left(1+\frac{f}{700}\right) \tag{4-12}$$

通过频率变换可以模拟人耳在 Mel 谱中的线性感知特性。人耳在此频率上具有非常优秀的性能，可以提取出语义信息，说话人特征。并且人耳具有滤波器的特性，可以将整个频带中的信号分段通过，临界频率带宽随着频率变化而变化，并且和 Mel 谱的增长保持一致[60]。此滤波器可以近似为一系列三角滤波器，其在 Mel 谱上为线性分布，而在实际频谱上分布如下：

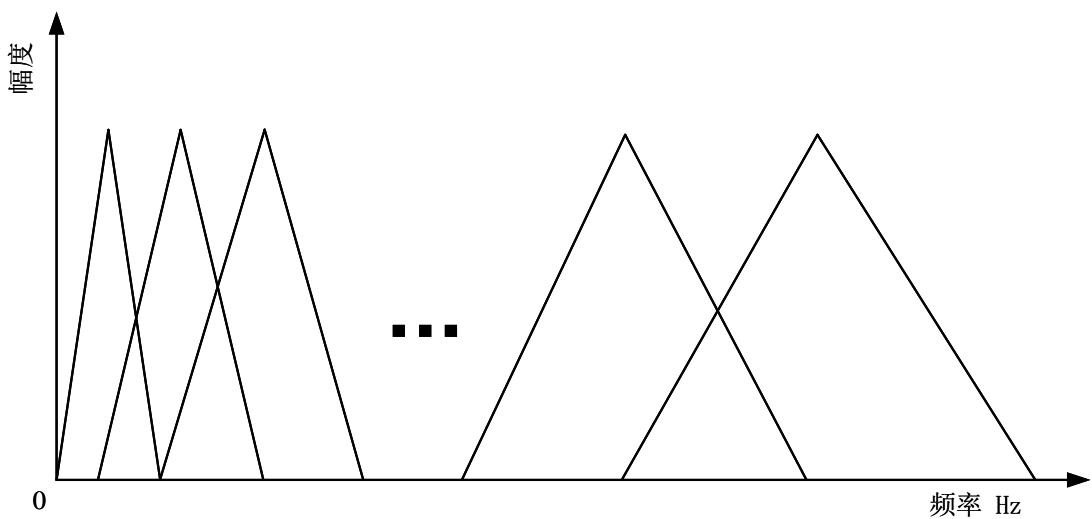

图 4.7 Mel 频率尺度滤波器组示意图

其中每个三角滤波器的中心频率在 Mel 谱上是等间隔分布的，记为 $c(l)$。假设和 $h(l)$ 是该滤波器的下限和上限。滤波器之间存在如下关系：

$$c(l) = h(l-1) = o(l+1) \tag{4-13}$$

将功率谱 $X_n(k)$ 通过上图所示的滤波器组后可以得到输出为：

$$m(l) = \sum_{k=o(l)}^{h(l)} W_l(k)|X_n(k)| \quad l=1,2\cdots,L \tag{4-14}$$

其中：

$$W_i(k) = \begin{cases} \dfrac{k - o(l)}{c(l) - o(l)} & o(l) \leq k \leq c(l) \\[2mm] \dfrac{h(l) - k}{h(l) - c(l)} & c(l) \leq k \leq h(l) \end{cases} \qquad \text{4-15}$$

经过三角滤波器组之后得到的信号不包含原有语音信号存在的音调和音高对说话人识别的影响因素。经过三角滤波器之后可以对频谱进行平滑，消除谐波，而且可以降低算法的运算量。

在获得 Mel 滤波之后，对滤波器输出的进行对数运算，然后进一步进行 DCT，就可以得到 MFCC：

$$c_{mfcc}(i) = \sqrt{\frac{2}{N}} \sum_{l=1}^{L} \lg\big(m(l)\big) \cos\left\{\left(l - \frac{1}{2}\right)\frac{i\pi}{L}\right\} \qquad \text{4-16}$$

其中 M 为滤波器个数，L 为 MFCC 系数阶数，通常取 12~16。本文在求语音信号的 I-Vector 特征的时候取 L=13。

由此方法得到的 MFCC 系数反应的是语音信号的静态特性，但是说话人特征包含一个人说话的持续过程。为了获得语音信号的动态特性，需要求 MFCC 的一阶差分和二阶差分：

$$d_t = \begin{cases} c_{t+1} - c_t & t < K \\[2mm] \dfrac{\displaystyle\sum_{k=1}^{K} k\left(c_{t+k} - c_{t-k}\right)}{\sqrt{2\displaystyle\sum_{k=1}^{K} k^2}} & \text{其他} \\[2mm] c_t - c_{t-1} & t \geq L - K \end{cases} \qquad \text{4-17}$$

其中 $d_t$ 表示第 $t$ 个一阶差分，$c_t$ 表示第 $t$ 个倒谱系数。$L$ 表示 MFCC 阶数，而 $K$ 表示一阶差分的时间差，可以为 1 或者 2。本文通过上式求得完整的 MFCC 系数为 MFCC 系数加上其一阶差分、二阶差分参数和信号当前帧能量。

### 4.2.2 GMM- UBM 训练

高斯混合模型（GMM）是隐马尔可夫模型（HMM）的一个状态时的形态。常用于拟合说话人系统中的声学模型参数分布概率。鉴于说话人声学特征参数是由不同时刻的声学特征和不同的特征空间的参数的集合。因此多个高数概率模型可以很完美的拟合这个任意分布的声学特征参数分布概率。

假设有一个 M 阶的 GMM，其可以通过下式表示：

$$p(x \mid \lambda) = \sum_{i=1}^{M} p(x, i \mid \lambda) = \sum_{i=1}^{M} w_i \, p(x \mid i, \lambda) \qquad \text{4-18}$$

其中 $x$ 表示声学特征、$i$ 是高数分量的序号、$\lambda$ 表示 GMM 模型、$M$ 表示高斯密度函数的个数。$w_i$ 表示第 $i$ 个高斯函数的权值，也就是其对应第 $i$ 个函数的先验概率。满足总和为 1。其中每个子分布复合 K 维联合高斯概率分布：

$$b_i(x) = p(x|i,\lambda) = \frac{1}{(2\pi)^{K/2}|\Sigma_i|^{1/2}} \exp\left\{-\frac{1}{2}(x-\mu_i)'\Sigma_i^{-1}(x-\mu_i)\right\} \qquad 4\text{-}19$$

上式中 $\mu_i$ 是均值向量，$\Sigma_i$ 是协方差函数。GMM 参数集合 $\lambda$ 可以由均值参数、协方差矩阵和高斯函数权重三个元素构成：

$$\lambda = \{w_i, \mu_i, \Sigma_i\} \qquad \text{其中} i = 1,2,\cdots,M \qquad 4\text{-}20$$

均值 $\mu_i$ 决定高斯函数的位置，方差决定高斯函数的分布，权值 $w_i$ 决定了高斯函数的幅值。对于任意分布，都可以找到相应的均值、方差和权重来拟合。

GMM 模型通过 MFCC 进行训练，最常用的方法是 EM（Expectation Maximization）算法估计参数 $\lambda$。EM 算法首先设置　的值为初始值开始，然后采用 EM 算法估计出新的参数 $\hat{\lambda}$，保证新的参数下似然度 $P(x/\hat{\lambda}) \geq P(x/\lambda)$。然后使用该参数作为新一轮训练参数使用，直到整个模型收敛。但是实际中这种算法需要大量的训练数据，少样本的情况下不能较好的拟合出模型。D.A.Reynolds 提出了 UBM 概念，可以解决训练数据不充分，信道环境失配的问题[61]。

对于一段说话人语音判断其是否是某个人说出来的可以由下式判断：

$$\sigma = \frac{p(v|g_0)}{p(v|g_1)} \qquad \text{其中}\begin{cases} \sigma \geq \theta \ \text{取} g_0 \\ \sigma < \theta \ \text{取} g_1 \end{cases} \qquad 4\text{-}21$$

其中 $g_0$ 表示 $v$ 是目标人物，$g_1$ 表示 $v$ 不是目标人物。$p(v|g_i)$ 表示似然函数，$\theta$ 为判决门限。套用上面的 GMM 模型来说，对于目标人物。如果假设为 $g_0$，则目标人物的 GMM 参数为 $\lambda_0$；如果假设为 $g_1$，其 GMM 参数为 $\lambda_1$。可以将 4-21 式改成：

$$S(X) = \log(\sigma) = \log\left(\frac{p(X|\lambda_0)}{p(X|\lambda_1)}\right) = \log(p(X|\lambda_0)) - \log(p(X|\lambda_1)) \qquad 4\text{-}22$$

其中 X 表示语音信号 MFCC 参数矢量。通过上式可以看出如果实际上是 $g_0$ 成立，则语音信号来自目标说话人，那么通过对目标人物的注册训练可以得到精确的估计。但是如果实际上是 $g_1$ 成立，对于 $\lambda_1$ 来说，其表示所有非目标人的特征空间。对于所有非目标人的特征空间使用 UBM 去处理。UBM 中包含人类语音信息，环境噪声，信道噪声等。对 UBM 进行充分的训练生成高斯混合模型，训练的越充分则非目标人特征因素越充分，区分效果越好[62]。

通过训练集语音训练好 UBM，当测试集语音特征送进了的时候，可以通过 MAP

自适应算法得到测试语音的 GMM。假设送入的测试语音特征为 $O = \{o_1, o_2, \cdots o_L\}$，则其与 UBM 的输出似然概率为：

$$p(i \mid O) = \frac{w_i b_i(O)}{\sum_{j=1}^{M} w_j b_j(O)} \quad \text{其中} i = 1, 2, \cdots M \qquad 4\text{-}23$$

然后使用 $p(i \mid O)$ 和 $O$ 计算新的 GMM 权值，均值和方差：

$$w_i' = \sum_{l=1}^{L} p(i \mid o_l) \qquad 4\text{-}24$$

$$E_i(o) = \frac{1}{w_i'} \sum_{l=1}^{L} p(i \mid o_l) o_l \qquad 4\text{-}25$$

$$E_i(o^2) = \frac{1}{w_i'} \sum_{l=1}^{L} p(i \mid o_l) o_l^2 \qquad 4\text{-}26$$

最后得到新的自适应的目标人物语音模型的权值、均值和方差由下式得到：

$$\hat{w}_i = [\frac{a_i^w w_i'}{L} + (1 - a_i^w) w_i] \gamma \qquad 4\text{-}27$$

$$\hat{\mu}_i = a_i^m E_i(o) + (1 - a_i^m) \mu_i \qquad 4\text{-}28$$

$$\hat{\Sigma}_i = a_i^d E_i(o^2) + (1 - a_i^d)(\Sigma_i^2 + \mu_i^2) - \hat{\mu}_i^2 \qquad 4\text{-}29$$

式中，$\gamma$ 为保证权值 能维持 $\sum_{i=1}^{M} \hat{w}_i = 1$ 而设立的权值规整化因子。$a_i^w$、$a_i^m$、$a_i^d$ 是用来控制第 i 个高斯分量的旧参数更新到新参数的步进尺度。也称为权值，均值和方差的调整因子[63]。

UBM 使用大量说话人语音进行训练生成一个高阶高斯模型，其中包含了大量说话人信息。把 GMM 和 UBM 结合起来之后，GMM 可以提现说话人的特有信息，UBM 表示大量说话人的共有特性。要体现某一个说话人的特征信息，通过式子 4-24、4-25 和 4-26 对 UBM 进行最大后验概率自适应训练。训练的过程也是对 UBM 得到的高斯函数修正的过程。训练 UBM 的时候训练样本越多，得到的模型包含的信道信息越全面，包含的说话人数越多，泛华性能也就会越好[64]。

### 4.2.3 全局差异空间估计

在训练全局差异空间的时候，假定所有的数据来自不同的说话人。训练差异空间 T 矩阵，首先计算每段语音的 Baum-Welch 统计量。假设该段语音每帧特征量为 $X = (x_1, \cdots x_L)$，对应的说话人为 s，$\lambda$ 表示 $M$ 阶的 UBM 模型。首先计算两个统计量：

$$N_{j,h}(s) = \sum_{i=1}^{L} p(j \mid x_i, x_i) \qquad 4\text{-}30$$

$$F_{j,h}(s) = \sum_{i=1}^{L} p(j \mid x_i, \lambda) x_i \qquad 4\text{-}31$$

其中 $j \in [1,M]$ 是 UBM 高斯混合模型中高斯函数的序号。 $p(j \mid X_i, \lambda)$ 是第 $X$ 帧语音对于第 $j$ 个高斯函数的后验概率。定义 Baum-Welch 的一阶中心统计量为：

$$\hat{F}_{j,h}(s) = \sum_{l=1}^{L} p(j \mid X_i, \lambda)(X_i - m_j)$$
$$= F_{j,h}(s) - N_{j,h}(s)m_j \qquad \text{4-32}$$

为后续计算方便，把统计量以矩阵的形式表示：

$$NN(s) = \begin{pmatrix} N_1(s) & & 0 \\ & \cdots & \\ 0 & & N_m(s) \end{pmatrix} \qquad \text{4-33}$$

$$FF(s) = \begin{pmatrix} \tilde{F}_1(s) \\ \cdots \\ \tilde{F}_M(s) \end{pmatrix} \qquad \text{4-34}$$

得到统计量之后，使用 EM 算法对全局差异空间矩阵 T 进行训练求解：

E（Expectation）步骤：首先要对 T 矩阵进行随机初始化，然后计算说话人因子的方差和均值：

$$l_T(s) = I + T^T \sum{}^{-1} NN(s)T \qquad \text{4-35}$$

$$\bar{y}(s) = l_T^{-1}(s)T^T \sum{}^{-1} FF(s) \qquad \text{4-36}$$

M（Maximization）步骤：训练所有的语音统计量来求最大似然估计：

$$N_c = \sum_s N_c(s) \qquad \text{4-37}$$

$$A_c = \sum_s N_c(s)l_T^{-1}(s) \qquad \text{4-38}$$

$$C = \sum_s FF(s)\bar{y}(s)^T \qquad \text{4-39}$$

通过训练所有的语音数据之后得到上述统计量，然后用它们更新全局差异矩阵：

$$V = \begin{pmatrix} V_1 \\ \vdots \\ V_c \end{pmatrix} = \begin{pmatrix} A_1^{-1}C_1^T \\ \vdots \\ A_C^{-1}C_C^T \end{pmatrix} \qquad \text{4-40}$$

在 EM 算法迭代开始之前设置初始次数，要是没有达到设定的次数，再次进行 E 步骤和 M 步骤，直到完成设定的迭代次数[65]。通常设置迭代次数为 10 次左右。

### 4.2.4 I-Vector 特征提取

语音信号的 I-Vector 特征提取公式思想如下：

$$M = m + Tw \qquad \text{4-41}$$

其中 M 为目标说话人的 GMM 均值超矢量、T 是上小节所求的全局差异空间矩阵，

是一个低秩矩阵，其可以说话人信息和空间信道信息的空间分布；$m$ 为信道与说话人无关的超矢量，即 UBM 的均值矢量；$w$ 是服从标准正态分布的随即向量，即 I-Vector。

## 4.3 说话人识别深度神经网络结构

在前面两节对说话人语音提取出 PLP 和 I-Vector 特征，本节使用深度神经网络结合这两个特征训练说话人识别模型[66]。并通过此模型的识别率进一步验证这两种特征融合之后对说话人识别的效果。在说话人识别部分使用的深度神经网络为 DBN[67]。此神经网络的结构如下图：

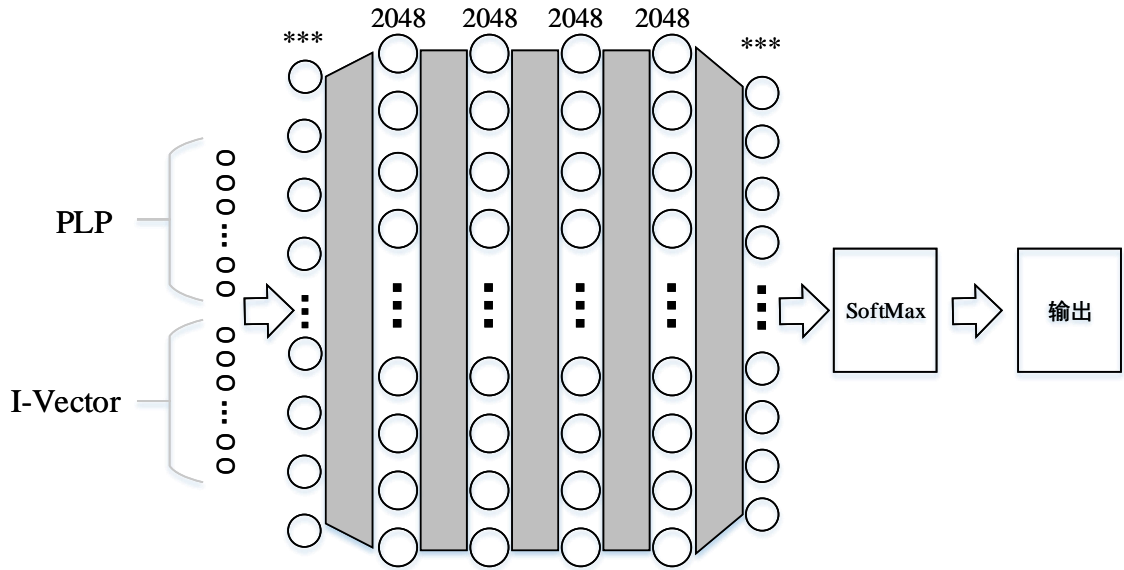

图 4.8 说话人识别深度神经网络结构

本文使用了 5 隐层的 DBN 网络结构，其中第一层为输入数据，中间的隐层节点数为 2048。最后一层的维度为需要分类的语音类别总数。最后将最后一层的特征送入 SoftMax 进行分类。

## 4.4 说话人识别模型数据库

本文说话人识别使用的数据库包括 TED_LIUM、NIST04、NIST05、NIST06、NIST07。其中 NIST 系列数据库是 NIST SRE 从 2004 年到 2007 年的说话人识别竞赛提供的语音数据。其中 NIST 04 为采样率为 8K，8bit 量化，$\mu$ 率压缩的 sph 文件。包含 616 个说话人，男性有 246 个，女性有 370 个。05~07 年的数据相对于 04 年在声音采集方式，环境噪声等方面有一定的出入，但大致上保持一致。对于 I-Vector 特征提取部分，UBM 的训练使用的是 NIST04 和 NIST05 的数据库。全局差异空间矩阵 T 使用的是 NIST 06 和 NIST 07 数据集。

TED-LIUM 数据库由缅因州大学计算机系创建。数据库中包含的内容为 TED 演讲

的音频数据。由 1242 个人演讲的 1495 个音频文件，男性包含 141 个小时，女性包含 66 个小时。音频采样率为 16kHZ。根据实验的需求，本文针对该数据库进行了相应处理。根据 1242 个人的音频语料，将每个人的说话时长分割成 5s、10s、15s 和 20s。每种长度提供 25 个样本。从 1242 个人中随机抽取 1000 人分入训练集（训练样本和训练过程中的测试样本），另外的 242 个人分入测试集。对于每个语音样本，首先通过 4.1 节中的算法得到其每帧数据的 PLP 系数，然后对 PLP 系数进行 PCA 降维。再提取该样本的 I-Vector 特征。将降维后的 PLP 系数和 I-Vector 特征拼接成完整的说话人语音特征。具体数据分布如下表：

表 6 说话人识别模型数据库

| 时长/秒 集合类别 | 5 | 10 | 15 | 20 |
|---|---|---|---|---|
| 训练集 | 20×1000 | 20×1000 | 20×1000 | 20×1000 |
| 训练测试集 | 5×1000 | 5×1000 | 5×1000 | 5×1000 |
| 测试集 | 25×242 | 25×242 | 25×242 | 25×242 |

## 4.5 说话人识别模型训练

本文使用的 DBN 的结构由 5 个 RBM 和一个 SoftMax 层组成。其中隐层的节点数为：2048×2048×2048×2048。在网络预训练时，网络最底层的受限玻尔兹曼机为高斯-伯努利类型，用来对输入的语音特征进行建模，并且对所有训练样本进行均值方差归一化处理。第一层 RBM 训练结束后，将其输出作为下一个 RBM 的输入，采用贪婪算法依次逐层向上训练第二层、第三层和第四层的参数，这里的受限玻尔兹曼机为伯努利-伯努利型。在预训练结束后，得到的 DBN 网络最上层增加 softmax 输出层，此层的网络权重通过随机初始化得到。然后用 2.12 节介绍的误差反向传播的 BP 算法，通过输出层的估计值和输入样本标签之间的交叉熵为目标函数对整个网络进行有监督的参数调节。

本文使用了 4 种长度的语音数据对网络进行了训练，语音帧长为 25ms，帧叠为 5ms，每帧提取 PLP 系数为 13 维。最后对 PLP 特征进行 PCA 降维到 500 维。拼接上 I-Vector 特征，总共语音特征长度为 1100 维。训练过程中学习速率为 0.005 并使用阶段性衰减调节学习了，衰减系数为 0.95，衰减步进长度为 2000，最大迭代次数为 40000。最后实验的识别率和目标函数损失曲线图如下：

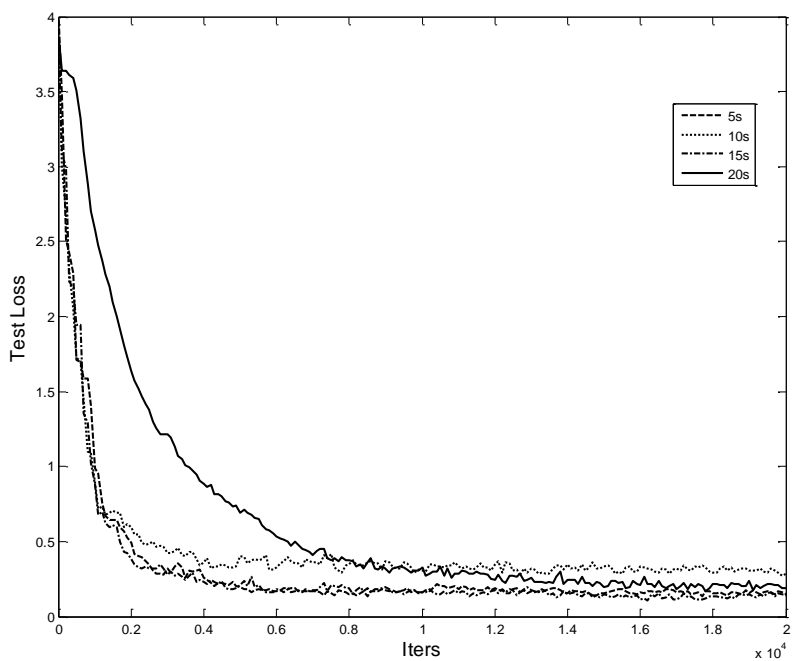

图 4.9 测试集损失函数值随迭代次数变化曲线

从上图可以看出，在模型训练过程中，5s、10s 和 15s 长的语音特征训练网络模型时，迭代次数在 200~2000 这一段时，测试损失函数下降很快，从 2000 次之后，测试集损失值趋于变缓。20s 长的语音特征训练网络的时候，迭代次数在 600~6000 这段区间内，测试集损失函数值下降较快。可以看出 20s 相比于前 3 种语音长度的网络训练过程，测试集的损失函数值下降较为缓慢。并且 4 种语音长度最终的损失函数值依次稳定在 0.12、0.27、0.14、0.18。测试集的损失函数反应的是深度神经网络模型在训练过程中，对其闭集内的测试数据的错误值。也是反应训练模型对训练数据学习收敛的一个过程，最终的损失函数值不能反应那种长度的数据集对网络的训练好坏。只能反应该数据集合训练当前网络的速度。通过测试损失函数可以看出，目前网络模型学习的程度，当测试损失函数值趋于稳定的时候，表示该模型对此训练集合已经达到饱和的状态。继续训练下去，网络模型对该训练集合也不会有进一步的优化。

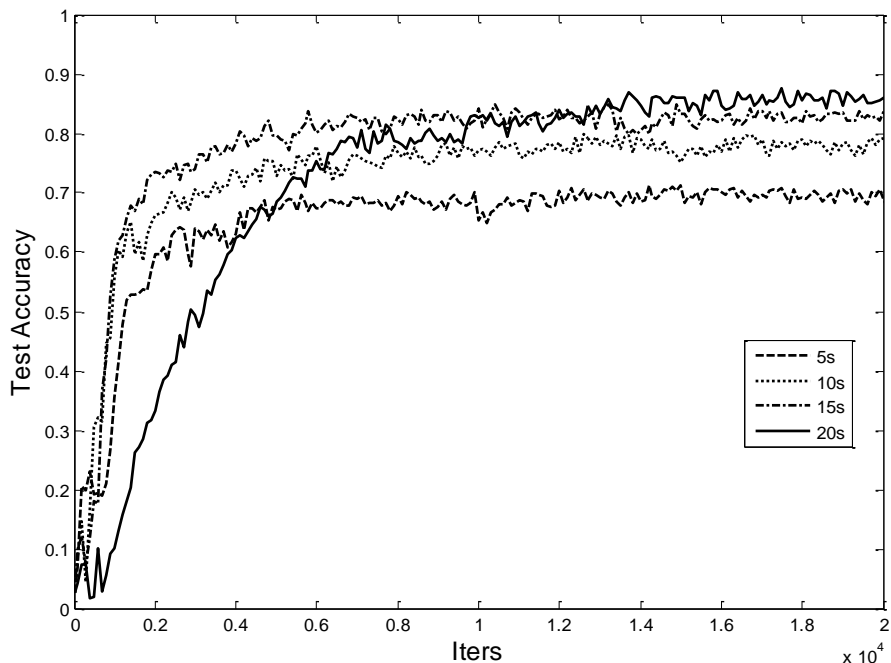

图 4.10 说话人识别模型识别率随迭代次数变化曲线

上图为模型对测试集样本的识别率随着迭代次数变化曲线。可以看到 5s、10s 和 15s 长的说话人数据训练网络的时候，符合其损失函数变化的状态，在最开始 200~2000 次迭代的过程中，识别率上升较为迅速，之后识别率增加开始放缓。最后稳定在各自的最终识别率范围内。而 20s 长的语音特征数据的准确率变化曲线，在最开始的 600 次迭代中，识别率反复变化，经过 600 次的迭代之后，开始有一个稳定的上升区间，大约在 6000 次迭代的时候，上升速度开始放缓。其识别率相较于前三种时长的语音数据，需要更多的迭代次数才能达到其最终的识别率。通过实验结果得到最终各时长的说话人数据训练得到的模型在其各自训练集中达到的识别率分别为：70.84%、79.21%、85.8%、87.76%。

用此模型在 242 个人的开集中进行测试。242 个人，每人 25 条语音，通过随机配对一共生成 1000 对正集和 1000 对负集。将两个语音分别送入模型中进行计算，得到其特征之后，使用余弦距离来判别其相似程度。归一化余弦距离之后，部分识别结果如下表：

表 7 说话人识别模型开集测试部分结果

| 阈值 | 正样本（正确/错误） | | | | 负样本（正确/错误） | | | |
|---|---|---|---|---|---|---|---|---|
| 样本类型 时长 | 5s | 10s | 15s | 20s | 5s | 10s | 15s | 20s |
| 0.4653 | 684/316 | 716/284 | 844/156 | 851/149 | 684/316 | 748/252 | 857/143 | 870/130 |
| 0.4863 | 698/302 | 728/272 | 853/147 | 862/138 | 667/333 | 728/272 | 840/160 | 855/145 |
| 0.4781 | 692/308 | 723/277 | 849/151 | 853/147 | 672/328 | 738/262 | 849/151 | 870/130 |
| 0.4840 | 695/305 | 728/272 | 852/148 | 860/140 | 670/330 | 732/268 | 845/155 | 860/140 |

表中，归一化阈值仅限时长相同的说话人语音类内对比，不同时长的归一化阈值之间没有必然联系。其中阈值（0.4653、0.4863、0.4781、0.4840）分别为（5s、10s、15s、20s）在等错误率的时候的归一化阈值。

最终得到模型的 ROC 曲线如图 4-11。可以从图中看到，在等错误率点处，各时长数据训练的模型在开集测试集上的识别率分别 68.43%、72.83%、84.90%、85.97%。开集测试集的识别率相对于闭集测试集的识别率都有一定程度的降低。总结起来，导致开集识别率没有闭集识别率高的原因有以下几点：

1. 训练集样本数量略少，导致训练的模型泛化能力还不够强；

2. 模型训练过程中，产生了一定的过拟合，导致开集测试识别率有所下降；

例如 20s 长度说话人的语音特征训练识别模型，开集测试识别率只有 85.97%，但从闭集训练的识别率曲线可以看出，PLP 系数和 I-Vector 特征融合之后，通过深度神经网络训练生成的识别模型对说话人区分具有不错的效果。但是距离实际使用还是需要更大的样本集进行训练。

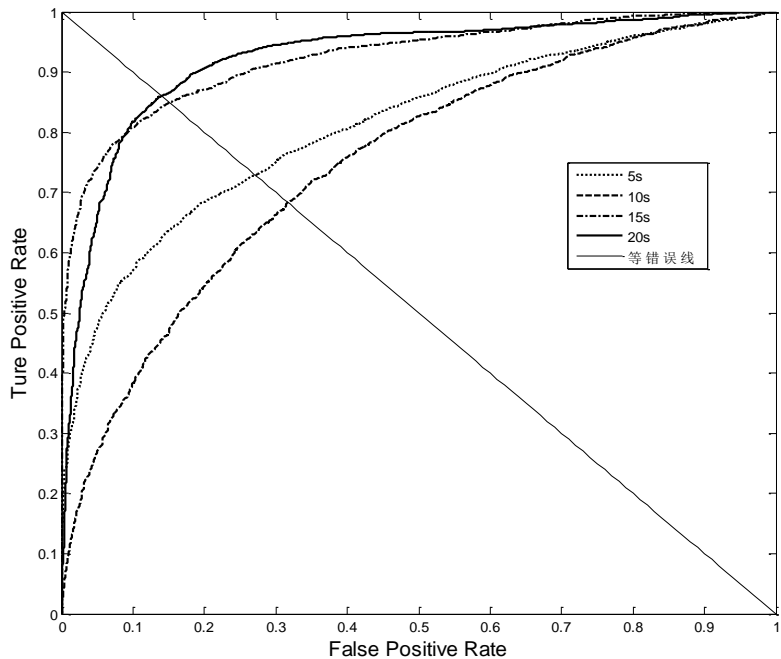

图 4.11 说话人识别模型 ROC 曲线

可以从最终的识别率可以得到，时间较长的语音数据提取出来的特征训练网络模型，具有更优秀的识别效果，实际情况也是必然如此，在一定范围内，说话人语音数据越长，其中包含的说话人信息越多。提取的 I-Vector 特征包含说话人信息相对于其他信息更加鲁棒。相同的训练时间下，越长的测试数据特征训练得到的识别模型的识别率越高。但是由于数据长度加长，特征提取所需要花费更多的时间，并且实际应用场景中，

提取不到足够长的语音数据，因此训练网络模型的输入数据时长要选择时长适中的数据。

## 4.6 本章小结

本章首先提出了说话人识别部分的语音信号的特征选取方案（基于 PLP 和 I-Vector 两种特征融合）。然后推导了感知线性预测系数的求解方法，利用人耳的听觉感知特性，通过对语音信号进行临界带分析、等响度预加重和强度-响度转化等处理之后进行全极点参数估计，最后求得 PLP 系数。对于语音信号的 I-Vector 特征，本文推导了 MFCC 特征提取方法，首先是利用 MFCC 对 GMM-UBM 进行训练得到通用背景模型，进而对语音信号进行参数估计得到超向量。然后用 MFCC 使用 EM 算法训练全局差异空间矩阵，利用超向量和全局差异空间矩阵求出语音信号的 I-Vector 特征。基于 YK_Speakers 语音库，使用 5 隐层的 DBN 对其进行训练，得到识别模型，最后取得了 87% 的识别率。通过对训练好的模型的识别率进行分析，得到 PLP 加上 I-Vector 作为说话人识别特征具有很好地性能。

# 第五章 双模态生物特征识别实验与分析

在第三章中使用卷积神经网络对人脸特征进行提取，通过卷积层和池化层的交替抽象得到基于深度神经网络的人脸特征，然后将这个无法用语言描述的特征送入全连接层再次进行训练最后通过 SoftMax 层对其进行分类，并且对训练模型进行测试，得到了不错的识别效果。然后在第四章中提出了利用说话人语音信号的 PLP 系数和 I-Vector 来做说话人识别的特征参数，并通过 DBN 网络进行验证其可靠性[68]。在本章中将会使用到经过卷积层和池化层交替抽象得到的图像特征，以及语音信号的 PLP 系数和 I-Vector 参数。通过对这两类生物特征的参数处理，并且设计其专有的深度神经网络对其训练得到识别模型[69]。

## 5.1 双模态特征融合

由于图像和语音在数据层的结构特性相差较大，本文并未使用数据层融合方式。本文使用的融合方式为特征层融合。特征层融合将多个特征集合成一个新的融合特征，然后使用深度神经网络对这个新的特征进行训练分类模型[70][71]。特征融合具有两个很重要的优势：1. 融合特征相比于单个特征而言，融合后的特征具有更多的数据信息，更加具有判别性；2. 融合特征可以得到有效、冗余少的特征，提高分类性能。特征层融合策略有并行特征融合和串行特征融合两种方式。针对于不同的实际处理对象，需要选取不同的融合方式。

假设 $A$ 和 $B$ 是样本空间内两个不同类型的的特征集合，那么对于样本空间中任意的样本而言，特征空间 $A$ 和 $B$ 的特征向量为 $a$ 和 $b$。并且假设特征的串行组合后的特征为 $c$，其中 $c$ 表达式见式子 5-1。要是 $a$ 是 $n$ 维，$b$ 是 $m$ 维特征向量，那么 $c$ 的维度为 $n+m$。那么所有样本新的融合特征向量构成维度为 $n+m$ 的向量空间。并且该空间内的特征数据为实数。

$$c = \begin{pmatrix} a \\ b \end{pmatrix} \qquad\qquad 5\text{-}1$$

若数据进行并行组合策略，则构成复数向量空间。同样构成样本的两类特征向量空间为 $A$ 和 $B$，其中的特征向量为 $a$ 和 $b$。并行的特征向量 $c = a + i \cdot b$。假如两个特征向量的维度不等，则将低维度的特征向量结尾补齐再进行特征融合。由于本文对图像特征和语音特征进行融合，使用并行融合方式不具有其物理含义，并且本文设计的的深度神经网络也不支持对复数的运算，因此选用串行组合特征的策略。通过训练好的卷积神

网络模型对图像进行特征提取，然后和语音信号的 PLP（PCA 降维）系数和 I-Vector 特征串行融合生成新的融合特征[72]。具体的融合流程如下图：

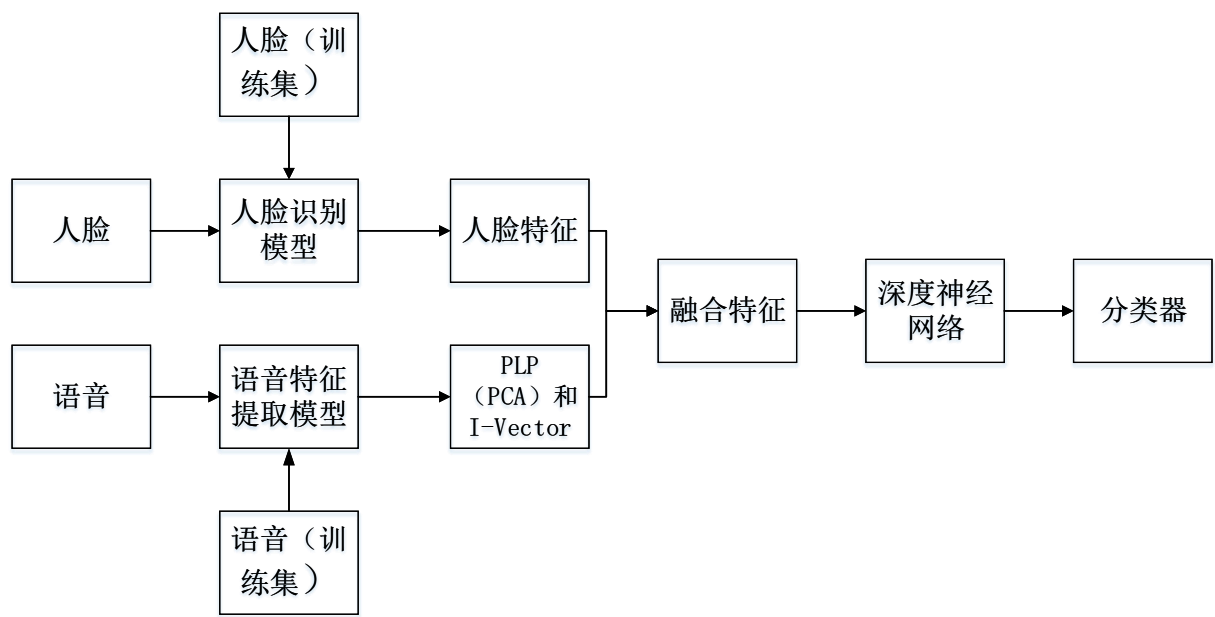

图 5.1 特征层融合流程图

## 5.2 数据库构建

本文使用的数据库为自建库，目前适用于深度神经网络的人脸和说话人双模态融合识别还没有较好的数据库。之前的融合数据库由于数据样本较少，无法满足深度神经网络的训练需求。在此将 CASIA-WebFace-B 库和 TED_LIUM 数据库进行组合，创建一个融合数据库。实际录入一个人脸语音混合库，每个人的人脸和其语音是未知的。通过人脸库和语音库新建一个人脸和说话人识别的混合库，随机选取人脸和语音样本进行匹配。这样生成的混合库样本分布符合实际录入混合库中的人脸和语音样本的分布情况。

在第三章中已经提出使用 CASIA-WebFace-A 来训练人脸特征提取模型，每个类别中随机抽取 45 张作为训练集，另外的 5 张作为训练过程测试集。TED_LIUM 库包含的 1242 个人的样本，在 CASIA-WebFace-B 中随机选取 1242 个类别并且每个类别随机选出 25 张人脸照片。将 1242 个人的语音和人脸按照类别随机匹配到一起，生成混合库。混合库中每个类别包含 25 个语音-人脸对。混合库将其中 1000 人的样本用来训练融合识别模型。其中每类取 20 个样本用来训练模型，另外 5 个作为训练过程测试样本。剩下的 242 类作为训练好的模型的开集测试集。在这 242 类开集测试集中，通过类内和类间样本的随机搭配，生成 1000 对正测试样本（同一个人）和 1000 对负测试样本（非同一个人）来测试训练好的模型。

本文实验使用的数据库具体分布如下表：

表 3 双模态融合数据库

| 数据库类型 | 人脸特征提取模型训练集 | 双模态识别模型 | | |
|---|---|---|---|---|
| | | 训练集 | 训练测试集 | 测试集 |
| CASIA-WebFace-A | 8216*50 | —— | | |
| CASIA-WebFace-B | —— | $1000 \times 20$ | $1000 \times 5$ | $242 \times 25$ |
| TED_LIUM | —— | $1000 \times 20$ | $1000 \times 5$ | $242 \times 25$ |

## 5.3 实验平台

本文实验使用的是 Ubuntu（64 位）操作系统，内存 16GB。CPU 是 Intel i5 3450 主频 3.1GHz ，6M 三级缓存。GPU 为 NVIDIA GeForce GTX Titan X，显存 12G。人脸特征提取模型及双模态识别模型的训练使用的深度学习框架为 Caffe，语音信号的 PLP（PCA）及 I-Vector 特征的提取使用的是 Matlab 软件平台。

## 5.4 双模态生物特征识别模型结构

本文双模态生物特征识别的系统识别网络模型使用的是卷积神经网络和深度置信网络。首先根据第三章所述的内容，通过人脸训练库对人脸特征提取模型进行训练，使用的训练网络为本文提出的 Vgg_Face 改进网络结构。根据第三章的实验显示，训练好的网络模型在 LFW 库上能够实现 94.97% 的识别率。提取出倒数第二个全连接层输出的数据作为图像特征，维度为 2048。然后将此 2048 维图像特征和说话人语音的 PLP（PCA）+I-Vector 特征进行串行融合形成双模态融合特征，此特征共 3148 维。将融合特征送入含有 4 个隐层的 DBN 中，此 DBN 由 4 个 RBM 和一个 SoftMax 层构成。其中 4 个 RBM 层的节点数分别为（$3148 \times 3000$、$3000 \times 3000$、$3000 \times 3000$、$3000 \times ***$，** 为训练库样本类型数）。双模态特征融合识别具体网络结构如下：

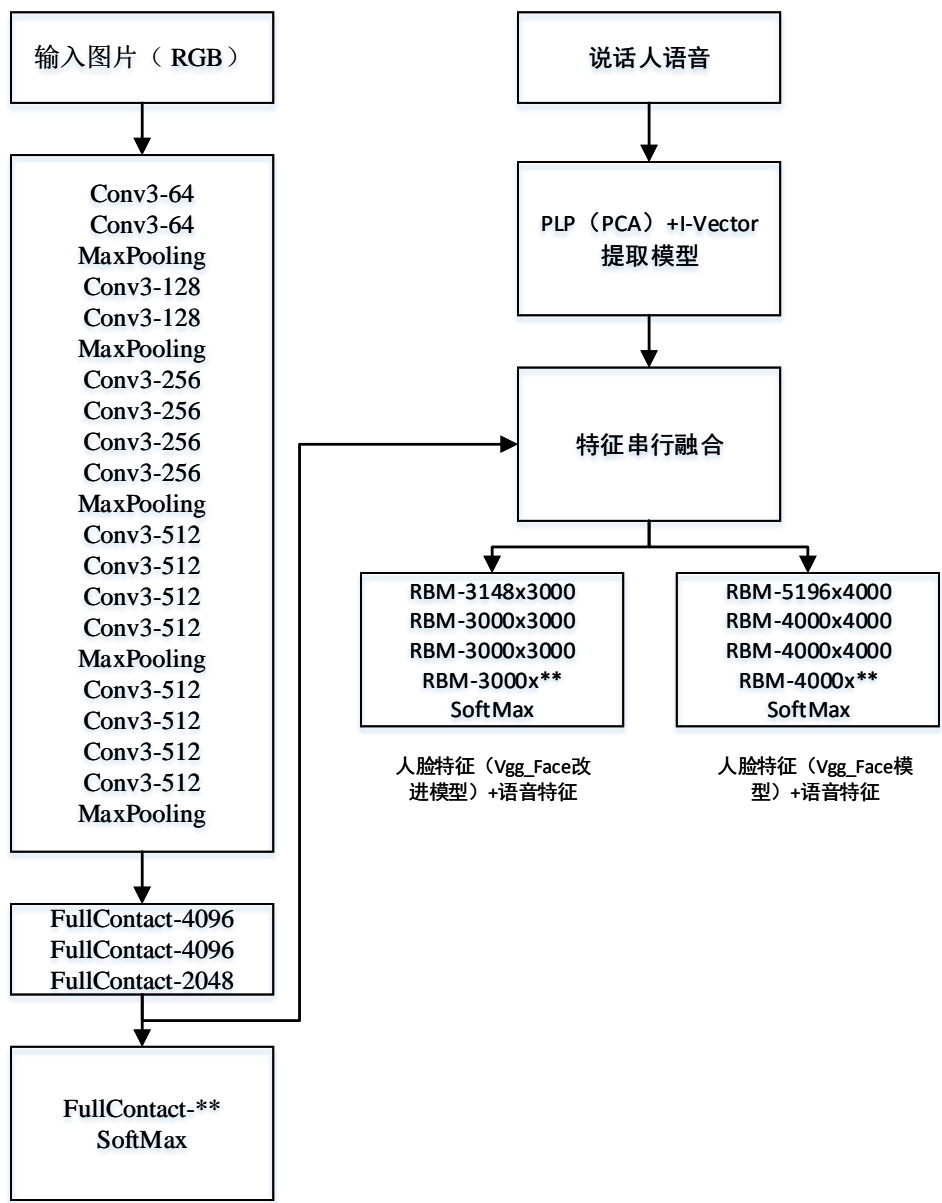

图 5.2 双模态生物特征识别网络模型结构图

## 5.5 实验仿真及分析

对于本文的双模态生物特征融合识别实验，进行了如下的实验步骤和对比设置：

1. 对于人脸特征提取模型，本文实验使用 CASIA-WebFace-A 库来训练网络，为了后面的对比试验，使用的网络包括 Vgg_Face 和 Vgg_Face 改进型。对于两个类型的网络模型,下面的实验步骤和数据样本的分步都是一样的,在此对 Vgg_Face 改进网络进行分析,此步骤和样本分步对 Vgg_Face 模型的实验同样适用。每类 50 张样本，将其中 45 张用于网络参数的迭代训练，另外的 5 张用于训练过程中的测试。通过参数调节，最终获得两种人脸特征提取模型。将 SoftMax 层去除，用最后一层全连接层的输出作为人脸特征。用此模型对双模态混合库中的人脸

图片进行特征提取，包括用于训练融合特征识别网络模型的 1000 类和用于融合特征识别网络模型测试的 242 类。一共 1242 类，共 31050 张人脸照片。

2. 对于说话人特征提取部分，本文使用 NIST04 和 NIST05 训练 UBM，然后使用 NIST06 和 NIST07 训练全局差异空间矩阵。使用训练好的 UBM 和全局差异空间矩阵，对融合识别样本库中的语音部分进行 I-Vector 特征提取，一共 1242 类，31050 句话。每个样本提取长度为 600 的 I-Vector 特征。然后对这 31050 个样本提取 PLP 系数，并且使用 PCA 对其降维到 500 维。将 I-Vector 特征和 PLP（PCA）特征进行串行融合，生成说话人语音特征，每个特征长度为 1100 维。

3. 通过前两步提取到人脸和说话人语音特征，使用串行特征融合方式生成双模态融合特征。至此，得到了融合特征识别模型的训练集样本（1000*25 个）特征，正样本测试集样本（1000 对）特征，负样本测试集样本（1000 对）特征。

4. 使用训练集对融合特征识别网络模型进行训练，并记录下测试过程的识别率变化，当识别率达到自己的预期时，对此网络使用正负样本集进行开集测试。

对于 Vgg_face 模型，Omkar M. Parkhi 已经对其进行了优秀的参数优化，该网络模型可以直接用于人脸特征提取，但是为了保证实验的公平性，本文保证 Omkar M. Parkhi 设定的原始参数，并且使用 CASIA-WebFace-A 对该网络进行微调，以保证该网络和 Vgg_Face 改进模型的训练过程一致。Vgg_Face 模型的微调过程识别率变化曲线如下图所示：

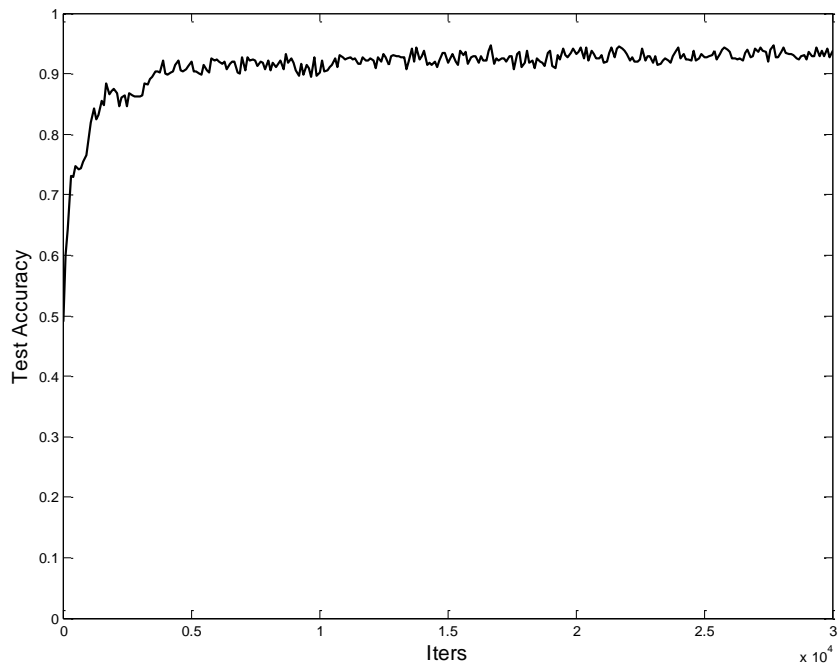

图 5.3 Vgg_Face 模型微调识别率变化曲线

可以从上图中看到，在微调 Vgg_Face 模型的时候，训练测试库识别率第一次迭代测试的识别率就达到了 54.76%，并且上升速度很快。由于 Vgg_Face 模型通过 2.6M 数据训练之后，对人脸提取特征具有很好的区分性。经过对 CASIA-WebFace-A 所有样本迭代一次之后，具有对 CASIA-WebFace-A 样本的区分性，所以第一次迭代测试就能达到 50% 多的识别效果。经过 30000 次迭代训练，最终的识别率稳定在 94.12%。

训练好之后使用 CASIA-WebFace-B 库对该模型进行开集测试，并获得其 ROC 曲线为：

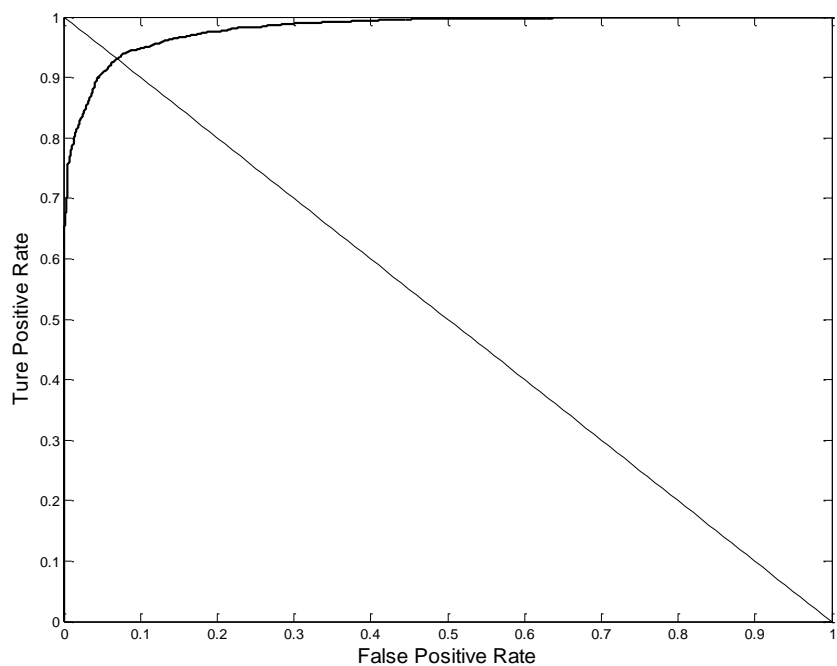

图 5.4 Vgg_Face 模型在 CASIA-WebFace-B 库上的 ROC 曲线

在 CASIA-WebFace-B 库上开集测试，通过调节距离阈值，在等错误率点处，模型对 CASIA-WebFace-B 的识别率为 95.26%。这个开集测试的识别效果要比训练的时候要好，正常情况下，网络模型经过大量数据进行训练之后，在开集测试的正负样本之间做目标确认要比目标识别识别率高。Vgg_Face 模型微调之后在 CASIA-WebFace-B 库上的识别率没有其原模型在 LFW 库上的识别率高的原因有：

1. CASIA-WebFace-B 相对于 LFW 库中的样本更加复杂，样本含噪跟多；

2. 本文对 Vgg_Face 模型在 CASIA-WebFace-B 库上微调时，设置的训练参数没有其原来的参数好，导致识别率下降；

对于 Vgg_Face 改进模型，添加了一个全连接层，并且将最后一层全连接层的输出节点调整为 2048，降低了输出特征的维度。提取处 Vgg_Face 模型的节点参数信息，并且保存在 Vgg_Face 改进网络相应的节点处。随机初始化新添加的全连接层参数。使用

CASIA-WebFace-A 对该网络进行训练，这个部分实验在第三章中已经完成，使用第三章中训练好的模型对 CASIA-WebFace-B 做开集测试。模型在 CASIA-WebFace-B 库上的 ROC 曲线如下图：

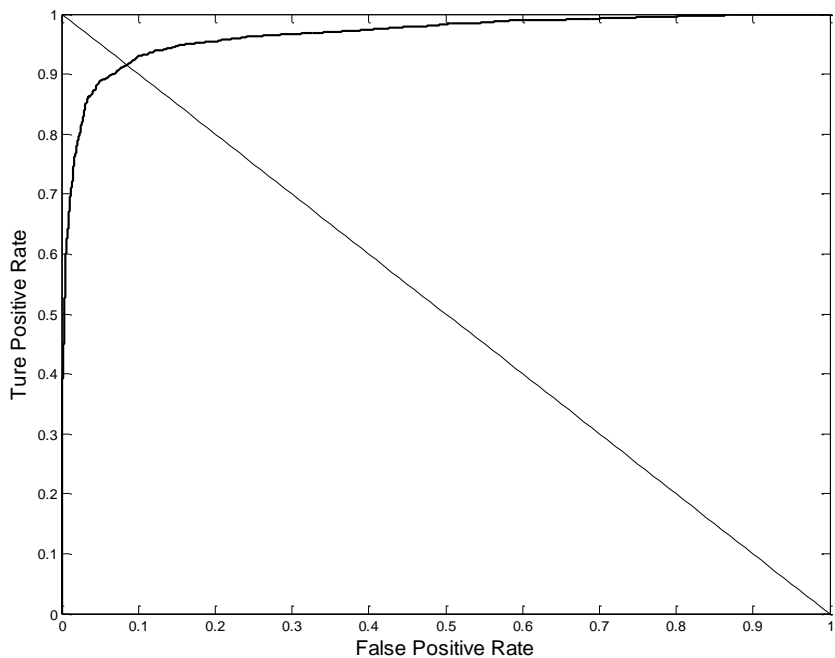

图 5.5 Vgg_Face 改进模型在 CASIA-WebFace-B 库上的 ROC 曲线

  Vgg_Face 改进模型在 CASIA-WebFace-B 库上的开集测试识别率为 93.46%，这个识别结果和其在 LFW 库上的结果相比也有一定的下降。

  对于 5.4 节中系统网络模型中的融合识别模型部分，将 Vgg_Face 和 Vgg_Face 改进模型提取出来的人脸特征和语音（20s）特征融合后，对其进行训练。针对两种融合识别模型的差异性，经过多次调参实验，得到下面两组针对不同模型的参数设置。以保证训练得到的模型具有其最优秀的性能。

  Vgg_Face 模型提取的人脸特征（4096 维）和语音特征（1100 维）的融合特征识别网络的训练参数设置为：

1. 初始学习率为 0.005；

2. 学习率下降策略为"Step"；

3. 学习率下降步长为 10000 迭代次数；

4. 学习率下降 Gamma 参数为 0.90；

5. 总迭代次数为 150000 次；

6. 训练数据的 batchsize 为 64，训练测试数据的 batchsize 为 32；

  Vgg_Face 改进模型提取的人脸特征（2048）和语音特征（1100）的融合特征识别

网络的训练参数设置为：

1. 初始学习率为 0.0005；

2. 学习率下降策略为"Step"；

3. 学习率下降步长为 10000 迭代次数；

4. 学习率下降 Gamma 参数为 0.96；

5. 总迭代次数为 150000 次；

6. 训练数据的 batchsize 为 64，训练测试数据的 batchsize 为 32；

迭代学习过程中训练测试集的损失函数值变化如下图：

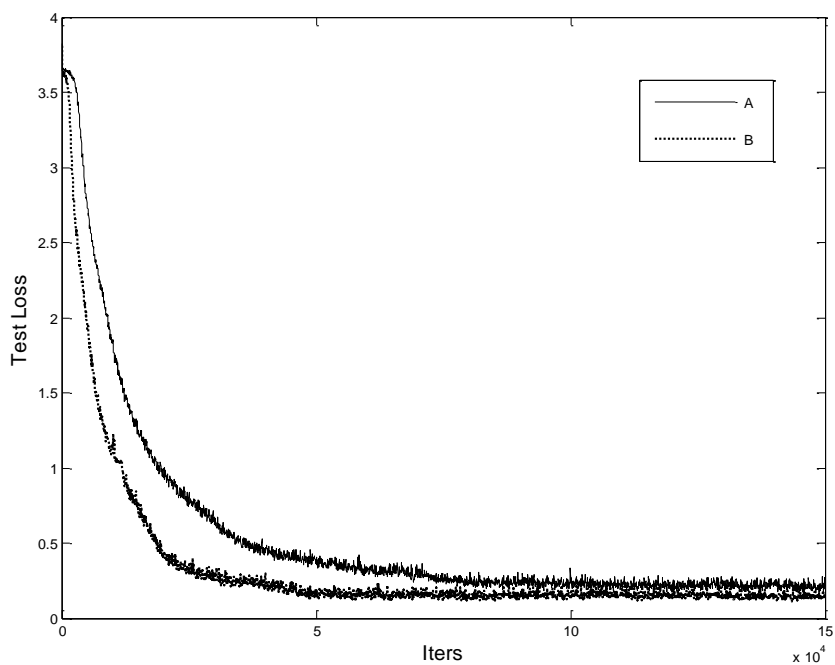

图 5.6 融合识别网络模型训练测试损失函数值变化曲线

线 A：融合 Vgg_Face 模型人脸特征的模型的训练测试损失函数值得变化曲线

线 B：融合 Vgg_Face 改进模型人脸特征的模型的训练测试损失函数值得变化曲线。

由于 Vgg_Face 模型提取出来的人脸特征维度为 4096，加上语音特征之后达到 5196 维。并且针对这个特征设计的 DBN 网络节点规模也比 Vgg_Face 改进模型对应的 DBN 网络节点规模大。导致模型迭代学习过程中训练测试集的损失函数值下降的要慢。在 7 万多次的迭代之后，才趋于稳定。而 Vgg_Face 改进模型对应的网络模型在 4 万多次的迭代就已经趋于稳定。并且训练的过程中，网络规模大的每一次迭代所花费的时间也要比规模较小的网络模型多。

这两个网络模型的训练过程的测试识别率变化曲线如下：

将训练好的模型在正负样本集中进行开集测试，调节阈值，对 2000 个样本识别结

果统计，并给出 ROC 曲线：

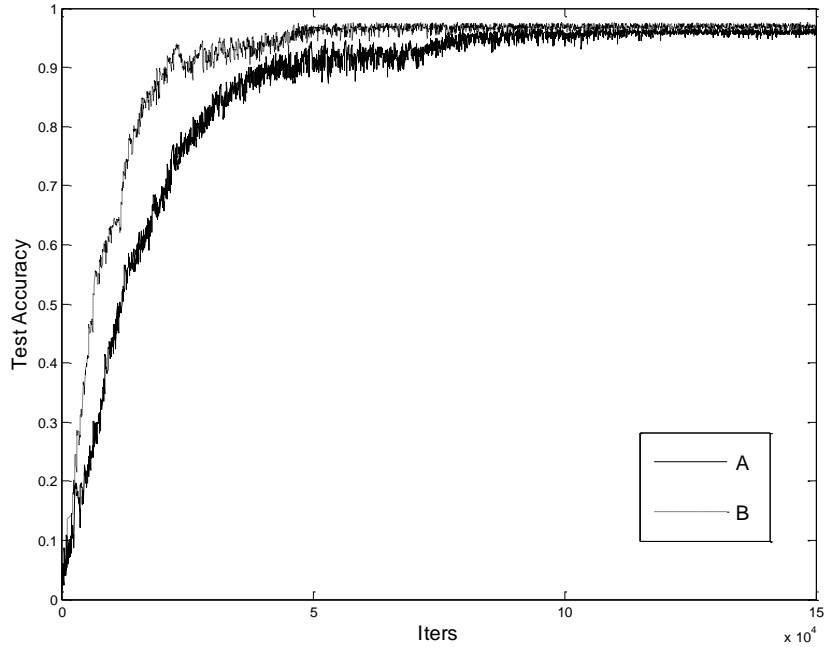

图 5.7 融合识别网络模型训练过程中测试识别率变化曲线

线 A： Vgg_Face 模型对应的融合识别模型的训练测试集识别率变化曲线

线 B： Vgg_Face 改进模型对应的融合识别模型的训练测试损失函数值得变化曲线。

从上图可以看到，Vgg_Face 改进模型对应的融合特征识别网络模型在训练过程中，其对训练样本的学习速度要比 Vgg_Face 模型对应的网络模型快。在 4 万次迭代的时候就已经达到 94.48% 的识别率，在后续的学习迭代中最终识别率稳定在 96.19%。Vgg_Face 模型对应的识别模型在 8.5 万次迭代时达到 94% 左右，最终的识别率稳定在 95.46%。训练过程中的识别率不能很好地反应该识别模型的识别效果。因此在 242 类组成的 2000 对正负样本上对两个模型做开集测试。调节余弦距离阈值，并且统计识别结果。通过 ROC 曲线反应如下：

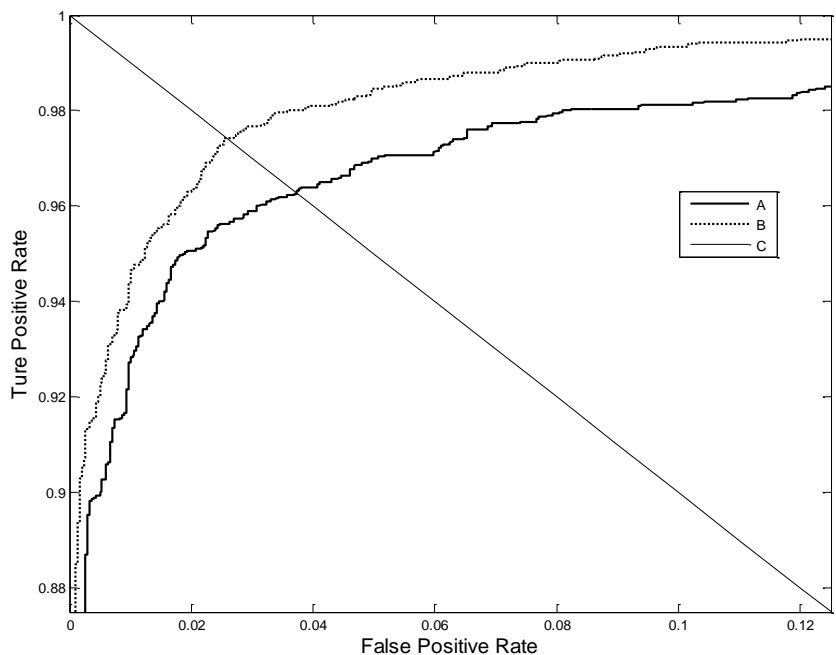

图 5.8 融合识别网络模型对于开集正负样本测试集的 ROC 曲线
线 A：Vgg_Face 模型对应的融合识别模型 ROC 曲线
线 B：Vgg_Face 改进模型对应的融合识别模型的 ROC 曲线
线 C：等错误率线

在等错误率处，Vgg_Face 对应的融合识别模型在开集上的识别率为 96.24%，Vgg_Face 改进网络对应的融合识别模型的识别率为 97.38%。改进型网络模型的识别率高于原模型的识别率，是由于人脸特征维度和说话人语音特征维度相差太大，导致特征分布不平衡。可以得出，Vgg_Face 改进网络提取出来的人脸特征融合上说话人语音特征要比直接使用 Vgg_Face 模型提取出来的人脸特征融合说话人语音特征优秀。虽然人脸特征提取模型在训练的时候，改进型的网络结构要比 Vgg_Face 的复杂，并且训练出来的模型提取人脸特征相较于原模型性能要差一点。但是新模型提取出来的人脸特征是原模型的一半，并且和说话人语音特征相结合之后，融合深度神经网络的训练速度大幅提高，并且最终的识别率要比未改进的识别率高。

针对于单独的人脸识别和说话人识别结果来说，融合特征识别网络在识别率上具有一定的优势，在使用相同的人脸图像和说话人语音的情况下，相比于单独的人脸识别系统提高了 3.92% 的识别率，相比于说话人识别结果提高了 11.41%。但是双模态特征融合识别系统更为复杂，需要同时提取人脸和说话人特征。并且前期的特征提取模型训练需要更多的数据样本和训练时间。这些问题需要在实际环境应用中细加考略，权衡识别效果和计算量的代价。

## 5.6 本章小结

　　在本章中，提出了本文重要的双模态生物特征融合识别的算法和系统识别模型。首先在第一小节中提出了双模态特征融合的方法为串行融合方式，即将人脸特征和说话人语音特征以串行的方式连接起来送入新的识别模型中，融合的时候不需要考虑两种特征的权重问题。通过深度神经网络的节点参数调节，可以更好的为每一维特征计算出其对应的权重。然后在第二节中提出了本文双模态识别模型训练的数据库，为了得到适合深度神经网络训练的大数据库，本文将 TED-LIUM 和 CASIA-Webface-B 中的样本按照类别随机组合在一起，生成新的融合数据库。在第三节中介绍了本文实验的硬件和软件平台。然后在第四节中提出了本文创新的双模态生物特征识别系统的识别模型结构。该网络模型包含人脸特征和说话特征提取部分和融合特征识别网络模型。具体的人脸提取模块，说话人语音特征提取模块和融合识别模块都可以根据实际需求进行更换，具有很好的灵活性。在最后一节中，对上节提出的识别系统进行了实验论证，通过对比 Vgg_Face 模型改进前后的准确率，分析了人脸特征提取模型的改进对融合识别模型的影响。然后对比了人脸和说话人语音融合前后的识别率，证实了双模态生物特征识别相较于单模太生物特征识别的优越性。

# 第六章 总结与展望

## 6.1 工作总结

得益于深度学习的快速发展，使得基于传统机器学习的模式识别性能大幅提升。人脸识别和说话人识别分别在其各自的领域内都得到了飞速的发展。作为人类特有的特征，人脸和语音信息具有优秀的特异性。并且这两类生物特征识别具有广泛的使用场景和应用价值。本文将人脸和说话人语音信号合并到一个系统中，通过深度学习对融合特征训练得到基于人脸与说话人语音融合特征的识别模型。该系统具有深渊的研究价值和实际应用场景。

在上面各章中，已经对双模态生物特征识别系统的各个关键点做了详细的解释和推导。现在针对前面的全部工作做一个简要的总结：

1. 首先介绍了人脸识别、说话人识别和双模态融合识别的研究历史、发展现状和目前主要存在的研究难点。经过了将近 90 年的发展历史，人脸和说话人识别已经得到了长远的发展历程，全世界越来越多的研究人员加入这个领域。并且在现实生活中，人们已经广泛地使用基于人脸和说话人识别技术。

2. 针对目前最新的深度学习研究成果，计算和推导了本文使用的卷积神经网络和深度置信网络的相关关键算法。并且对深度神经网络训练过程中的优化与调参原理做了详尽的介绍。对后续的深度神经网络的训练有着至关重要的帮助。

3. 对系统人脸特征提取部分，做了详尽的理论推导和实验验证。首先使用 AdaBoost 训练 Haar-Like 特征得到人脸检测模型。然后使用 SDM 算法找到人脸关键点的位置，调用图像放射变换算法对人脸进行校正。对 Vgg_Face 网络模型进行深入研究，并且在其网络基础上结合融合特征的需求进行了一定的改进。最后使用 CASIA-Webface-A 对改进网络进行训练，在 LFW 库上对识别模型进行开集测试，最终得到 95.01%的识别率。

4. 针对说话人语音特征提取部分，本文推导了求 PLP 系数的算法流程，并且提出使用 PCA 对说话人语音特征进行降维处理。然后对提取出来的语音 MFCC 特征系数使用常向量的方法映射到不变子空间内，并且使用 MFCC 特征系数作为 GMM-UBM 和全局差异特征矩阵的输入进行提取模型的训练。得到 I-Vector 特征之后，对 PLP（PCA）特征和 I-Vector 特征进行串行特征融合处理。为了验证提取的语音参数具有说话人识别的能力。在 TED-LIUM 库中随机抽取 1000 个样本作为训练数据集合，并且手工切分出 5s、10s、15s、20s 长度的子样本。

本文提出使用 DBN 对训练集得到的语音参数训练说话人识别网络模型。最终在开集测试库上得到了 87.76%（20s）的识别率。进一步证实了说话人语音特征的可用性。

5. 最后本文提出了双模态生物特征的融合识别算法和系统框架。通过模块化的系统设计，方便了双模态生物特征识别系统的调试和改进。鉴于深度学习网络训练需要大量数据的双模态人脸-语音融合识别库，本文提出了使用 TED_LIUM 库和 CASIA-Webface 库基于类别随机组合成融合数据库。并通过该融合数据库训练提出的系统网络识别模型。通过和单模态识别结果对比和 Vgg_Face 原模型提取特征的识别系统对比，验证了本文设计的双模态生物特征识别系统的性能。

## 6.2 研究展望

在硕士论文的撰写过程中，本人认真总结了过去三年的研究工作。笔者深知此文只是双模态生物特征识别中的九牛之一毛，还有很多深入的问题还没来得及仔细思考。距离达到更优秀的实际应用，本文提出的观点和研究结果还需要进一步的改进和突破。对今后的研究工作有以下展望：

1. 本文使用的人脸检测的 AdaBoost 算法训练过程较长，识别效果还没有达到非常优秀的程度，和目前最新的人脸检测算法比还有很大的进步空间。鉴于 AdaBoost 算法已经满足本文实验要求的指标，本文并未对人脸检测算法有过多讨论。今后可以使用更新的人脸检测算法，可以进一步增快系统的速度。

2. 本文使用的人脸识别网络是对 Vgg_Face 模型的网络结构进行一定的改进，此改进仅为了提高双模态生物特征识别识别率而做出的策略。今后的工作中不限于 Vgg_Face 改进模型提取的人脸特征，可以使用 FaceNet 或者 DeepID 等网络提取出来的人脸特征作为输入特征。

3. 本文中语音特征使用的是 PLP 和 I-Vector 特征，在今后的研究中，可以使用其他的特征代替这两个特征。或者使用另外的特征和 I-Vector 或是 PLP 相融合。

4. 双模态生物特征识别系统中的 DBN 的设计是笔者在实际工作中手动调节出来的网络结构，不排除有其他更优秀的网络模型作为融合识别模型。

5. 本文使用的融合数据训练库包含 1000 个类别，25000 个样本。这个数量的样本对于深度学习来说还有进一步增大的空间。并且开集测试的 2000 对正负样本还有进一步增加的需求。

6. 对于深度神经网络训练的参数设定和训练数据训练量是笔者手动调节的最优结果，今后的工作中可以设计自动设定参数，逐步优化初始参数的算法，以减少手动调节的缺陷，得到最优的网络结构和训练参数。

# 参考文献

# 致谢

　　转眼研究生学生生涯即将结束。在东南大学求学的三年里，我收获了丰富的专业知识，更重要的是收获了宝贵的人生阅历。在此，我首先要感谢我的父母，女朋友对我在生活上的支持，有你们做我坚持的后盾，使我能够全心全意的完成学业。

　　谨向我的导师邹采荣教授表示深深的感谢，感谢邹老师在这三年中对我学业的悉心指导。此外，感谢实验室的赵力教授，感谢赵老师对我研究生三年的关怀和指导。赵老师严谨的治学态度，求实的作风以及对研究孜孜不倦的追求一直都在陶冶和熏陶着我，让我终生受益。

　　另外，我还要向我研究生期间实习过的南京凡豆信息科技公司的黄程韦博士和公司的相关工作人员表示衷心的感谢。黄程韦博士对我在专业知识上给予了相当大的帮助。他渊博的学识、个人魅力、积极的工作态度给我的人生产生了深远的影响，将成为我终生享用的宝贵财富。

　　感谢实验室的所有同仁，包括梁瑞宇博士、张明阳博士、谢跃博士、章勤杰、丁一坤、张奇、邹炉庚、刘成宇、皮慧、卢小芳、陈晟、黄晨、何旭。感谢这个积极向上的团队，对我研究生求学生涯宝贵的指导和帮助。感谢我的室友刘耘、张希龙和张祺威。感谢他们对我学习和生活上的帮助，从他们身上得到很多知识和快乐。

　　感谢答辩组的各位老师们，感谢他们在百忙之中抽出宝贵的时间对我的论文给予指导和建议。

# 攻读硕士学位期间发表的论文